\documentclass[11pt]{article}

\usepackage[final]{acl}

\usepackage{times}
\usepackage{latexsym}

\usepackage[T1]{fontenc}

\usepackage[utf8]{inputenc}

\usepackage{microtype}

\usepackage{inconsolata}

\usepackage{graphicx}
\usepackage{amsmath}
\usepackage{booktabs} 
\usepackage{listings}
\usepackage{xcolor}
\definecolor{sectionhead}{HTML}{1F4E79}

\usepackage{makecell}
\usepackage{placeins}

\lstdefinestyle{promptstyle}{
  basicstyle=\ttfamily\footnotesize,
  breaklines=true,
  breakatwhitespace=true,
  escapeinside={(*}{*)},
  showstringspaces=false,
  columns=fullflexible,
  upquote=true,
  keepspaces=true
}

\usepackage{array}
\usepackage{lstautogobble}

\usepackage{longtable}
\usepackage{multirow}

\usepackage{xcolor}
\usepackage{xspace}
\usepackage{enumitem}

\newcommand{\pidim}[1]{{\color{purple!60!black}\textsc{#1}}\xspace}
\newcommand{\pisub}[1]{{\color{green!40!black}\texttt{#1}}\xspace}
\newcommand{\appeal}[1]{{\color{green!20!black}\textbf{#1}}\xspace}

\newcommand{\model}[1]{\texttt{#1}\xspace}
\newcommand{\mypar}[1]{\noindent\textbf{#1}}

\title{Persuasion Index: A Theory-Guided Framework for Persuasion Analysis}

\author{
\textbf{Liancheng Gong}$^{1}$ \quad
\textbf{Zhiyang Wang}$^{2}$ \quad
\textbf{Yiwei Xu}$^{1}$ \quad
\textbf{Julia Mendelsohn}$^{1}$ \\
$^{1}$University of Maryland, College Park \quad 
$^{2}$New York University \\
\texttt{\{gonglc, juliame\}@umd.edu}
}

\begin{document}
\maketitle

\begin{abstract}
Identifying persuasive rhetorical cues is critical across domains, from detecting information manipulation and improving AI safety to advancing public health communication. We propose the \textbf{Persuasion Index} (PI), a taxonomy of 15 dimensions grounded in persuasion theories from psychology and communication, and one transparent implementation using 55 sub-features built from lexicons and rule-based detectors. The taxonomy is modular: individual detectors can be replaced while preserving the theoretical structure. We evaluate PI on four public datasets for English argumentative text that vary in domain, style, and outcome measures, and show that PI provides a shared feature space for interpreting rhetorical patterns associated with persuasion-related outcomes. Linear models show that PI features carry meaningful predictive signal while remaining computationally lightweight. Dimension-level analyses reveal recurring associations between PI dimensions and persuasion outcomes across datasets, while also highlighting topic- and stance-specific variation. We release PI as an open-source package and web interface for principled and auditable analysis of human and AI-mediated communication.\footnote{The Python package, source code, and other reproducibility materials are available at \url{https://github.com/krystalgong/Persuasion_Index_Code}. The interactive PI web interface is available at \url{https://pi.umiacs.umd.edu}.}

\end{abstract}

\section{Introduction}
\begin{figure}[t]
  \includegraphics[width=\columnwidth]{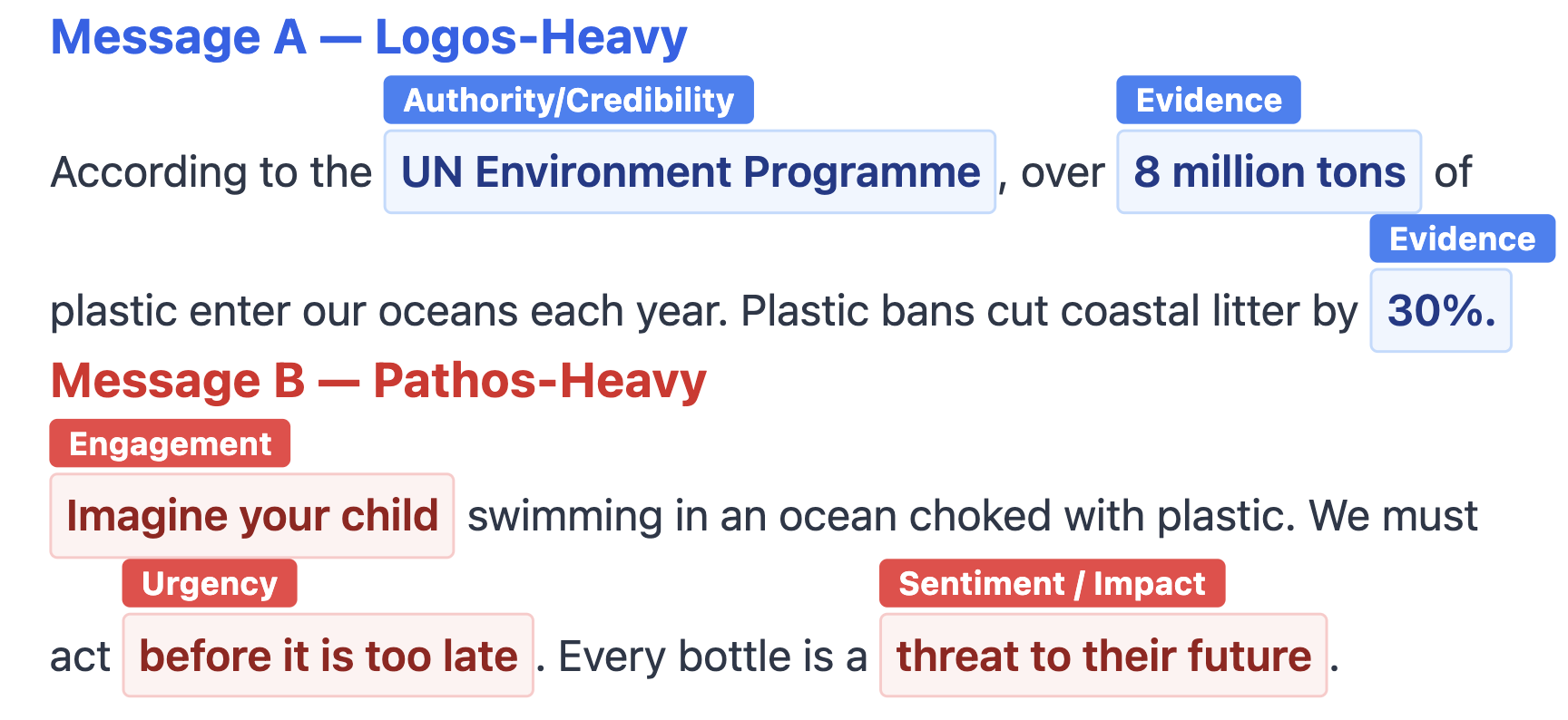}
  \caption{Two arguments against plastic pollution. Message A (\appeal{Logos}-heavy) relies on factual grounding through statistics and institutional citation; Message B (\appeal{Pathos}-heavy) relies on emotional engagement, urgency, and threat framing.}
  \label{fig:two_arguments}
\end{figure}

Persuasion is the process whereby a message changes an audience's beliefs, attitudes, or behavior \citep{okeefePersuasionTheoryResearch2015}. Understanding persuasion has become increasingly urgent as AI systems gain influence over public discourse at scale \citep{salviConversationalPersuasivenessGPT42025, hackenburgEvaluatingPersuasiveInfluence2024, bozdagMustReadComprehensive2026}. LLMs have demonstrated measurable attitude change in large-scale political persuasion experiments \citep{hackenburgLeversPoliticalPersuasion2025}; AI-generated content accelerates the spread of health misinformation \citep{augensteinFactualityChallengesEra2024}; and conversational agents increasingly operate as persuaders in both human-facing and multi-agent systems \citep{breumPersuasivePowerLarge2024,matzPotentialGenerativeAI2024,dashPersuasivePotentialAIparaphrased2025,chengStrategicPersuasionLanguage2026}. In these settings, detecting persuasion is not enough; auditing it requires identifying the underlying rhetorical cues deployed, as logical, emotional, and credibility-based appeals call for distinct responses.

A common approach in prior NLP work is to operationalize persuasion as a prediction target, although what counts as a persuasive outcome varies substantially across datasets and tasks. These outcomes range from pairwise convincingness judgments \citep{habernalWhichArgumentMore2016} to self-reported opinion change \citep{tanWinningArgumentsInteraction2016} and experimentally measured pre--post attitude shifts \citep{MeasuringPersuasivenessLanguage}. Models ranging from bag-of-words classifiers to fine-tuned transformers and LLM-based judges can achieve strong predictive performance, but their outputs are typically binary labels or scalar scores \citep{yangLetsMakeYour2019,bassiDecodingPersuasionSurvey2024}. For example, Figure~\ref{fig:two_arguments} shows two arguments reaching the same conclusion but deploying entirely different rhetorical cues, a distinction that scalar persuasiveness predictions absorb but cannot expose. While post-hoc attribution methods can explain which tokens or features influenced a particular model, they do not provide a stable vocabulary for comparing persuasive language across datasets, domains, or communicative settings. This creates a need for an interpretable framework that represents persuasive language directly in terms of theory-grounded rhetorical dimensions, independent of any particular prediction model or dataset-specific label scheme.

We address this gap by proposing the \textbf{Persuasion Index} (PI) with two distinct components. The core component is a taxonomy of 15 dimensions grounded in well-established theories from psychology, communication, and rhetoric, mapped onto the Aristotelian triad of \appeal{Logos}, \appeal{Ethos}, and \appeal{Pathos}. The second component is an implementation of this taxonomy through 55 granular sub-features, implemented via theory-seeded lexicons, LLM-based expansion, validated computational linguistics resources, and lightweight rule-based detectors. The current PI implementation targets English text. Importantly, PI scores characterize observable rhetorical strategies encoded in language. They do not determine whether a claim is factually true, an inference is logically valid, or an argument is normatively high quality. PI is modular by design: we provide a complete implementation, but individual sub-feature measurements are interchangeable and extensible without disrupting the taxonomy. Unlike opaque neural representations, PI extracts features directly from text, ensuring every score traces back to specific lexical or structural patterns. 

We make the following contributions:
\begin{enumerate}[topsep=0pt, noitemsep]
    \item A theory-guided taxonomy of 15 persuasive rhetorical dimensions
    \item An implementation via interpretable lexicon and rule-based features
    \item Construct and predictive validation through human judgments and four heterogeneous persuasion datasets
    \item Analysis of PI dimensions and persuasive outcomes, revealing both regularities and variation across datasets, topics, and stances
    \item An open-source package and public web interface for real-time argument analysis (Appendix~\ref{appendix:web_interface}), enabling researchers to apply and extend the framework on their own data
\end{enumerate}

The PI can broadly support interdisciplinary persuasion research, with use cases including content analysis or experimental stimuli validation in social science research, as well as AI safety audits of persuasive content in human-LLM or multi-agent interactions.

\section{Background}
\label{sec:related}
Research in psychology and communication has long treated persuasion as multidimensional rather than a single scalar property. Dual-process theories such as the Elaboration Likelihood Model and the Heuristic-Systematic Model distinguish between \textit{central/systematic processing}, characterized by rigorous evaluation of evidence and argument quality, and \textit{peripheral/heuristic processing}, characterized by reliance on cues such as source credibility and emotional tone \citep{pettyElaborationLikelihoodModel1986, chaikenHeuristicSystematicInformation1980}. PI builds on this tradition by including rhetorical dimensions corresponding to both routes. Section~\ref{sec:framework} details the theoretical anchors for each PI dimension.

Prior NLP work has largely operationalized persuasion through task- and domain-specific taxonomies, each emphasizing a narrow subset of persuasive mechanisms. Research on altruistic requests, crowdfunding and charitable giving foregrounds peripheral cues with an emphasis on social and relational strategies \citep{althoffHowAskFavor2014,mitraLanguageThatGets2014,wangPersuasionGoodPersonalized2019,yangLetsMakeYour2019,chenWeaklySupervisedHierarchicalModels2021}. Conversely, argument mining research primarily targets logic-oriented argument structure and quality \citep{wachsmuthComputationalArgumentationQuality2017,ghoshCoarsegrainedArgumentationFeatures2016,carlileGiveMeMore2018,gretzLargescaleDatasetArgument2020,toledoAutomaticArgumentQuality2019,joshiArgAnalysis35KLargescaleDataset2023,rombergPerspectivistTurnArgument2025}. Propaganda detection research defines and labels fine-grained persuasive techniques in news and political texts \citep{dasanmartinoFineGrainedAnalysisPropaganda2019,dimitrovSemEval2021Task62021,piskorskiSemEval2023Task32023,modzelewskiPCoTPersuasionAugmentedChain2025,sajwaniFRAPPEFRAmingPersuasion2024}. 

A second source of fragmentation is that datasets define persuasive success in heterogeneous ways, including pairwise convincingness judgments \citep{habernalWhichArgumentMore2016,gleizeAreYouConvinced2019,toledoAutomaticArgumentQuality2019,gretzLargescaleDatasetArgument2020}, scalar ratings of argument quality \citep{ghoshCoarsegrainedArgumentationFeatures2016,carlileGiveMeMore2018,joshiArgAnalysis35KLargescaleDataset2023}. A complementary paradigm measures pre--post attitude change, using explicit markers of opinion shifts or survey responses \citep{tanWinningArgumentsInteraction2016,luuMeasuringOnlineDebaters2019,montiLanguageOpinionChange2022,MeasuringPersuasivenessLanguage}. Thus, existing resources provide strong local accounts of persuasive language, but their domain-specific taxonomies and heterogeneous outcome labels make it difficult to compare how persuasive language functions across domains or determine if observed patterns generalize to other corpora.

Like many other tasks, NLP research often emphasizes predictive performance over explanatory power, modeling has trended toward higher-capacity, lower-interpretability architectures. Early work combines hand-crafted linguistic features, argument structure, and interaction dynamics with linear or tree-based models \citep{stabAnnotatingArgumentComponents2014,somasundaranDetectingArguingSentiment2007,danescu-niculescu-mizilComputationalApproachPoliteness2013,althoffHowAskFavor2014,mitraLanguageThatGets2014,ghoshCoarsegrainedArgumentationFeatures2016,wangWinningMeritsJoint2017}. Later studies adopt neural architectures for predicting persuasiveness, argument quality, and opinion change \citep{hideyPersuasiveInfluenceDetection2018,yangLetsMakeYour2019,wangPersuasionGoodPersonalized2019,toledoAutomaticArgumentQuality2019,gretzLargescaleDatasetArgument2020,chenWeaklySupervisedHierarchicalModels2021,joshiArgAnalysis35KLargescaleDataset2023}. More recently, LLMs act as persuaders in controlled experiments, where their messages are evaluated via human attitude shifts \citep{breumPersuasivePowerLarge2024,matzPotentialGenerativeAI2024,baiLLMgeneratedMessagesCan2025,hackenburgEvaluatingPersuasiveInfluence2024,hackenburgComparingPersuasivenessRoleplaying2025,salviConversationalPersuasivenessGPT42025,chengStrategicPersuasionLanguage2026,costelloDurablyReducingConspiracy2024,dashPersuasivePotentialAIparaphrased2025,sharmaGenerativeEchoChamber2024}, and as judges of persuasive strength and strategy use \citep{pauliMeasuringBenchmarkingLarge2025,breumPersuasivePowerLarge2024,salviConversationalPersuasivenessGPT42025,MeasuringPersuasivenessLanguage,modzelewskiPCoTPersuasionAugmentedChain2025}. As LLMs increasingly generate, personalize, and evaluate persuasive messages, it is necessary for NLP research to not just predict persuasion, but provide interpretable measures of underlying rhetorical strategies.

Recent work has begun to address this need by grounding predictions in theory. \citet{sudharsanCrossDomainPersuasionDetection2025} show that incorporating argumentative components and semantic types improve cross-domain generalization. \citet{hoangHybridTheoryDatadriven2025} combine psychological theory with LLM-generated feature ratings in interpretable classifiers. PI shares this motivation but differs in scope and operationalization. Rather than focusing primarily on argumentative structure, PI represents persuasive language across logical, affective, and credibility appeals. While the PI framework is modular, the implementation we provide uses transparent lexicon and rule-based features. These choices make PI a shared, auditable representation for comparing persuasive language across contexts and describing linguistic strategies beyond outcome prediction.

\section{Theory-Grounded Framework}
\label{sec:framework}

\begin{figure*}[t]
\centering
\includegraphics[width=.9\textwidth]{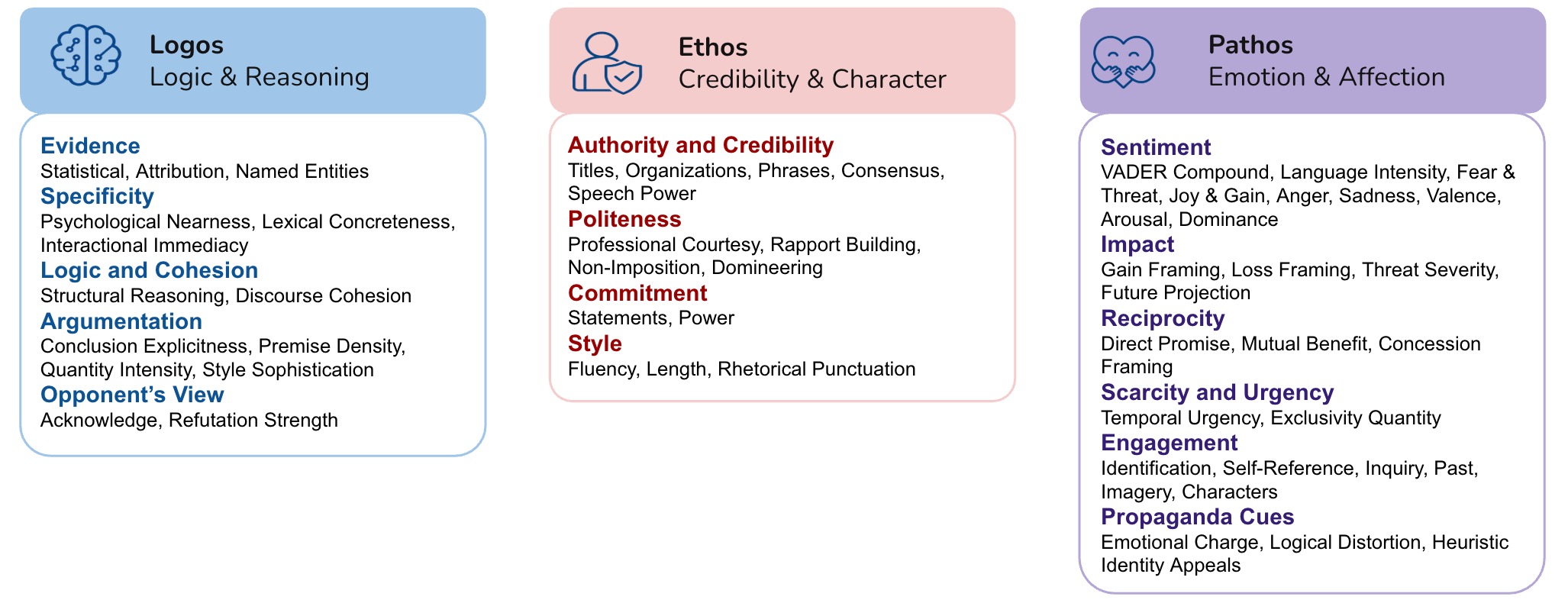}
  \caption{The Persuasion Index (PI) taxonomy. Fifteen dimensions are organized under the Aristotelian triad of \appeal{Logos}, \appeal{Ethos}, and \appeal{Pathos}, with sub-features listed under each dimension.}
  \label{fig:PI_features}
\end{figure*}

The PI maps its 15 dimensions onto the Aristotelian Triad of \appeal{Logos}, \appeal{Ethos}, and \appeal{Pathos} \citep{aristotleRhetoricTheoryCivic2006a}, a meta-framework for organizing the predictors of persuasion, also used in argument quality and rhetorical analysis \citep{carlileGiveMeMore2018, higginsEthosLogosPathos2012}. Under \appeal{Logos}, PI targets the central route of cognitive elaboration, including \pidim{Evidence}, \pidim{Specificity}, \pidim{Logic and Cohesion}, \pidim{Argumentation}, and \pidim{Opponent's View}. Under \appeal{Ethos}, PI captures heuristic credibility signals, including \pidim{Authority and Credibility}, \pidim{Politeness}, \pidim{Commitment}, and \pidim{Style}. Under \appeal{Pathos}, PI contains affective and social compliance triggers, including \pidim{Sentiment}, \pidim{Impact}, \pidim{Engagement}, \pidim{Reciprocity}, \pidim{Scarcity and Urgency}, and \pidim{Propaganda}. Figure~\ref{fig:PI_features} shows the 15 dimensions and 55 sub-features, with theoretical anchors and prior computational linguistic operationalizations mapped in Table~\ref{tab:pi_categories}.

\subsection{Logos: The Appeal to Reason}

\appeal{Logos} captures observable linguistic cues associated with factual grounding, explicit reasoning structure, and argumentative organization. These cues characterize how reasoning is presented in text but do not verify factual accuracy. Such cues may nevertheless facilitate central-route processing by making claims, evidence, and relations among propositions more explicit \citep{pettyElaborationLikelihoodModel1986}.

\mypar{\pidim{Evidence:}} linguistic cues that present claims as factually grounded, including statistics, source attribution, and named entities. These cues can increase the appearance or perception of evidential support, but PI does not determine whether a claim is true or whether an attributed source is reliable. Sub-features target cognitive beliefs by anchoring claims in facts and authoritative origins \citep{zebregsDifferentialImpactStatistical2015}. The inclusion of named entities is supported by Exemplification Theory: specific examples enhance perceived issue severity and message recall \citep{bigsbyExemplificationTheoryReview2019}.

\mypar{\pidim{Logic and Cohesion:}} explicit connective, inferential, and referential structures that make relations among statements easier to follow. High text cohesion facilitated by logical connectives reduces cognitive load, allowing readers to focus on content rather than processing structure \citep{kaakinenInfluenceTextCohesion2011}. This structural logic serves as a cognitive facilitator ensuring messages can be processed via systematic routes \citep{chaikenHeuristicSystematicInformation1980}.

\mypar{\pidim{Argumentation:}} structural explicitness of claims and the quantity and sophistication of supporting premises. Messages with explicitly articulated conclusions reduce audience misunderstanding and ensure the intended point is communicated \citep{okeefeStandpointExplicitnessPersuasive1997a}. For lower-involvement audiences, premise density additionally functions as a peripheral ``length-is-strength'' heuristic under ELM \citep{pettyEffectsInvolvementResponses1984}.

\mypar{\pidim{Specificity:}} linguistic abstraction and psychological distance of the argument's subject. Construal Level Theory argues that concrete, specific language is more persuasive for immediate targets as it facilitates perceptual vividness and increases perceived truth \citep{fujitaInfluencingAttitudesDistant2008, hansenTruthLanguageTruth2010, packardHowConcreteLanguage2021}. Interactional immediacy via second-person pronoun use further triggers self-referencing, deepening receiver involvement \citep{cruzSecondPersonPronouns2017}.

\mypar{\pidim{Opponent's View:}} acknowledgment and refutation of opposing positions. Refutational two-sided messages, which acknowledge then attack the opposing position, are more persuasive than one-sided messages because they preemptively resolve receiver objections \citep{okeefeHowHandleOpposing1999}. Non-refutational two-sided messages are consistently less effective than one-sided appeals \citep{allenMetaanalysisComparingPersuasiveness1991}.

\subsection{Ethos: The Appeal to Character}

Ethos captures credibility signals embedded in linguistic choices, functioning as heuristic shortcuts that shape receiver trust independently of argument content \citep{hovlandInfluenceSourceCredibility1951, chaikenHeuristicSystematicInformation1980}.

\mypar{\pidim{Authority and Credibility:}} expertise and institutional trust. High-credibility markers generally enhance persuasion, though their influence is dynamic and contingent on receiver involvement \citep{pornpitakpanPersuasivenessSourceCredibility2004}. The PI also considers trust-building markers to counteract the psychological reactance that excessive authority distance can trigger \citep{hovlandInfluenceSourceCredibility1951, songSourceEffectsPsychological2018}.

\mypar{\pidim{Politeness:}} how an argument mitigates imposition on the receiver, reducing psychological reactance. Successful communicators mitigate the ``Face Threatening Acts'' inherent in persuasive requests \citep{brownPolitenessUniversalsLanguage1987}. Meta-analyses of the ``But You Are Free'' technique demonstrate that acknowledging a listener's autonomy significantly increases compliance by reducing the perceived threat to freedom \citep{carpenterMetaAnalysisEffectivenessYou2013}.

\mypar{\pidim{Commitment:}} communicator's resolve and investment. The ``Foot-in-the-Door'' mechanism shows that expressed prior effort shifts a receiver's perception of the speaker's consistency \citep{burgerFootintheDoorComplianceProcedure1999}. Persuasive effects are stronger when there is verifiable proof of action \citep{pallakCommitmentEnergyConservation1980}.

\mypar{\pidim{Style:}} surface-level markers of communicative competence that influence the text's perceptual fluency. Individuals often mistake the ease of processing a text for factual accuracy, using fluency as a heuristic for truth \citep{reberEffectsPerceptualFluency1999}. Terminal punctuation and message length function as social cues for sincerity and emotional intensity in digital communication \citep{kempPeriodReallyPissed2025}.

\subsection{Pathos: The Appeal to Emotion}

Pathos captures affective and social compliance triggers, operating primarily through peripheral routes where emotional resonance influences judgment without requiring deep logical scrutiny \citep{forgasMoodJudgmentAffect1995, pettyElaborationLikelihoodModel1986}.

\mypar{\pidim{Sentiment (Emotion):}} the affective intensity of language. Discrete emotions produce distinct action tendencies that influence processing: fear promotes protection-seeking while anger promotes heuristic evaluation \citep{pettyMultipleRolesAffect1991, schwarzMoodMisattributionJudgments1983}. Meta-analyses on fear appeals suggest they are highly effective when threat is balanced with efficacy \citep{witteMetaanalysisFearAppeals2000}.

\mypar{\pidim{Impact:}} future consequences with projected gains or losses. The linguistic framing of costs and benefits significantly alters receivers' psychological cost-benefit analysis \citep{kahnemanProspectTheoryAnalysis1979, okeefeRelativePersuasivenessGainframed2007}. The persuasive force of these frames is further conditioned by projected outcome severity, as messages depicting high-severity consequences drive receivers toward danger-control processing \citep{witteMetaanalysisFearAppeals2000}.

\mypar{\pidim{Engagement:}} textual antecedents hypothesized to stimulate psychological immersion (i.e., transportation) or prompt audience self-referencing. While narrative transportation and identification represent subjective audience states rather than inherent message properties, they are systematically activated by distinct story-driven markers; audience transportation promotes persuasion by reducing the receiver's ability to counter-argue, bypassing logical scrutiny \citep{greenRoleTransportationPersuasiveness2000, slaterEntertainmentEducationElaboration2002}. Self-referencing effects further confirm that direct audience address integrates message content with the receiver's self-concept, making persuasive outcomes more persistent \citep{burnkrantEffectsSelfreferencingPersuasion1995}.

\mypar{\pidim{Reciprocity:}} concession or indebtedness prompting an obligation to comply. The ``Door-in-the-Face'' technique is driven by reciprocal concession mechanisms, where perceived social sacrifice triggers obligation in the receiver \citep{okeefeDoorintheFaceInfluenceStrategy1998}. The ``That's-Not-All'' technique suggests that pre-emptive deal-sweetening creates immediate indebtedness \citep{burgerIncreasingComplianceImproving1986}.

\mypar{\pidim{Scarcity and Urgency:}} limited time or availability. Scarcity heightens perceived value and acts as a competitive signal that escalates decision urgency \citep{ladeiraMetaanalysisEffectsProduct2023}. Per Psychological Reactance Theory: when freedom of choice is threatened by limited supply or deadlines, individuals are driven to pursue the restricted object more intensely \citep{brehmTheoryPsychologicalReactance1966, worchelEffectsSupplyDemand1975}.

\mypar{\pidim{Propaganda:}} identity-based heuristics and logical distortions that secure compliance regardless of logical depth. Even modest propaganda exposure produces significant shifts in pro-regime attitudes, though effects vary considerably across contexts \citep{shenDoesAuthoritarianPropaganda2025}. Propaganda further functions through deterrence, using markers of strength and danger to suppress counter-argumentation through fear \citep{karasmanPropagandaMechanismManipulation, jowettPropagandaPersuasion2012}.
\section{Methods}
\label{sec:methods}

We present a transparent implementation of PI using lexicon and rule-based sub-features. Crucially, the framework is implementation-agnostic: lexicons can be swapped for other models while preserving the theoretically-grounded PI structure. Section \ref{sec:experimental-setup} describes our empirical setup.

\subsection{Feature Construction}
\label{sec:methods-feature_construction}

We represent each text as a fixed-dimensional vector corresponding to the PI sub-features. Table \ref{tab:pi_methods_one_cue} contains specific operationalizations for all sub-features. We first leverage existing computational linguistic resources when available, including concreteness norms \citep{brysbaertConcretenessRatings402014} (\pidim{Specificity}), LIWC and NRC-VAD lexicons for \pidim{Sentiment} and \pidim{Engagement} \citep{boydDevelopmentPsychometricProperties,mohammadNRCVADLexicon2025}, and VADER for polarity scoring \citep{huttoVADERParsimoniousRuleBased2014}. We also use regular expressions to detect structural cues, e.g., numbers and URLs under \pidim{Evidence}.

For remaining sub-features, we first manually constructed seed lexicons based on foundational persuasion theories (Table~\ref{tab:pi_categories}). For example, the \pisub{Causal} sub-feature of the \pidim{Logic and Cohesion} dimension has a seed lexicon containing causal connectives (``because'', ``therefore'') drawn from argumentation theory \citep{toulminUsesArgument2003}. We then expand these seed lexicons with a series of structured prompts for \model{GPT-5.4-mini} that specify the sub-feature label, theoretical anchor, inclusion/exclusion criteria, and one of seven register- or morphology-conditioned slices (e.g., formal, colloquial, domain-specific, inflectional variants). Generated candidates are retained or filtered out using the validation procedure described in Section~\ref{sec:methods-validation}. Importantly, \model{GPT-5.4-mini} is used only for offline lexicon construction. The released PI scorer uses frozen lexicons and deterministic rules at runtime and makes no generative-model calls. Full prompts, expansion procedures, and finalized lexicons are provided in Appendix \ref{appendix:llm_expansion}.

We do not assume lexical detectors are inherently superior to pretrained models. The current implementation prioritizes inspectable mappings from textual cues to rhetorical constructs, while using pretrained components where appropriate, such as spaCy NER for evidential anchors. PI’s modular design allows richer semantic or discourse-aware detectors to replace these components without changing the taxonomy.

\subsection{Validation}
\label{sec:methods-validation}

We validate PI at multiple levels. First, during lexicon construction, we manually screen seed terms against their theoretical definitions, inspect feature distributions and top- and bottom-scoring texts, and filter LLM-expanded candidates through pilot human annotation, consensus coding, and confidence-tier review. See full procedures in Appendix~\ref{appendix:llm_expansion}.

Second, we assess lexicon stability using split-half consistency, independently rescoring the corpus with two non-overlapping halves of lexicons. 

Third, we conduct blinded dimension-level human validation by asking annotators to identify which of paired high- and low-scoring texts more strongly exhibits the target rhetorical construct.

Finally, we assess the robustness of coefficient interpretation on UKP using correlation and variance inflation factor (VIF) diagnostics, leave-one-dimension-out ablations, and leave-one-subfeature-out ablations. For each ablation, we remove the target feature set, refit the model, and measure the resulting change in predictive performance. Full procedures and results are reported in Appendix~\ref{appendix:robustness}.

\subsection{PI Score Generation}
\label{sec:methods-PI_score}

Each sub-feature is scored in [0,1] as either a saturating density $1 - e^{-k \cdot r}$ (where r is the per-100-token match rate and k=0.5 controls saturation speed) or a binary indicator for sparse signals. The value k=0.5 is chosen so that r=2 (two matches per 100 tokens) maps to a moderate score of $\approx 0.63$ and saturation approaches 1 for $r\geq 8$, calibrating the curve to the typical match density observed in our corpora. Overlapping lexicon matches are resolved by retaining only the longest non-contained span. Subfeatures whose theoretical direction is negative are reverse-coded before dimension aggregation. For example, hedges, hesitations, and tag questions are transformed so that greater powerless-speech use lowers the speech-power component of \pidim{Authority and Credibility}. Dimension scores are the mean of their sub-features, yielding a 15-dimensional category-mean vector or a full 55-dimensional sub-feature vector.

\subsection{Evaluation Datasets and Models}
\label{sec:experimental-setup}

We evaluate PI on four publicly released persuasion datasets (Appendix~\ref{appendix:datasets}): \textbf{UKPConvArg1 (UKP)} \citep{habernalWhichArgumentMore2016} (n=11,650, pairwise convincingness), \textbf{CMV} \citep{tanWinningArgumentsInteraction2016} (n=8,526, $\Delta$-based binary outcomes), \textbf{IBM Argument Quality} \citep{gretzLargescaleDatasetArgument2020} (n=14,003, expert quality ratings with pairwise comparisons), and \textbf{Anthropic Persuasion} \citep{MeasuringPersuasivenessLanguage} (n=3,882, argument-level annotations). All datasets are normalized into a unified binary persuasion label. For pairwise datasets (UKP and IBM), we construct difference vectors $\Delta\mathbf{x} = \mathbf{x}_A - \mathbf{x}_B$ from each argument pair's PI feature vectors, with label 1 if argument A is judged more persuasive. For CMV, each reply is labeled 1 if the original poster awarded a $\Delta$, and 0 otherwise. For Anthropic, we label 1 if post-exposure attitude improves by more than 1 point after removing participants with extreme initial attitudes due to ceiling and floor effects (Appendix~\ref{sec:persuasiveness-signal-construction}).

We compare $\ell_2$-regularized logistic regression on PI feature vectors (15-dimensional category means or full 55-dimensional sub-features) against fine-tuned \model{RoBERTa} \citep{liuRoBERTaRobustlyOptimized2019} and zero-shot \model{GPT-4o} as opaque high-capacity baselines. All models use a fixed 80/20 stratified split at the instance level (Appendix~\ref{appendix:model}).
\section{Results}
\label{sec:results}

\paragraph{Predictive Performance}


\begin{table}[t]
\centering
\small
\setlength{\tabcolsep}{4pt}
\caption{Persuasion prediction performance. Precision, recall, and F1 are computed for the positive persuasion class; accuracy is overall accuracy. For RoBERTa, values are means across 10 runs, with standard deviations reported in parentheses below. PI-sub uses 55 subfeatures and PI-mean uses 15 dimension means.}
\label{tab:f1_overview}
\begin{tabular}{llcccc}
\toprule
Dataset & Model & P & R & Acc & F1 \\
\midrule
\multirow{4}{*}{Anthropic}
 & GPT4o   & 0.326 & \textbf{0.984} & 0.328 & \textbf{0.490} \\
 & PI-mean & 0.350 & 0.553 & 0.516 & 0.429 \\
 & PI-sub  & \textbf{0.359} & 0.541 & 0.533 & 0.432 \\
 & RoBERTa &
   \makecell{0.357\\{\scriptsize (0.056)}} &
   \makecell{0.424\\{\scriptsize (0.303)}} &
   \makecell{\textbf{0.540}\\{\scriptsize (0.108)}} &
   \makecell{0.333\\{\scriptsize (0.126)}} \\
\midrule
\multirow{4}{*}{CMV}
 & GPT4o   & 0.563 & 0.436 & 0.528 & 0.492 \\
 & PI-mean & 0.582 & 0.561 & 0.560 & 0.571 \\
 & PI-sub  & \textbf{0.611} & 0.567 & \textbf{0.585} & \textbf{0.588} \\
 & RoBERTa &
   \makecell{0.528\\{\scriptsize (0.018)}} &
   \makecell{\textbf{0.590}\\{\scriptsize (0.256)}} &
   \makecell{0.510\\{\scriptsize (0.022)}} &
   \makecell{0.528\\{\scriptsize (0.148)}} \\
\midrule
\multirow{4}{*}{IBM}
 & GPT4o   & \textbf{0.702} & 0.503 & \textbf{0.650} & 0.586 \\
 & PI-mean & 0.565 & 0.594 & 0.575 & 0.579 \\
 & PI-sub  & 0.581 & \textbf{0.599} & 0.590 & \textbf{0.590} \\
 & RoBERTa &
   \makecell{0.496\\{\scriptsize (0.187)}} &
   \makecell{0.441\\{\scriptsize (0.261)}} &
   \makecell{0.543\\{\scriptsize (0.061)}} &
   \makecell{0.440\\{\scriptsize (0.213)}} \\
\midrule
\multirow{4}{*}{UKP}
 & GPT4o   & \textbf{0.895} & \textbf{0.794} & \textbf{0.855} & \textbf{0.842} \\
 & PI-mean & 0.677 & 0.691 & 0.690 & 0.684 \\
 & PI-sub  & 0.768 & 0.768 & 0.775 & 0.768 \\
 & RoBERTa &
   \makecell{0.682\\{\scriptsize (0.141)}} &
   \makecell{0.715\\{\scriptsize (0.191)}} &
   \makecell{0.697\\{\scriptsize (0.139)}} &
   \makecell{0.690\\{\scriptsize (0.156)}} \\
\bottomrule
\end{tabular}
\end{table}

Our straightforward implementation of PI carries predictive signal for persuasion, with performance comparable to larger black-box models. \model{PI-sub} consistently outperforms \model{PI-mean} across all datasets (Table~\ref{tab:f1_overview}), confirming that preserving sub-feature granularity aids persuasion modeling. 

\model{PI-sub} performs competitively with our black-box LLM baselines. On UKP, \model{PI-sub} reaches parity with the original benchmark (0.76--0.78) \citep{habernalWhichArgumentMore2016}, outperforming \model{RoBERTa} and trailing zero-shot \model{GPT-4o}. On CMV and IBM, \model{PI-sub} achieves the highest F1 among all models considered. \model{RoBERTa} exhibits sensitivity to stochastic fine-tuning. Across 10 random seeds, the mean F1 is below PI-sub on all four datasets, although the large standard deviations show that individual runs can vary considerably. Overall, \model{PI-sub} and \model{GPT-4o} each lead on two datasets, but \model{PI-sub} attains this with a transparent and computationally lightweight feature pipeline. 

We statistically compare models via McNemar's test, bootstrap difference tests, and equivalence testing (TOST). \model{PI-sub} significantly outperforms \model{RoBERTa} (with seed 42) on UKP, CMV, and IBM, and significantly outperforms \model{GPT-4o} on CMV and Anthropic, while \model{GPT-4o} significantly outperforms \model{PI-sub} on UKP. See full results in Appendix~\ref{appendix:results}.

These results show that PI carries predictive signal for persuasion-related outcomes across four datasets that vary substantially in domain, style, and outcome measures. However, we caution that these are simple LLM baselines: \model{RoBERTa} is fine-tuned with 10 random seeds but no hyperparameter search, and \model{GPT-4o} is evaluated zero-shot with a minimal prompt. Rather than claim state-of-the-art performance, we treat these comparisons as evidence of predictive validity. PI provides a transparent feature space that is not only interpretable for descriptive analysis, but also predictive enough to support meaningful comparisons of rhetorical patterns across domains.

\paragraph{Coefficients by Dataset}
\label{result-dataset}

\begin{figure}[t]
  \includegraphics[width=\columnwidth]{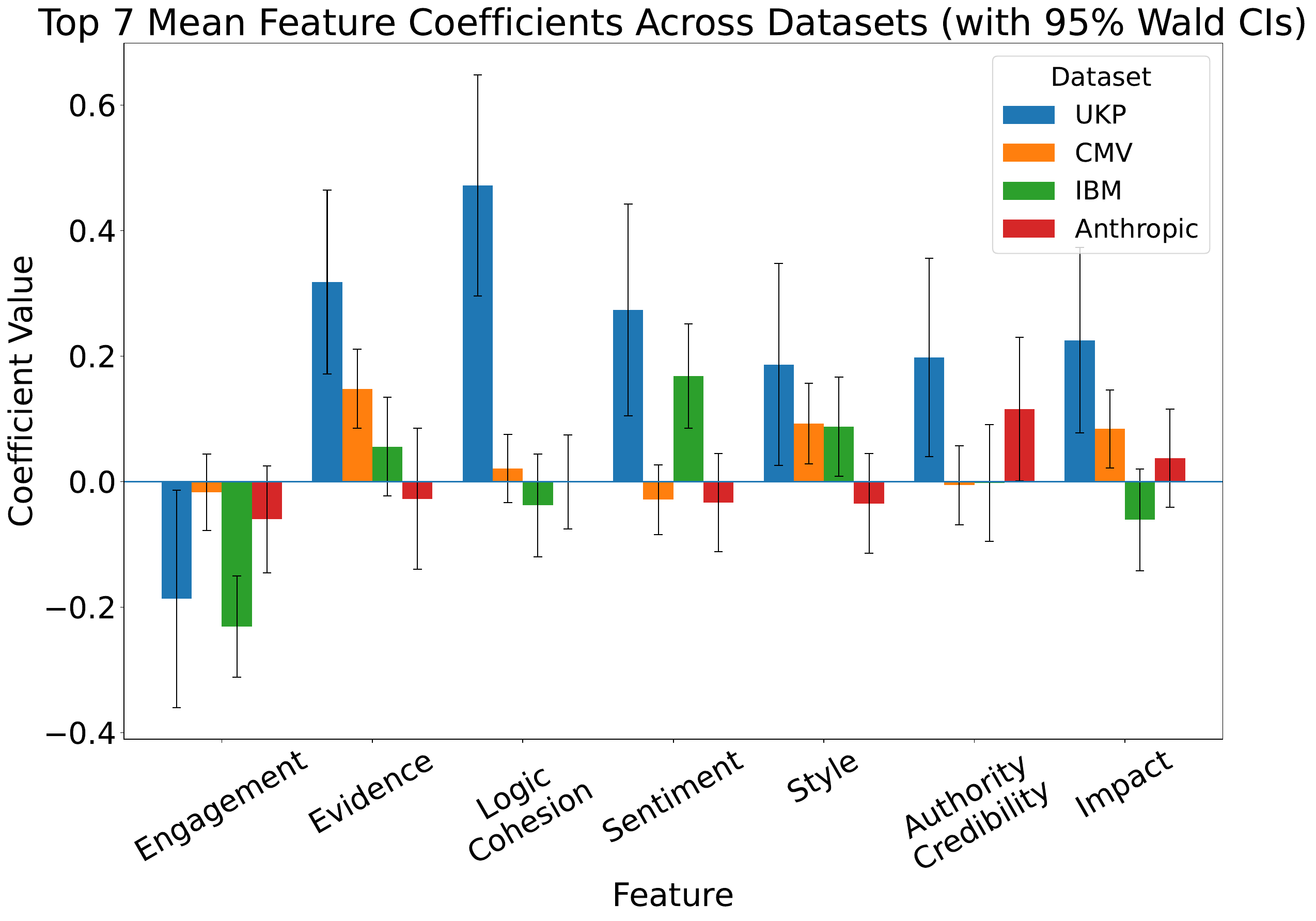}
\caption{Top 7 \model{PI-mean} dimensions by mean absolute coefficient across datasets (95\% Wald CI).}
  \label{fig:mean_coefficient_8_comparison}
\end{figure}

Figure~\ref{fig:mean_coefficient_8_comparison} shows 7 PI dimensions with the largest mean coefficient magnitudes across datasets (coefficient robustness is tested via VIF and ablation analyses; Appendix~\ref{appendix:robustness}). UKP exhibits the largest and most consistently positive coefficients overall, led by \pidim{Logic and Cohesion}, \pidim{Evidence}, and \pidim{Sentiment}, patterns consistent with the structured, evidence-oriented nature of its crowdsourced pairwise convincingness task. \pidim{Logic and Cohesion} is highly domain-specific: dominant on UKP but statistically indistinguishable from zero on CMV, IBM, and Anthropic. \pidim{Evidence} generalizes more broadly, retaining a clearly positive effect on CMV but attenuating on IBM and Anthropic. Each remaining dataset contributes one distinctive signal of its own, \pidim{Sentiment} on IBM and \pidim{Authority and Credibility} on Anthropic, while their other coefficients are small with confidence intervals that frequently span zero. \pidim{Engagement} has negative coefficients across all datasets, suggesting that personal pronoun density and narrative immersion do not reliably distinguish more persuasive arguments in these corpora.

\paragraph{Coefficients by Topic}
\label{result-topic}

\begin{figure}[t]
  \includegraphics[width=\columnwidth]{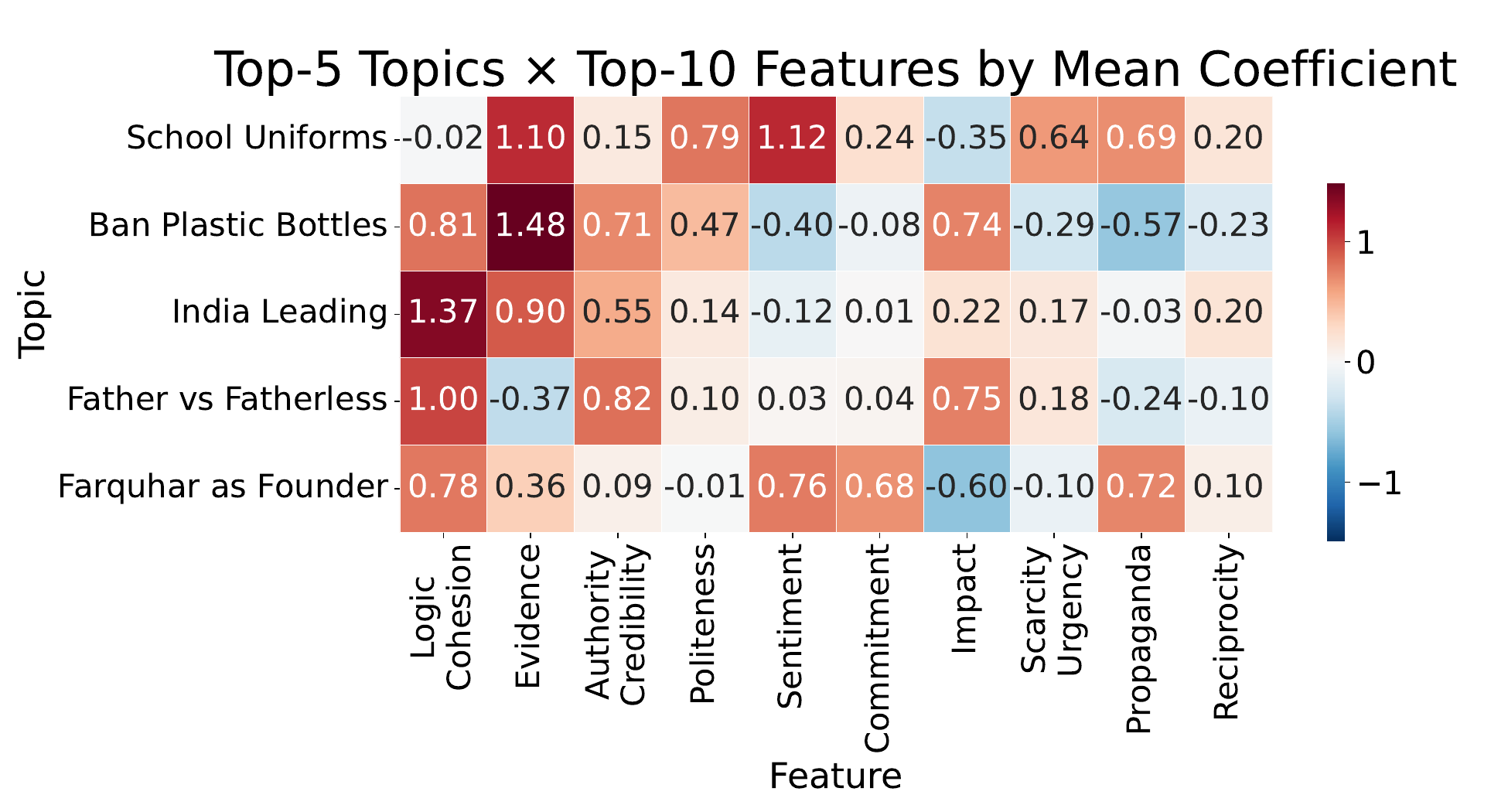}
\caption{Top 10 \model{PI-mean} dimensions by mean absolute coefficient for the 5 highest-accuracy UKP topics. Cells show standardized coefficients.}
  \label{fig:ukp_coefficient_topic_subset}
\end{figure}

Figure~\ref{fig:ukp_coefficient_topic_subset} shows that persuasive strategies shift substantially across debate topics. \pidim{Evidence} is the most consistent predictor, peaking in \textit{Ban Plastic Bottles} (1.48), where \pidim{Logic and Cohesion} (0.81) and \pidim{Authority and Credibility} (0.71) also rank highly, consistent with policy topics rewarding factual, structured reasoning. \textit{India Leading} shows a similar Logos-dominant profile, with \pidim{Logic and Cohesion} reaching its highest value (1.37) alongside strong \pidim{Evidence} (0.90). Interpersonal topics shift toward affective and relational dimensions: \textit{Father vs.\ Fatherless} is driven by \pidim{Logic and Cohesion} (1.00) and \pidim{Impact} (0.75), while \textit{Farquhar as Founder} relies on \pidim{Sentiment} (0.76) and \pidim{Commitment} (0.68) with a notably negative \pidim{Impact} ($-$0.60). \textit{School Uniforms} presents a distinct mix of \pidim{Evidence} (1.10), \pidim{Sentiment} (1.12), and \pidim{Propaganda} (0.69), reflecting a topic where normative pressure and emotional appeals dominate over logical structure. These patterns confirm that while \pidim{Evidence} and \pidim{Logic and Cohesion} provide stable cross-topic signals, the relative importance of credibility-based, affective, and persuasion-pressure dimensions varies substantially with topic framing. Full results across all 16 UKP topics and all features, as well as IBM topic-level results, are provided in Appendix~\ref{appendix:results-ibm-topic}.

\paragraph{Coefficients by Stance}
\label{sec:result-stance}

\begin{figure}[t]
  \includegraphics[width=\columnwidth]{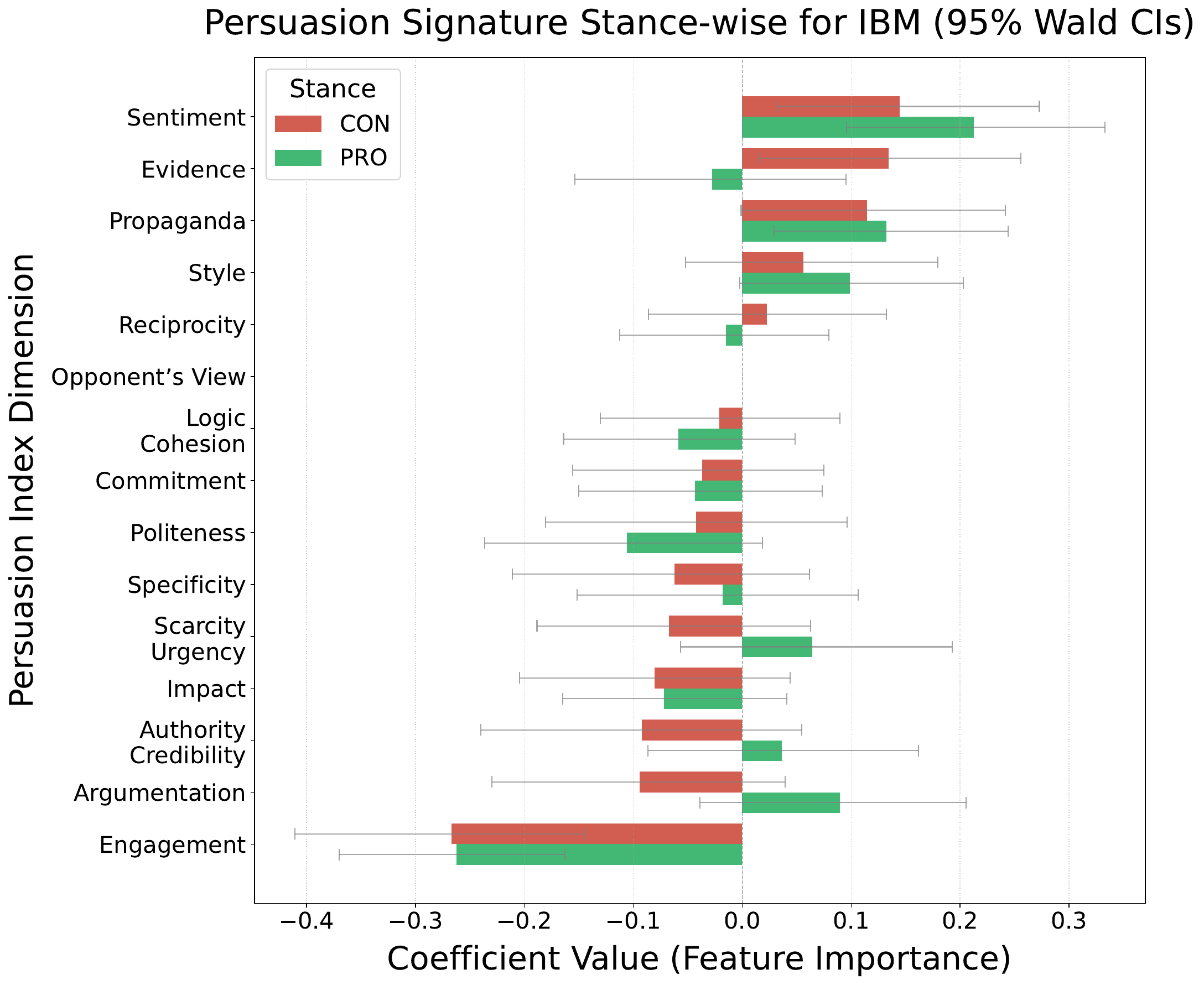}
\caption{Coefficients for all 15 PI dimensions by stance in the IBM dataset (95\% Wald CI). PRO (CON) arguments advocate (oppose) the stated premise.}
  \label{fig:ibm_stance_features_comparison}
\end{figure}

Each topic in the IBM dataset has arguments on both sides: PRO advocates for a stated premise (e.g., \textit{We should adopt vegetarianism}) while CON opposes it (e.g., \textit{We should abandon vegetarianism}) (Appendix~\ref{appendix:datasets}). Models are fitted separately for each stance. Persuasion is stance-contingent in this setting (Figure~\ref{fig:ibm_stance_features_comparison}). Both stances share \pidim{Sentiment} as the strongest positive predictor and \pidim{Engagement} as the strongest negative predictor. PRO arguments also weight \pidim{Argumentation}, \pidim{Authority and Credibility}, and \pidim{Scarcity and Urgency} positively, suggesting that supporting a claim benefits from explicit reasoning, credibility appeals, and decision pressure. CON arguments weight \pidim{Evidence} and \pidim{Reciprocity} more heavily, consistent with a strategy of factual challenge and concession framing when opposing a premise.

\paragraph{Validation and Robustness.}
PI scores generally align with the intended rhetorical constructs: blinded human judgments agree with PI's high--low ordering in 86.2\% of 289 valid comparisons, with agreement of at least 75\% for 12 of 15 dimensions. Split-half rescoring also shows high aggregate consistency ($r=.90$ across lexicon-driven sub-features; $r=.97$ across dimension means). Robustness analyses further show that 12 of 15 dimensions have VIF $<10$, while \pidim{Style}, \pidim{Sentiment}, and \pidim{Logic and Cohesion} exhibit greater multicollinearity. Leave-one-dimension-out ablations identify \pidim{Style}, \pidim{Sentiment}, and \pidim{Engagement} as the largest marginal contributors on UKP. At the sub-feature level, much of \pidim{Style}'s contribution is concentrated in length, whereas \pidim{Sentiment}'s contribution is distributed across several sub-features. Full validation and robustness results are reported in Appendices~\ref{appendix:llm_expansion} and~\ref{appendix:robustness}.
\section{Discussion}
Persuasion research faces a persistent tension between predictive performance and interpretability. Neural and LLM-based models achieve strong results but treat persuasion as a scalar outcome, offering no insight into which rhetorical mechanisms drive success. PI addresses this gap by grounding persuasion measurement in an explicit, theory-derived feature space, and implemented through transparent lexicons and rule-based detectors, where every score traces back to specific lexical or structural patterns in the text.

Empirical validation on four datasets indicates that PI features can capture meaningful predictive signal for persuasion-related outcomes. \model{PI-sub} performs competitively, and sometimes outperforms, opaque RoBERTa and GPT-4o models. Unlike post-hoc methods that approximate feature importance after an opaque model has already predicted, PI is interpretable by construction: the features used by the predictive model are the same features visible to the researcher, and learned coefficients provide insights into associations between rhetorical dimensions and persuasion outcomes.

The most substantively important finding is not that PI predicts persuasion, but that comparative analyses across datasets, topics, and stances reveal regularities and variation in what kind of language tends to be persuasive. \pidim{Evidence}, \pidim{Sentiment}, and \pidim{Logic and Cohesion} are broadly positive predictors, but their relative weights vary across settings. UKP places more weight on logical structure and factual grounding; CMV shows more variable patterns, consistent with informal debate; Anthropic assigns higher weights to authority and credibility. \pidim{Evidence} and \pidim{Argumentation} are linked to persuasion for policy topics, while \pidim{Sentiment} and \pidim{Impact} are stronger predictors for interpersonal topics. At the stance level, PRO arguments (supporting a given claim) have higher weights for \pidim{Argumentation} and \pidim{Authority and Credibility}, while CON arguments (opposing a given claim) rely on \pidim{Evidence} and \pidim{Reciprocity}. 

The consistently negative coefficient for \pidim{Engagement} is notable. In these corpora, high personal pronoun density, direct address, and rhetorical questions do not reliably distinguish more persuasive arguments. However, this does not necessarily imply that narrative or engagement-oriented strategies are ineffective; rather, the surface linguistic markers we can extract from text may be weak proxies for audience-side psychological processes, such as transportation, identification, or involvement. The effectiveness of engagement-oriented language likely depends on factors beyond the text itself, including audience prior attitudes, medium, and context.

These patterns demonstrate that persuasion is better treated as a context-sensitive rhetorical configuration than a single generalizable signal. From an ELM perspective, structured debate audiences may engage the central route more systematically, rewarding logical coherence, while informal settings elevate peripheral cues. PI makes these shifts quantifiable in an interpretable feature space and helps move NLP and computational social science persuasion research toward the kind of theoretical explanation that has long been central to psychology and communication scholarship.

PI opens several practical directions. It facilitates content analysis across domains. For social scientists who study persuasive message effects, it provides a principled framework for constructing stimuli and conducting efficient, low-cost, and reliable manipulation checks without requiring human subjects. By verifying that only the targeted rhetorical dimensions vary across experimental conditions while others are held strictly constant, it helps researchers eliminate confounding variables, a level of precision difficult to achieve with manual text validation or with opaque neural systems. For researchers studying human--LLM or multi-agent persuasive dynamics, PI offers a shared feature space applicable to both human- and model-generated text, enabling direct rhetorical comparison across sources. As LLMs become increasingly capable persuaders \citep{hackenburgLeversPoliticalPersuasion2025, breumPersuasivePowerLarge2024}, characterizing which rhetorical mechanisms they deploy and how these compare to human norms becomes critical for accountability. 

PI is also designed to grow. Because the taxonomy and the implementation are decoupled, any individual dimension or sub-feature detector can be updated. For example, lexicons can be replaced with fine-tuned classifiers or discourse-aware models without disturbing the 15-dimension structure or its theoretical anchors. This separation distinguishes two questions that prior work tends to conflate: what rhetorical dimensions matter for persuasion (the taxonomy) and how best to detect each one in text (the implementation). PI offers a stable answer to the first while inviting community contributions to the second. To support this, we release PI as an open-source Python package and a public web interface for real-time argument analysis (Appendix~\ref{appendix:web_interface}). We envision PI as a community-driven resource where researchers test the framework on new domains, swap in stronger detectors for individual sub-features, and extend the feature pipeline, progressively expanding its theoretical coverage and cross-domain generalizability.

The interpretability--performance tradeoff in persuasion modeling is smaller than often assumed. What PI sacrifices in raw predictive power, it recovers in theoretical legibility, domain portability, and practical utility, offering a foundation for persuasion analysis that is competitive empirically, auditable by design, and directly connected to theoretical constructs that make findings meaningful.
\section{Limitations}

A fundamental limitation of PI, and of text-based persuasion modeling more broadly, is that persuasiveness is not fully determined by message content alone. Relational context, source reputation, audience prior attitudes, and social dynamics all shape persuasive outcomes but are not recoverable from text. PI operationalizes persuasion as a property of the message as a linguistic artifact, capturing the rhetorical choices a communicator encodes in text but not how those choices interact with receiver characteristics or interactional context.

PI's lexicon- and rule-based implementations prioritize transparency but are limited to surface-level operationalization. Strategies that depend on pragmatic interpretation, such as implicit framing, irony, and sarcasm, as well as discourse-level phenomena such as long-range narrative structure and cross-sentence argumentation, remain underrepresented. PI's modular design partially mitigates this: individual sub-feature implementations can be replaced or augmented with richer methods, including neural or LLM-based detectors, without altering the taxonomic structure or its theoretical grounding.

Several dimensions show confidence intervals overlapping zero in some datasets, reflecting both genuine cross-context variation in persuasive norms and the noisiness of sparse lexicon coverage. More data would yield more precise coefficient estimates, and continued lexicon refinement remains necessary. Per-feature internal consistency results are reported in Table~\ref{tab:internal_consistency} as a calibrated disclosure of which dimensions provide robust signals and which require further development.

Finally, PI is evaluated exclusively on English-language argumentative text from online debate, crowdfunding, and controlled experiment settings. Generalization to other languages, cultural rhetorical norms, spoken discourse, or multimodal contexts remains an open question. Lexicons constructed from English argumentative conventions may systematically fail to capture persuasive strategies prevalent in other linguistic and cultural traditions. Extending PI to multilingual and multimodal settings is an important direction for future work.

\section{Ethics Statement}
The Persuasion Index is designed as a transparent analytical tool for studying persuasive language. However, a framework that identifies effective persuasion strategies could in principle be misused to engineer more persuasive misinformation, manipulative political messaging, or targeted influence campaigns. We emphasize that PI is intended for analysis and auditing of persuasive content rather than its generation. Its transparency is itself a safeguard: by making persuasion mechanisms explicit and traceable, PI enables detection and critique of manipulative rhetoric as much as it describes it.
Lexicon-based features calibrated on English argumentative text may not generalize equally across languages, cultural communities, or demographic groups, as rhetorical norms vary substantially across contexts. We encourage users to audit PI scores for systematic disparities before deploying the framework in applied settings.
All datasets used are publicly released and contain no personally identifiable information. Lexicon annotation involved only word-level judgments with no collection of personal data.

\bibliography{references}

@article{smithEffectsPowerfulPowerless1998,
	title = {The {Effects} of {Powerful} and {Powerless} {Speech} {Styles} and {Speaker} {Expertise} {On} {Impression} {Formation} and {Attitude} {Change}},
	volume = {15},
	url = {https://aquila.usm.edu/fac_pubs/16286},
	number = {1},
	journal = {Communication Research Reports},
	author = {Smith, Victoria and Siltanen, Susan and Hosman, Lawrence},
	month = jan,
	year = {1998},
	pages = {27--35},
}

@book{brehmTheoryPsychologicalReactance1966,
	address = {Oxford, England},
	title = {A theory of psychological reactance},
	publisher = {Academic Press},
	author = {Brehm, Jack W.},
	year = {1966},
}

@article{greenRoleTransportationPersuasiveness2000,
	title = {The role of transportation in the persuasiveness of public narratives},
	volume = {79},
	issn = {1939-1315},
	doi = {10.1037/0022-3514.79.5.701},
	number = {5},
	journal = {Journal of Personality and Social Psychology},
	publisher = {American Psychological Association},
	author = {Green, Melanie C. and Brock, Timothy C.},
	year = {2000},
	pages = {701--721},
}

@book{SAGEHandbookPersuasion2025,
	address = {Thousand Oaks, CA},
	edition = {2},
	title = {The {SAGE} {Handbook} of {Persuasion}: {Developments} in {Theory} and {Practice}},
	doi = {10.4135/9781452218410},
	language = {en},
	urldate = {2025-09-30},
	publisher = {SAGE Publications, Inc.},
	editor = {Dillard, James Price and Shen, Lijiang},
	year = {2012},
}

@article{grantMultipleRolesScarcity2014,
	title = {The multiple roles of scarcity in compliance: {Elaboration} as a moderator of scarcity mechanisms},
	volume = {9},
	issn = {1553-4510},
	shorttitle = {The multiple roles of scarcity in compliance},
	url = {https://doi.org/10.1080/15534510.2013.796891},
	doi = {10.1080/15534510.2013.796891},
	number = {2},
	urldate = {2026-04-10},
	journal = {Social Influence},
	publisher = {Routledge},
	author = {Grant, Naomi K. and Fabrigar, Leandre R. and Forzley, Adelle and Kredentser, Maia},
	month = apr,
	year = {2014},
	pages = {149--161},
}

@article{mccroskeyEthosCredibilityConstruct1981,
	title = {Ethos and credibility: {The} construct and its measurement after three decades},
	volume = {32},
	issn = {0008-9575},
	shorttitle = {Ethos and credibility},
	url = {https://doi.org/10.1080/10510978109368075},
	doi = {10.1080/10510978109368075},
	number = {1},
	urldate = {2026-04-16},
	journal = {Central States Speech Journal},
	publisher = {Routledge},
	author = {McCroskey, James C. and Young, Thomas J.},
	month = mar,
	year = {1981},
	pages = {24--34},
}

@article{mccroskeyGoodwillReexaminationConstruct1999,
	title = {Goodwill: {A} reexamination of the construct and its measurement},
	volume = {66},
	issn = {0363-7751},
	shorttitle = {Goodwill},
	url = {https://doi.org/10.1080/03637759909376464},
	doi = {10.1080/03637759909376464},
	number = {1},
	urldate = {2026-04-16},
	journal = {Communication Monographs},
	publisher = {Routledge},
	author = {McCroskey, James C. and Teven, Jason J.},
	month = mar,
	year = {1999},
	pages = {90--103},
}

@article{hornikQuantitativeEvaluationPersuasive2016,
	title = {Quantitative evaluation of persuasive appeals using comparative meta-analysis},
	volume = {19},
	issn = {1071-4421},
	url = {https://doi.org/10.1080/10714421.2016.1195204},
	doi = {10.1080/10714421.2016.1195204},
	number = {3},
	urldate = {2026-03-29},
	journal = {The Communication Review},
	publisher = {Routledge},
	author = {Hornik, Jacob and Ofir, Chezy and Rachamim, Matti},
	month = jul,
	year = {2016},
	pages = {192--222},
}

@incollection{dillardLanguageStylePersuasion2014,
	address = {New York, NY},
	title = {Language, style, and persuasion},
	booktitle = {The {Oxford} {Handbook} of {Language} and {Social} {Psychology}},
	publisher = {Oxford University Press},
	author = {Dillard, James P.},
	editor = {Holtgraves, Thomas M.},
	month = jan,
	year = {2014},
	pages = {177--187},
}

@incollection{pallakCommitmentEnergyConservation1980,
	address = {US},
	title = {Commitment and energy conservation},
	volume = {1},
	issn = {0196-4151},
	booktitle = {Applied {Social} {Psychology} {Annual}},
	publisher = {Sage Publications},
	author = {Pallak, Michael S. and Cook, David A. and Sullivan, John J.},
	editor = {Bickman, Leonard},
	year = {1980},
	pages = {235--253},
}

@article{hansenTruthLanguageTruth2010,
	title = {Truth from language and truth from fit: {The} impact of linguistic concreteness and level of construal on subjective truth},
	volume = {36(11)},
	shorttitle = {Truth {From} {Language} and {Truth} {From} {Fit}},
	doi = {10.1177/0146167210386238},
	journal = {Personality \& Social Psychology Bulletin},
	author = {Hansen, Jochim and Wänke, Michaela},
	month = nov,
	year = {2010},
	pages = {1576--88},
}

@book{okeefePersuasionTheoryResearch2015,
	address = {Thousand Oaks, CA},
	edition = {3},
	title = {Persuasion: {Theory} and {Research}},
	isbn = {978-1-4833-0971-2},
	shorttitle = {Persuasion},
	language = {en},
	publisher = {SAGE Publications},
	author = {O'Keefe, Daniel J.},
	year = {2015},
}

@book{brownPolitenessUniversalsLanguage1987,
	title = {Politeness: {Some} universals in language usage},
	isbn = {978-0-521-31355-1},
	shorttitle = {Politeness},
	url = {https://www.cambridge.org/highereducation/books/politeness/89113EE2FB4A1D254D4A8D2011E542E4},
	doi = {10.1017/CBO9780511813085},
	language = {en},
	urldate = {2025-12-08},
	publisher = {Cambridge University Press},
	author = {Brown, Penelope and Levinson, Stephen C.},
	year = {1987},
}

@misc{shenDoesAuthoritarianPropaganda2025,
	address = {Rochester, NY},
	type = {{SSRN} {Scholarly} {Paper}},
	title = {Does authoritarian propaganda persuade and deter? {A} meta-analysis of 70 {Chinese} propaganda experiments},
	shorttitle = {Does {Authoritarian} {Propaganda} {Persuade} and {Deter}?},
	url = {https://papers.ssrn.com/abstract=5895103},
	doi = {10.2139/ssrn.5895103},
	language = {en},
	urldate = {2026-04-11},
	publisher = {Social Science Research Network},
	author = {Shen, Shuyuan and Yin, Tianhong},
	year = {2025},
}

@misc{MeasuringPersuasivenessLanguage,
	title = {Measuring the {Persuasiveness} of {Language} {Models}},
	url = {https://www.anthropic.com/research/measuring-model-persuasiveness},
	language = {en},
	urldate = {2026-03-26},
	journal = {Anthropic},
	publisher = {Anthropic},
	author = {Durmus, Esin and Lovitt, Liane and Tamkin, Alex and Ritchie, Stuart and Clark, Jack and Ganguli, Deep},
	year = {2024},
}

@inproceedings{dimitrovSemEval2021Task62021,
	title = {{SemEval}-2021 {Task} 6: {Detection} of {Persuasion} {Techniques} in {Texts} and {Images}},
	shorttitle = {{SemEval}-2021 {Task} 6},
	url = {https://aclanthology.org/2021.semeval-1.7/},
	doi = {10.18653/v1/2021.semeval-1.7},
	urldate = {2025-10-27},
	booktitle = {Proceedings of the 15th {International} {Workshop} on {Semantic} {Evaluation} ({SemEval}-2021)},
	publisher = {Association for Computational Linguistics},
	author = {Dimitrov, Dimitar and Bin Ali, Bishr and Shaar, Shaden and Alam, Firoj and Silvestri, Fabrizio and Firooz, Hamed and Nakov, Preslav and Da San Martino, Giovanni},
	year = {2021},
	pages = {70--98},
}

@inproceedings{hoangHybridTheoryDatadriven2025,
	title = {A {Hybrid} {Theory} and {Data}-driven {Approach} to {Persuasion} {Detection} with {Large} {Language} {Models}},
	url = {https://workshop-proceedings.icwsm.org/abstract.php?id=2025_38},
	doi = {10.36190/2025.38},
	urldate = {2026-04-29},
	booktitle = {Proceedings of the {International} {AAAI} {Conference} on {Web} and {Social} {Media} {Workshops}},
	publisher = {AAAI Press},
	author = {Hoang, Gia Bao and Ransom, Keith J. and Stephens, Rachel and Semmler, Carolyn and Fay, Nicolas and Mitchell, Lewis},
	year = {2025},
}

@inproceedings{chengStrategicPersuasionLanguage2026,
	title = {Towards {Strategic} {Persuasion} with {Language} {Models}},
	url = {https://openreview.net/forum?id=aTCXvJKnkE},
	urldate = {2025-10-05},
	booktitle = {Proceedings of the {International} {Conference} on {Learning} {Representations}},
	author = {Cheng, Zirui and You, Jiaxuan},
	year = {2026},
	note = {arXiv:2509.22989 [cs]},
}

@article{bozdagMustReadComprehensive2026,
	title = {Must {Read}: {A} {Comprehensive} {Survey} of {Computational} {Persuasion}},
	volume = {58},
	shorttitle = {Must {Read}},
	url = {https://dl.acm.org/doi/full/10.1145/3800687},
	doi = {10.1145/3800687},
	number = {12},
	urldate = {2026-05-15},
	journal = {ACM Computing Surveys},
	publisher = {Association for Computing Machinery},
	author = {Bozdag, Nimet Beyza and Mehri, Shuhaib and Yang, Xiaocheng and Ha, Hyeonjeong and Cheng, Zirui and Durmus, Esin and You, Jiaxuan and Ji, Heng and Tur, Gokhan and Hakkani-Tür, Dilek},
	month = may,
	year = {2026},
	pages = {1--39},
}

@inproceedings{stabAnnotatingArgumentComponents2014,
	address = {Dublin, Ireland},
	title = {Annotating {Argument} {Components} and {Relations} in {Persuasive} {Essays}},
	url = {https://aclanthology.org/C14-1142/},
	urldate = {2026-05-22},
	booktitle = {Proceedings of {COLING} 2014, the 25th {International} {Conference} on {Computational} {Linguistics}: {Technical} {Papers}},
	publisher = {Dublin City University and Association for Computational Linguistics},
	author = {Stab, Christian and Gurevych, Iryna},
	editor = {Tsujii, Junichi and Hajic, Jan},
	month = aug,
	year = {2014},
	pages = {1501--1510},
}

@inproceedings{dasanmartinoFineGrainedAnalysisPropaganda2019,
	address = {Hong Kong, China},
	title = {Fine-{Grained} {Analysis} of {Propaganda} in {News} {Articles}},
	url = {https://aclanthology.org/D19-1565/},
	doi = {10.18653/v1/D19-1565},
	urldate = {2026-05-22},
	booktitle = {Proceedings of the 2019 {Conference} on {Empirical} {Methods} in {Natural} {Language} {Processing} and the 9th {International} {Joint} {Conference} on {Natural} {Language} {Processing} ({EMNLP}-{IJCNLP})},
	publisher = {Association for Computational Linguistics},
	author = {Da San Martino, Giovanni and Yu, Seunghak and Barrón-Cedeño, Alberto and Petrov, Rostislav and Nakov, Preslav},
	editor = {Inui, Kentaro and Jiang, Jing and Ng, Vincent and Wan, Xiaojun},
	month = nov,
	year = {2019},
	pages = {5636--5646},
}

@book{aristotleRhetoricTheoryCivic2006a,
	address = {New York},
	edition = {2nd ed.},
	title = {On rhetoric: {A} theory of civic discourse},
	isbn = {978-0-19-530509-8},
	publisher = {Oxford University Press},
	author = {Aristotle},
	translator = {Kennedy, George A.},
	year = {2007},
}

@incollection{sharmaGenerativeEchoChamber2024,
	series = {{ACM} {Conferences}},
	title = {Generative {Echo} {Chamber}? {Effect} of {LLM}-{Powered} {Search} {Systems} on {Diverse} {Information} {Seeking}},
	isbn = {979-8-4007-0330-0},
	shorttitle = {Generative {Echo} {Chamber}?},
	url = {https://dl.acm.org/doi/10.1145/3613904.3642459},
	doi = {10.1145/3613904.3642459},
	urldate = {2026-05-20},
	booktitle = {Proceedings of the 2024 {CHI} {Conference} on {Human} {Factors} in {Computing} {Systems}},
	author = {Sharma, Nikhil and Liao, Q. Vera and Xiao, Ziang},
	month = may,
	year = {2024},
	pages = {1--17},
}

@inproceedings{sudharsanCrossDomainPersuasionDetection2025,
	address = {Suzhou, China},
	title = {Cross-{Domain} {Persuasion} {Detection} with {Argumentative} {Features}},
	isbn = {979-8-89176-340-1},
	url = {https://aclanthology.org/2025.starsem-1.30/},
	doi = {10.18653/v1/2025.starsem-1.30},
	urldate = {2026-05-15},
	booktitle = {Proceedings of the 14th {Joint} {Conference} on {Lexical} and {Computational} {Semantics} (*{SEM} 2025)},
	publisher = {Association for Computational Linguistics},
	author = {Sudharsan, Bagyasree and Pacheco, Maria Leonor},
	editor = {Frermann, Lea and Stevenson, Mark},
	month = nov,
	year = {2025},
	pages = {372--380},
}

@inproceedings{sajwaniFRAPPEFRAmingPersuasion2024,
	address = {St. Julians, Malta},
	title = {{FRAPPE}: {FRAming}, {Persuasion}, and {Propaganda} {Explorer}},
	shorttitle = {{FRAPPE}},
	url = {https://aclanthology.org/2024.eacl-demo.22/},
	doi = {10.18653/v1/2024.eacl-demo.22},
	urldate = {2026-05-15},
	booktitle = {Proceedings of the 18th {Conference} of the {European} {Chapter} of the {Association} for {Computational} {Linguistics}: {System} {Demonstrations}},
	publisher = {Association for Computational Linguistics},
	author = {Sajwani, Ahmed and El Setohy, Alaa and Mekky, Ali and Turmakhan, Diana and Hassan, Lara and El Zeftawy, Mohamed and El Herraoui, Omar and Mohammed Afzal, Osama and Liao, Qisheng and Mahmoud, Tarek and Muhammad Mujahid, Zain and Salman, Muhammad Umar and Arslan Manzoor, Muhammad and Baali, Massa and Piskorski, Jakub and Stefanovitch, Nicolas and Da San Martino, Giovanni and Nakov, Preslav},
	editor = {Aletras, Nikolaos and De Clercq, Orphee},
	month = mar,
	year = {2024},
	pages = {207--213},
}

@article{hovlandInfluenceSourceCredibility1951,
	title = {The influence of source credibility on communication effectiveness},
	volume = {15},
	issn = {0033-362X},
	doi = {10.1086/266350},
	number = {4},
	journal = {The Public Opinion Quarterly},
	publisher = {[Oxford University Press, American Association for Public Opinion Research]},
	author = {Hovland, Carl I. and Weiss, Walter},
	year = {1951},
	pages = {635--650},
}

@article{breumPersuasivePowerLarge2024,
	title = {The persuasive power of large language models},
	volume = {18},
	copyright = {Copyright (c) 2024 Association for the Advancement of Artificial Intelligence},
	issn = {2334-0770},
	url = {https://ojs.aaai.org/index.php/ICWSM/article/view/31304},
	doi = {10.1609/icwsm.v18i1.31304},
	language = {en},
	urldate = {2025-09-28},
	journal = {Proceedings of the International AAAI Conference on Web and Social Media},
	author = {Breum, Simon Martin and Egdal, Daniel Vædele and Mortensen, Victor Gram and Møller, Anders Giovanni and Aiello, Luca Maria},
	month = may,
	year = {2024},
	pages = {152--163},
}

@book{jowettPropagandaPersuasion2012,
	address = {Thousand Oaks, Calif},
	edition = {5th ed},
	title = {Propaganda \& persuasion},
	isbn = {978-1-4129-7782-1},
	language = {en},
	publisher = {SAGE},
	author = {Jowett, Garth S. and O'Donnell, Victoria},
	year = {2012},
}

@article{songSourceEffectsPsychological2018,
	title = {Source effects on psychological reactance to regulatory policies: {The} role of trust and similarity},
	volume = {40},
	shorttitle = {Source {Effects} on {Psychological} {Reactance} to {Regulatory} {Policies}},
	doi = {10.1177/1075547018791293},
	number = {4},
	journal = {Science Communication},
	author = {Song, Hwanseok and McComas, Katherine and Schuler, Krysten},
	month = aug,
	year = {2018},
	pages = {525--553},
}

@article{forgasMoodJudgmentAffect1995,
	title = {Mood and judgment: {The} affect infusion model ({AIM})},
	volume = {117},
	issn = {1939-1455},
	shorttitle = {Mood and judgment},
	doi = {10.1037/0033-2909.117.1.39},
	number = {1},
	journal = {Psychological Bulletin},
	publisher = {American Psychological Association},
	author = {Forgas, Joseph P.},
	year = {1995},
	pages = {39--66},
}

@article{kahnemanProspectTheoryAnalysis1979,
	title = {Prospect theory: {An} analysis of decision under risk},
	volume = {47},
	issn = {0012-9682},
	shorttitle = {Prospect {Theory}},
	url = {https://www.jstor.org/stable/1914185},
	doi = {10.2307/1914185},
	number = {2},
	urldate = {2026-05-01},
	journal = {Econometrica},
	publisher = {[Wiley, Econometric Society]},
	author = {Kahneman, Daniel and Tversky, Amos},
	year = {1979},
	pages = {263--291},
}

@article{youdenIndexRatingDiagnostic1950,
	title = {Index for rating diagnostic tests},
	volume = {3},
	issn = {0008-543X},
	doi = {10.1002/1097-0142(1950)3:1<32::aid-cncr2820030106>3.0.co;2-3},
	language = {eng},
	number = {1},
	journal = {Cancer},
	author = {Youden, W. J.},
	month = jan,
	year = {1950},
	pages = {32--35},
}

@misc{liuRoBERTaRobustlyOptimized2019,
	title = {{RoBERTa}: {A} {Robustly} {Optimized} {BERT} {Pretraining} {Approach}},
	shorttitle = {{RoBERTa}},
	url = {http://arxiv.org/abs/1907.11692},
	doi = {10.48550/arXiv.1907.11692},
	urldate = {2026-04-29},
	publisher = {arXiv},
	author = {Liu, Yinhan and Ott, Myle and Goyal, Naman and Du, Jingfei and Joshi, Mandar and Chen, Danqi and Levy, Omer and Lewis, Mike and Zettlemoyer, Luke and Stoyanov, Veselin},
	month = jul,
	year = {2019},
	note = {arXiv:1907.11692 [cs]},
}

@article{huttoVADERParsimoniousRuleBased2014,
	title = {{VADER}: {A} {Parsimonious} {Rule}-{Based} {Model} for {Sentiment} {Analysis} of {Social} {Media} {Text}},
	volume = {8},
	copyright = {Copyright (c) 2021 Proceedings of the International AAAI Conference on Web and Social Media},
	issn = {2334-0770},
	shorttitle = {{VADER}},
	url = {https://ojs.aaai.org/index.php/ICWSM/article/view/14550},
	doi = {10.1609/icwsm.v8i1.14550},
	language = {en},
	number = {1},
	urldate = {2026-04-29},
	journal = {Proceedings of the International AAAI Conference on Web and Social Media},
	author = {Hutto, C. and Gilbert, Eric},
	month = may,
	year = {2014},
	pages = {216--225},
}

@misc{mohammadNRCVADLexicon2025,
	title = {{NRC} {VAD} {Lexicon} v2: {Norms} for {Valence}, {Arousal}, and {Dominance} for over 55k {English} {Terms}},
	shorttitle = {{NRC} {VAD} {Lexicon} v2},
	url = {http://arxiv.org/abs/2503.23547},
	doi = {10.48550/arXiv.2503.23547},
	urldate = {2026-04-29},
	publisher = {arXiv},
	author = {Mohammad, Saif M.},
	month = mar,
	year = {2025},
	note = {arXiv:2503.23547 [cs]},
}

@article{boydDevelopmentPsychometricProperties,
	address = {Austin, TX},
	title = {The {Development} and {Psychometric} {Properties} of {LIWC}-22},
	language = {en},
	author = {Boyd, Ryan L and Ashokkumar, Ashwini and Seraj, Sara and Pennebaker, James W.},
	year = {2022},
}

@article{bassiDecodingPersuasionSurvey2024,
	title = {Decoding persuasion: a survey on {ML} and {NLP} methods for the study of online persuasion},
	volume = {9},
	issn = {2297-900X},
	shorttitle = {Decoding persuasion},
	url = {https://www.frontiersin.org/journals/communication/articles/10.3389/fcomm.2024.1457433/full},
	doi = {10.3389/fcomm.2024.1457433},
	language = {English},
	urldate = {2026-04-29},
	journal = {Frontiers in Communication},
	publisher = {Frontiers},
	author = {Bassi, Davide and Fomsgaard, Søren and Pereira-Fariña, Martín},
	month = oct,
	year = {2024},
}

@article{murakiConcretenessRatings620002023,
	title = {Concreteness ratings for 62,000 {English} multiword expressions},
	volume = {55},
	issn = {1554-3528},
	url = {https://doi.org/10.3758/s13428-022-01912-6},
	doi = {10.3758/s13428-022-01912-6},
	language = {en},
	number = {5},
	urldate = {2026-04-27},
	journal = {Behavior Research Methods},
	author = {Muraki, Emiko J. and Abdalla, Summer and Brysbaert, Marc and Pexman, Penny M.},
	month = aug,
	year = {2023},
	pages = {2522--2531},
}

@article{witteMetaanalysisFearAppeals2000,
	title = {A meta-analysis of fear appeals: implications for effective public health campaigns},
	volume = {27},
	issn = {1090-1981},
	shorttitle = {A meta-analysis of fear appeals},
	doi = {10.1177/109019810002700506},
	language = {eng},
	number = {5},
	journal = {Health Education \& Behavior},
	author = {Witte, K. and Allen, M.},
	month = oct,
	year = {2000},
	pages = {591--615},
}

@article{chaikenHeuristicSystematicInformation1980,
	address = {US},
	title = {Heuristic versus systematic information processing and the use of source versus message cues in persuasion},
	volume = {39},
	issn = {1939-1315},
	doi = {10.1037/0022-3514.39.5.752},
	number = {5},
	journal = {Journal of Personality and Social Psychology},
	publisher = {American Psychological Association},
	author = {Chaiken, Shelly},
	year = {1980},
	pages = {752--766},
}

@article{karasmanPropagandaMechanismManipulation,
	title = {Propaganda as a mechanism of manipulation and encouragement to action},
	volume = {9},
	language = {en},
	journal = {RIThink},
	author = {Skuhala Karasman, Ivana},
	year = {2020},
}

@article{zebregsDifferentialImpactStatistical2015,
	title = {The differential impact of statistical and narrative evidence on beliefs, attitude, and intention: {A} meta-analysis},
	volume = {30},
	issn = {1041-0236},
	shorttitle = {The {Differential} {Impact} of {Statistical} and {Narrative} {Evidence} on {Beliefs}, {Attitude}, and {Intention}},
	url = {https://doi.org/10.1080/10410236.2013.842528},
	doi = {10.1080/10410236.2013.842528},
	number = {3},
	urldate = {2026-03-01},
	journal = {Health Communication},
	publisher = {Routledge},
	author = {Zebregs, Simon and van den Putte, Bas and Neijens, Peter and de Graaf, Anneke},
	month = mar,
	year = {2015},
	pages = {282--289},
}

@book{toulminUsesArgument2003,
	address = {Cambridge},
	edition = {2},
	title = {The uses of argument},
	isbn = {978-0-521-82748-5},
	url = {https://www.cambridge.org/core/books/uses-of-argument/26CF801BC12004587B66778297D5567C},
	doi = {10.1017/CBO9780511840005},
	urldate = {2025-12-08},
	publisher = {Cambridge University Press},
	author = {Toulmin, Stephen E.},
	year = {2003},
}

@article{slaterEntertainmentEducationElaboration2002,
	title = {Entertainment—education and elaboration likelihood: understanding the processing of narrative persuasion},
	volume = {12},
	issn = {1468-2885},
	shorttitle = {Entertainment—{Education} and {Elaboration} {Likelihood}},
	url = {https://onlinelibrary.wiley.com/doi/abs/10.1111/j.1468-2885.2002.tb00265.x},
	doi = {10.1111/j.1468-2885.2002.tb00265.x},
	language = {en},
	number = {2},
	urldate = {2026-02-22},
	journal = {Communication Theory},
	author = {Slater, Michael D. and Rouner, Donna},
	year = {2002},
	pages = {173--191},
}

@article{sebaldaAnalyzingCoherenceCohesion2025,
	title = {Analyzing coherence and cohesion in {Sara} {Duterte}'s vice presidential inaugural speech},
	volume = {7},
	copyright = {Copyright (c) 2025 Resyl Sebalda, Taray, Celeste Faye},
	issn = {2704-7156},
	url = {https://ijlls.org/index.php/ijlls/article/view/2052},
	doi = {10.36892/ijlls.v7i2.2052},
	language = {en},
	number = {2},
	urldate = {2026-03-09},
	journal = {International Journal of Language and Literary Studies},
	author = {Sebalda, Resyl and Taray, Celeste Faye and Tulud, Donnie M.},
	month = mar,
	year = {2025},
	pages = {260--282},
}

@article{reberEffectsPerceptualFluency1999,
	title = {Effects of perceptual fluency on judgments of truth},
	volume = {8},
	issn = {1053-8100},
	url = {https://www.sciencedirect.com/science/article/pii/S1053810099903860},
	doi = {10.1006/ccog.1999.0386},
	number = {3},
	urldate = {2026-02-22},
	journal = {Consciousness and Cognition},
	author = {Reber, Rolf and Schwarz, Norbert},
	month = sep,
	year = {1999},
	pages = {338--342},
}

@article{pornpitakpanPersuasivenessSourceCredibility2004,
	title = {The persuasiveness of source credibility: {A} critical review of five decades' evidence},
	volume = {34},
	issn = {1559-1816},
	shorttitle = {The {Persuasiveness} of {Source} {Credibility}},
	url = {https://onlinelibrary.wiley.com/doi/abs/10.1111/j.1559-1816.2004.tb02547.x},
	doi = {10.1111/j.1559-1816.2004.tb02547.x},
	language = {en},
	number = {2},
	urldate = {2026-03-09},
	journal = {Journal of Applied Social Psychology},
	author = {Pornpitakpan, Chanthika},
	year = {2004},
	pages = {243--281},
}

@incollection{pettyElaborationLikelihoodModel1986,
	title = {The elaboration likelihood model of persuasion},
	volume = {19},
	url = {https://www.sciencedirect.com/science/article/pii/S0065260108602142},
	doi = {10.1016/S0065-2601(08)60214-2},
	urldate = {2025-09-22},
	booktitle = {Advances in {Experimental} {Social} {Psychology}},
	publisher = {Academic Press},
	author = {Petty, Richard E. and Cacioppo, John T.},
	editor = {Berkowitz, Leonard},
	month = jan,
	year = {1986},
	pages = {123--205},
}

@article{packardHowConcreteLanguage2021,
	title = {How concrete language shapes customer satisfaction},
	volume = {47},
	issn = {0093-5301},
	url = {https://doi.org/10.1093/jcr/ucaa038},
	doi = {10.1093/jcr/ucaa038},
	number = {5},
	urldate = {2025-12-08},
	journal = {Journal of Consumer Research},
	author = {Packard, Grant and Berger, Jonah},
	month = feb,
	year = {2021},
	pages = {787--806},
}

@article{okeefeDoorintheFaceInfluenceStrategy1998,
	title = {The door-in-the-face influence strategy: {A} random-effects meta-analytic review},
	volume = {21},
	issn = {2380-8985, 2380-8977},
	shorttitle = {The {Door}-in-the-{Face} {Influence} {Strategy}},
	url = {https://academic.oup.com/anncom/article/21/1/1/7850305},
	doi = {10.1080/23808985.1998.11678947},
	language = {en},
	number = {1},
	urldate = {2026-03-29},
	journal = {Annals of the International Communication Association},
	author = {O’Keefe, Daniel J. and Hale, Scott L.},
	month = jan,
	year = {1998},
	pages = {1--33},
}

@article{okeefeHowHandleOpposing1999,
	title = {How to handle opposing arguments in persuasive messages: {A} meta-analytic review of the effects of one-sided and two-sided messages},
	volume = {22},
	issn = {2380-8985},
	shorttitle = {How to {Handle} {Opposing} {Arguments} in {Persuasive} {Messages}},
	url = {https://doi.org/10.1080/23808985.1999.11678963},
	doi = {10.1080/23808985.1999.11678963},
	number = {1},
	urldate = {2025-09-22},
	journal = {Annals of the International Communication Association},
	publisher = {Routledge},
	author = {O’Keefe, Daniel J.},
	month = jan,
	year = {1999},
	pages = {209--249},
}

@article{okeefeStandpointExplicitnessPersuasive1997a,
	title = {Standpoint explicitness and persuasive effect: {A} meta-analytic review of the effects of varying conclusion articulation in persuasive messages},
	volume = {34},
	issn = {1051-1431, 2576-8476},
	shorttitle = {Standpoint {Explicitness} and {Persuasive} {Effect}},
	url = {https://www.tandfonline.com/doi/full/10.1080/00028533.1997.11978023},
	doi = {10.1080/00028533.1997.11978023},
	language = {en},
	number = {1},
	urldate = {2026-03-22},
	journal = {Argumentation and Advocacy},
	author = {O'Keefe, Daniel J.},
	month = jun,
	year = {1997},
	pages = {1--12},
}

@inproceedings{lukinArgumentStrengthEye2017,
	address = {Valencia, Spain},
	title = {Argument strength is in the eye of the beholder: audience effects in persuasion},
	shorttitle = {Argument {Strength} is in the {Eye} of the {Beholder}},
	url = {https://aclanthology.org/E17-1070/},
	urldate = {2026-01-26},
	booktitle = {Proceedings of the 15th {Conference} of the {European} {Chapter} of the {Association} for {Computational} {Linguistics}: {Volume} 1, {Long} {Papers}},
	publisher = {Association for Computational Linguistics},
	author = {Lukin, Stephanie and Anand, Pranav and Walker, Marilyn and Whittaker, Steve},
	editor = {Lapata, Mirella and Blunsom, Phil and Koller, Alexander},
	month = apr,
	year = {2017},
	pages = {742--753},
}

@article{hideyPersuasiveInfluenceDetection2018,
	title = {Persuasive influence detection: {The} role of argument sequencing},
	volume = {32},
	copyright = {Copyright (c)},
	issn = {2374-3468},
	shorttitle = {Persuasive {Influence} {Detection}},
	url = {https://ojs.aaai.org/index.php/AAAI/article/view/12003},
	doi = {10.1609/aaai.v32i1.12003},
	language = {en},
	number = {1},
	urldate = {2025-09-22},
	journal = {Proceedings of the AAAI Conference on Artificial Intelligence},
	author = {Hidey, Christopher and McKeown, Kathleen},
	month = apr,
	year = {2018},
}

@article{cruzSecondPersonPronouns2017,
	title = {Second person pronouns enhance consumer involvement and brand attitude},
	volume = {39},
	issn = {1094-9968},
	url = {https://journals.sagepub.com/action/showAbstract},
	doi = {10.1016/j.intmar.2017.05.001},
	language = {EN},
	number = {1},
	urldate = {2026-03-01},
	journal = {Journal of Interactive Marketing},
	publisher = {SAGE Publications},
	author = {Cruz, Ryan E. and Leonhardt, James M. and Pezzuti, Todd},
	month = aug,
	year = {2017},
	pages = {104--116},
}

@article{carpenterMetaAnalysisEffectivenessYou2013,
	title = {A meta-analysis of the effectiveness of the “but you are free” compliance-gaining technique},
	volume = {64},
	issn = {1051-0974},
	url = {https://doi.org/10.1080/10510974.2012.727941},
	doi = {10.1080/10510974.2012.727941},
	number = {1},
	urldate = {2026-03-29},
	journal = {Communication Studies},
	publisher = {Routledge},
	author = {Carpenter, Christopher J.},
	month = jan,
	year = {2013},
	pages = {6--17},
}

@inproceedings{carlileGiveMeMore2018,
	address = {Melbourne, Australia},
	title = {Give me more feedback: {Annotating} argument persuasiveness and related attributes in student essays},
	shorttitle = {Give {Me} {More} {Feedback}},
	url = {https://aclanthology.org/P18-1058/},
	doi = {10.18653/v1/P18-1058},
	urldate = {2025-10-18},
	booktitle = {Proceedings of the 56th {Annual} {Meeting} of the {Association} for {Computational} {Linguistics} ({Volume} 1: {Long} {Papers})},
	publisher = {Association for Computational Linguistics},
	author = {Carlile, Winston and Gurrapadi, Nishant and Ke, Zixuan and Ng, Vincent},
	editor = {Gurevych, Iryna and Miyao, Yusuke},
	month = jul,
	year = {2018},
	pages = {621--631},
}

@article{burgerFootintheDoorComplianceProcedure1999,
	title = {The foot-in-the-door compliance procedure: {A} multiple-process analysis and review},
	volume = {3},
	issn = {1088-8683},
	shorttitle = {The {Foot}-in-the-{Door} {Compliance} {Procedure}},
	url = {https://doi.org/10.1207/s15327957pspr0304_2},
	doi = {10.1207/s15327957pspr0304_2},
	language = {EN},
	number = {4},
	urldate = {2026-04-10},
	journal = {Personality and Social Psychology Review},
	publisher = {SAGE Publications Inc},
	author = {Burger, Jerry M.},
	month = nov,
	year = {1999},
	pages = {303--325},
}

@article{blumenauVariablePersuasivenessPolitical2024,
	title = {The variable persuasiveness of political rhetoric},
	volume = {68},
	copyright = {© 2022 The Authors. American Journal of Political Science published by Wiley Periodicals LLC on behalf of Midwest Political Science Association.},
	issn = {1540-5907},
	url = {https://onlinelibrary.wiley.com/doi/abs/10.1111/ajps.12703},
	doi = {10.1111/ajps.12703},
	language = {en},
	number = {1},
	urldate = {2026-02-27},
	journal = {American Journal of Political Science},
	author = {Blumenau, Jack and Lauderdale, Benjamin E.},
	year = {2024},
	pages = {255--270},
}

@article{bigsbyExemplificationTheoryReview2019,
	title = {Exemplification theory: {A} review and meta-analysis of exemplar messages},
	volume = {43},
	issn = {2380-8985},
	shorttitle = {Exemplification {Theory}},
	url = {https://doi.org/10.1080/23808985.2019.1681903},
	doi = {10.1080/23808985.2019.1681903},
	number = {4},
	urldate = {2026-03-01},
	journal = {Annals of the International Communication Association},
	author = {Bigsby, Elisabeth and Bigman, Cabral A. and Gonzalez, Andrea Martinez},
	month = sep,
	year = {2019},
	pages = {273--296},
}

@article{higginsEthosLogosPathos2012,
	series = {Analyzing the {Quality}, {Meaning} and {Accountability} of {Organizational} {Communication}},
	title = {Ethos, logos, pathos: {Strategies} of persuasion in social/environmental reports},
	volume = {36},
	issn = {0155-9982},
	shorttitle = {\textit{{Ethos}}, \textit{logos}, \textit{pathos}},
	url = {https://www.sciencedirect.com/science/article/pii/S0155998212000178},
	doi = {10.1016/j.accfor.2012.02.003},
	number = {3},
	urldate = {2026-04-26},
	journal = {Accounting Forum},
	author = {Higgins, Colin and Walker, Robyn},
	month = sep,
	year = {2012},
	pages = {194--208},
}

@inproceedings{pauliMeasuringBenchmarkingLarge2025,
	address = {Albuquerque, New Mexico},
	title = {Measuring and {Benchmarking} {Large} {Language} {Models}' {Capabilities} to {Generate} {Persuasive} {Language}},
	isbn = {979-8-89176-189-6},
	url = {https://aclanthology.org/2025.naacl-long.506/},
	doi = {10.18653/v1/2025.naacl-long.506},
	urldate = {2026-04-20},
	booktitle = {Proceedings of the 2025 {Conference} of the {Nations} of the {Americas} {Chapter} of the {Association} for {Computational} {Linguistics}: {Human} {Language} {Technologies} ({Volume} 1: {Long} {Papers})},
	publisher = {Association for Computational Linguistics},
	author = {Pauli, Amalie Brogaard and Augenstein, Isabelle and Assent, Ira},
	editor = {Chiruzzo, Luis and Ritter, Alan and Wang, Lu},
	month = apr,
	year = {2025},
	pages = {10056--10075},
}

@article{gallagherHealthMessageFraming2012,
	title = {Health message framing effects on attitudes, intentions, and behavior: a meta-analytic review},
	volume = {43},
	issn = {1532-4796},
	shorttitle = {Health message framing effects on attitudes, intentions, and behavior},
	doi = {10.1007/s12160-011-9308-7},
	language = {eng},
	number = {1},
	journal = {Annals of Behavioral Medicine: A Publication of the Society of Behavioral Medicine},
	author = {Gallagher, Kristel M. and Updegraff, John A.},
	month = feb,
	year = {2012},
	pages = {101--116},
}

@article{chenFrameworkModeratorsSocial2024,
	title = {A framework of moderators in social norm-based message persuasiveness based on a systematic review},
	volume = {50},
	issn = {1468-2958},
	url = {https://doi.org/10.1093/hcr/hqad043},
	doi = {10.1093/hcr/hqad043},
	number = {2},
	urldate = {2026-04-16},
	journal = {Human Communication Research},
	author = {Chen, Junhan and Xia, Shilin and Lin, Tong},
	month = apr,
	year = {2024},
	pages = {285--298},
}

@article{tropeConstrualLevelTheoryPsychological2010,
	title = {Construal-{Level} {Theory} of {Psychological} {Distance}},
	volume = {117},
	issn = {0033-295X},
	url = {https://pmc.ncbi.nlm.nih.gov/articles/PMC3152826/},
	doi = {10.1037/a0018963},
	number = {2},
	urldate = {2026-04-16},
	journal = {Psychological review},
	author = {Trope, Yaacov and Liberman, Nira},
	month = apr,
	year = {2010},
	pages = {440--463},
}

@article{blankenshipRhetoricalQuestionUse2006,
	title = {Rhetorical {Question} {Use} and {Resistance} to {Persuasion}: {An} {Attitude} {Strength} {Analysis}},
	volume = {25},
	issn = {0261-927X},
	shorttitle = {Rhetorical {Question} {Use} and {Resistance} to {Persuasion}},
	url = {https://doi.org/10.1177/0261927X06286380},
	doi = {10.1177/0261927X06286380},
	language = {EN},
	number = {2},
	urldate = {2026-04-11},
	journal = {Journal of Language and Social Psychology},
	publisher = {SAGE Publications Inc},
	author = {Blankenship, Kevin L. and Craig, Traci Y.},
	month = jun,
	year = {2006},
	pages = {111--128},
}

@article{kempPeriodReallyPissed2025,
	title = {Is the period really “pissed”? {The} effect of punctuation and message length on perceptions in digital communication},
	volume = {97},
	issn = {0736-5853},
	shorttitle = {Is the period really “pissed”?},
	url = {https://www.sciencedirect.com/science/article/pii/S0736585325000036},
	doi = {10.1016/j.tele.2025.102241},
	urldate = {2026-04-11},
	journal = {Telematics and Informatics},
	author = {Kemp, Nenagh and Kovacic, Rebecca and Beyersmann, Elisabeth},
	month = feb,
	year = {2025},
	pages = {102241},
}

@article{decheneTruthTruthMetaanalytic2010,
	title = {The truth about the truth: a meta-analytic review of the truth effect},
	volume = {14},
	issn = {1532-7957},
	shorttitle = {The truth about the truth},
	doi = {10.1177/1088868309352251},
	language = {eng},
	number = {2},
	journal = {Personality and Social Psychology Review: An Official Journal of the Society for Personality and Social Psychology, Inc},
	author = {Dechêne, Alice and Stahl, Christoph and Hansen, Jochim and Wänke, Michaela},
	month = may,
	year = {2010},
	pages = {238--257},
}

@article{burnkrantEffectsSelfreferencingPersuasion1995,
	address = {US},
	title = {Effects of self-referencing on persuasion},
	volume = {22},
	issn = {1537-5277},
	doi = {10.1086/209432},
	number = {1},
	journal = {Journal of Consumer Research},
	publisher = {Univ of Chicago Press},
	author = {Burnkrant, Robert E. and Unnava, H. Rao},
	year = {1995},
	pages = {17--26},
}

@article{ladeiraMetaanalysisEffectsProduct2023,
	title = {A meta‐analysis on the effects of product scarcity},
	volume = {40},
	doi = {10.1002/mar.21816},
	journal = {Psychology \& Marketing},
	author = {Ladeira, Wagner and Lim, Weng Marc and Santini, Fernando and Rasul, Tareq and Perin, Marcelo and Altinay, Levent},
	month = apr,
	year = {2023},
	pages = {1267--1279},
}

@article{kimMetaanalysisGainLoss2025,
	title = {A meta-analysis of gain–loss framing effects in narrative persuasion},
	volume = {75},
	issn = {0021-9916},
	url = {https://doi.org/10.1093/joc/jqaf002},
	doi = {10.1093/joc/jqaf002},
	number = {3},
	urldate = {2026-03-29},
	journal = {Journal of Communication},
	author = {Kim, Hye Kyung and Chen, Minyi},
	month = jun,
	year = {2025},
	pages = {220--230},
}

@article{okeefeRelativePersuasivenessGainframed2007,
	title = {The relative persuasiveness of gain-framed and loss-framed messages for encouraging disease prevention behaviors: a meta-analytic review},
	volume = {12},
	issn = {1081-0730},
	shorttitle = {The relative persuasiveness of gain-framed and loss-framed messages for encouraging disease prevention behaviors},
	doi = {10.1080/10810730701615198},
	language = {eng},
	number = {7},
	journal = {Journal of Health Communication},
	author = {O'Keefe, Daniel J. and Jensen, Jakob D.},
	year = {2007},
	pages = {623--644},
}

@article{burgerIncreasingComplianceImproving1986,
	address = {US},
	title = {Increasing compliance by improving the deal: {The} that's-not-all technique},
	volume = {51},
	issn = {1939-1315},
	shorttitle = {Increasing compliance by improving the deal},
	doi = {10.1037/0022-3514.51.2.277},
	number = {2},
	journal = {Journal of Personality and Social Psychology},
	publisher = {American Psychological Association},
	author = {Burger, Jerry M.},
	year = {1986},
	pages = {277--283},
}

@incollection{dillardDiscreteEmotionsPersuasion,
	title = {Discrete {Emotions} and {Persuasion}},
	doi = {https://doi.org/10.4135/9781412976046.n15},
	booktitle = {The {Persuasion} {Handbook}: {Developments} in {Theory} and {Practice}},
	publisher = {SAGE Publications, Inc.},
	author = {Nabi, Robin L.},
	year = {2002},
	pages = {289--308},
}

@article{rothmanStrategicUseGain2006,
	address = {United Kingdom},
	title = {The {Strategic} {Use} of {Gain}- and {Loss}-{Framed} {Messages} to {Promote} {Healthy} {Behavior}: {How} {Theory} {Can} {Inform} {Practice}},
	volume = {56},
	issn = {1460-2466},
	shorttitle = {The {Strategic} {Use} of {Gain}- and {Loss}-{Framed} {Messages} to {Promote} {Healthy} {Behavior}},
	doi = {10.1111/j.1460-2466.2006.00290.x},
	number = {Suppl 1},
	journal = {Journal of Communication},
	publisher = {Blackwell Publishing},
	author = {Rothman, Alexander J. and Bartels, Roger D. and Wlaschin, Jhon and Salovey, Peter},
	year = {2006},
	pages = {S202--S220},
}

@article{ciukMoralConvictionEmotion2020,
	title = {Moral {Conviction}, {Emotion}, and the {Influence} of {Episodic} versus {Thematic} {Frames}},
	volume = {38},
	doi = {10.1080/10584609.2020.1793847},
	journal = {Political Communication},
	author = {Ciuk, David and Rottman, Joshua},
	month = sep,
	year = {2020},
	pages = {1--20},
}

@inproceedings{mohammadWordAffectIntensities2018,
	address = {Miyazaki, Japan},
	title = {Word {Affect} {Intensities}},
	url = {https://aclanthology.org/L18-1027/},
	urldate = {2026-03-25},
	booktitle = {Proceedings of the {Eleventh} {International} {Conference} on {Language} {Resources} and {Evaluation} ({LREC} 2018)},
	publisher = {European Language Resources Association (ELRA)},
	author = {Mohammad, Saif},
	editor = {Calzolari, Nicoletta and Choukri, Khalid and Cieri, Christopher and Declerck, Thierry and Goggi, Sara and Hasida, Koiti and Isahara, Hitoshi and Maegaard, Bente and Mariani, Joseph and Mazo, Hélène and Moreno, Asuncion and Odijk, Jan and Piperidis, Stelios and Tokunaga, Takenobu},
	month = may,
	year = {2018},
}

@inproceedings{stabIdentifyingArgumentativeDiscourse2014,
	address = {Doha, Qatar},
	title = {Identifying {Argumentative} {Discourse} {Structures} in {Persuasive} {Essays}},
	url = {https://aclanthology.org/D14-1006/},
	doi = {10.3115/v1/D14-1006},
	urldate = {2026-03-25},
	booktitle = {Proceedings of the 2014 {Conference} on {Empirical} {Methods} in {Natural} {Language} {Processing} ({EMNLP})},
	publisher = {Association for Computational Linguistics},
	author = {Stab, Christian and Gurevych, Iryna},
	editor = {Moschitti, Alessandro and Pang, Bo and Daelemans, Walter},
	month = oct,
	year = {2014},
	pages = {46--56},
}

@inproceedings{farkasCoNLL2010SharedTask2010,
	address = {Uppsala, Sweden},
	title = {The {CoNLL}-2010 {Shared} {Task}: {Learning} to {Detect} {Hedges} and their {Scope} in {Natural} {Language} {Text}},
	shorttitle = {The {CoNLL}-2010 {Shared} {Task}},
	url = {https://aclanthology.org/W10-3001/},
	urldate = {2026-03-25},
	booktitle = {Proceedings of the {Fourteenth} {Conference} on {Computational} {Natural} {Language} {Learning} – {Shared} {Task}},
	publisher = {Association for Computational Linguistics},
	author = {Farkas, Richárd and Vincze, Veronika and Móra, György and Csirik, János and Szarvas, György},
	editor = {Farkas, Richárd and Vincze, Veronika and Szarvas, György and Móra, György and Csirik, János},
	month = jul,
	year = {2010},
	pages = {1--12},
}

@inproceedings{dasanmartinoSemEval2020Task112020,
	address = {Barcelona (online)},
	title = {{SemEval}-2020 {Task} 11: {Detection} of {Propaganda} {Techniques} in {News} {Articles}},
	shorttitle = {{SemEval}-2020 {Task} 11},
	url = {https://aclanthology.org/2020.semeval-1.186/},
	doi = {10.18653/v1/2020.semeval-1.186},
	urldate = {2026-03-25},
	booktitle = {Proceedings of the {Fourteenth} {Workshop} on {Semantic} {Evaluation}},
	publisher = {International Committee for Computational Linguistics},
	author = {Da San Martino, Giovanni and Barrón-Cedeño, Alberto and Wachsmuth, Henning and Petrov, Rostislav and Nakov, Preslav},
	editor = {Herbelot, Aurelie and Zhu, Xiaodan and Palmer, Alexis and Schneider, Nathan and May, Jonathan and Shutova, Ekaterina},
	month = dec,
	year = {2020},
	pages = {1377--1414},
}

@inproceedings{rietscheSpecificityHelpfulnessPeertoPeer2022,
	address = {Seattle, Washington},
	title = {The {Specificity} and {Helpfulness} of {Peer}-to-{Peer} {Feedback} in {Higher} {Education}},
	url = {https://aclanthology.org/2022.bea-1.15/},
	doi = {10.18653/v1/2022.bea-1.15},
	urldate = {2026-03-24},
	booktitle = {Proceedings of the 17th {Workshop} on {Innovative} {Use} of {NLP} for {Building} {Educational} {Applications} ({BEA} 2022)},
	publisher = {Association for Computational Linguistics},
	author = {Rietsche, Roman and Caines, Andrew and Schramm, Cornelius and Pfütze, Dominik and Buttery, Paula},
	editor = {Kochmar, Ekaterina and Burstein, Jill and Horbach, Andrea and Laarmann-Quante, Ronja and Madnani, Nitin and Tack, Anaïs and Yaneva, Victoria and Yuan, Zheng and Zesch, Torsten},
	month = jul,
	year = {2022},
	pages = {107--117},
}

@article{zhaoMeasurePerceivedArgument2011,
	title = {A {Measure} of {Perceived} {Argument} {Strength}: {Reliability} and {Validity}},
	volume = {5},
	issn = {1931-2458},
	shorttitle = {A {Measure} of {Perceived} {Argument} {Strength}},
	url = {https://pmc.ncbi.nlm.nih.gov/articles/PMC4283835/},
	doi = {10.1080/19312458.2010.547822},
	number = {1},
	urldate = {2026-03-23},
	journal = {Communication methods and measures},
	author = {Zhao, Xiaoquan and Strasser, Andrew and Cappella, Joseph N. and Lerman, Caryn and Fishbein, Martin},
	month = mar,
	year = {2011},
	pages = {48--75},
}

@article{pettyEffectsInvolvementResponses1984,
	address = {US},
	title = {The effects of involvement on responses to argument quantity and quality: {Central} and peripheral routes to persuasion},
	volume = {46},
	issn = {1939-1315},
	shorttitle = {The effects of involvement on responses to argument quantity and quality},
	doi = {10.1037/0022-3514.46.1.69},
	number = {1},
	journal = {Journal of Personality and Social Psychology},
	publisher = {American Psychological Association},
	author = {Petty, Richard E. and Cacioppo, John T.},
	year = {1984},
	pages = {69--81},
}

@article{hovlandExperimentalComparisonConclusiondrawing1952,
	address = {US},
	title = {An experimental comparison of conclusion-drawing by the communicator and by the audience},
	volume = {47},
	issn = {0096-851X},
	doi = {10.1037/h0059833},
	number = {3},
	journal = {The Journal of Abnormal and Social Psychology},
	publisher = {American Psychological Association},
	author = {Hovland, Carl I. and Mandell, Wallace},
	year = {1952},
	pages = {581--588},
}

@article{kaakinenInfluenceTextCohesion2011,
	title = {Influence of text cohesion on the persuasive power of expository text},
	volume = {52},
	copyright = {© 2011 The Authors. Scandinavian Journal of Psychology © 2011 The Scandinavian Psychological Associations},
	issn = {1467-9450},
	url = {https://onlinelibrary.wiley.com/doi/abs/10.1111/j.1467-9450.2010.00863.x},
	doi = {10.1111/j.1467-9450.2010.00863.x},
	language = {en},
	number = {3},
	urldate = {2026-03-09},
	journal = {Scandinavian Journal of Psychology},
	author = {Kaakinen, Johanna K. and Salonen, Jonna and Venäläinen, Paula and Hyönä, Jukka},
	year = {2011},
	pages = {201--208},
}

@article{bulleePersuasionSecurityAwareness2015,
	title = {The persuasion and security awareness experiment: reducing the success of social engineering attacks},
	volume = {11},
	issn = {1572-8315},
	shorttitle = {The persuasion and security awareness experiment},
	url = {https://doi.org/10.1007/s11292-014-9222-7},
	doi = {10.1007/s11292-014-9222-7},
	language = {en},
	number = {1},
	urldate = {2026-03-09},
	journal = {Journal of Experimental Criminology},
	author = {Bullée, Jan-Willem H. and Montoya, Lorena and Pieters, Wolter and Junger, Marianne and Hartel, Pieter H.},
	month = mar,
	year = {2015},
	pages = {97--115},
}

@article{rainsNaturePsychologicalReactance2013,
	title = {The {Nature} of {Psychological} {Reactance} {Revisited}: a {Meta}-{Analytic} {Review}},
	volume = {39},
	issn = {0360-3989},
	shorttitle = {The {Nature} of {Psychological} {Reactance} {Revisited}},
	url = {https://doi.org/10.1111/j.1468-2958.2012.01443.x},
	doi = {10.1111/j.1468-2958.2012.01443.x},
	number = {1},
	urldate = {2026-03-04},
	journal = {Human Communication Research},
	author = {Rains, Stephen A.},
	month = jan,
	year = {2013},
	pages = {47--73},
}

@article{fujitaInfluencingAttitudesDistant2008,
	title = {Influencing attitudes toward near and distant objects},
	volume = {44},
	issn = {0022-1031},
	url = {https://www.sciencedirect.com/science/article/pii/S0022103107001503},
	doi = {10.1016/j.jesp.2007.10.005},
	number = {3},
	urldate = {2026-03-01},
	journal = {Journal of Experimental Social Psychology},
	author = {Fujita, Kentaro and Eyal, Tal and Chaiken, Shelly and Trope, Yaacov and Liberman, Nira},
	month = may,
	year = {2008},
	pages = {562--572},
}

@article{kimTestingAdditiveModel2012,
	title = {Testing an additive model for the effectiveness of evidence on the persuasiveness of a message},
	volume = {7},
	issn = {1553-4510},
	url = {https://doi.org/10.1080/15534510.2012.658285},
	doi = {10.1080/15534510.2012.658285},
	number = {2},
	urldate = {2026-03-01},
	journal = {Social Influence},
	publisher = {Routledge},
	author = {Kim, Sang-Yeon and Allen, Mike and Gattoni, Ali and Grimes, Denis and Herrman, Anna M. and Huang, Han and Kim, Jihyun and Lu, Shan and Maier, Melissa and May, Amy and Omachinski, Kim and Omori, Kikuko and Tenzek, Kelly and Turkiewicz, Katie LaPlant and Zhang, Yan},
	month = apr,
	year = {2012},
	pages = {65--77},
}

@article{allenComparingPersuasivenessNarrative1997,
	title = {Comparing the persuasiveness of narrative and statistical evidence using meta‐analysis},
	volume = {14},
	issn = {0882-4096},
	url = {https://doi.org/10.1080/08824099709388654},
	doi = {10.1080/08824099709388654},
	number = {2},
	urldate = {2026-03-01},
	journal = {Communication Research Reports},
	publisher = {Routledge},
	author = {Allen, Mike and Preiss, Raymond W.},
	month = mar,
	year = {1997},
	pages = {125--131},
}

@article{worchelEffectsSupplyDemand1975,
	address = {US},
	title = {Effects of supply and demand on ratings of object value},
	volume = {32},
	issn = {1939-1315},
	doi = {10.1037/0022-3514.32.5.906},
	number = {5},
	journal = {Journal of Personality and Social Psychology},
	publisher = {American Psychological Association},
	author = {Worchel, Stephen and Lee, Jerry and Adewole, Akanbi},
	year = {1975},
	pages = {906--914},
}

@article{schwarzMoodMisattributionJudgments1983,
	address = {US},
	title = {Mood, misattribution, and judgments of well-being: {Informative} and directive functions of affective states},
	volume = {45},
	issn = {1939-1315},
	shorttitle = {Mood, misattribution, and judgments of well-being},
	doi = {10.1037/0022-3514.45.3.513},
	number = {3},
	journal = {Journal of Personality and Social Psychology},
	publisher = {American Psychological Association},
	author = {Schwarz, Norbert and Clore, Gerald L.},
	year = {1983},
	pages = {513--523},
}

@incollection{pettyMultipleRolesAffect1991,
	address = {Elmsford, NY, US},
	series = {International series in experimental social psychology},
	title = {Multiple roles for affect in persuasion},
	isbn = {978-0-08-040236-9 978-0-08-040235-2},
	booktitle = {Emotion and social judgments},
	publisher = {Pergamon Press},
	author = {Petty, Richard E. and Gleicher, Faith and Baker, Sara M.},
	year = {1991},
	pages = {181--200},
}

@inproceedings{shenHEARTfeltNarrativesTracing2024,
	address = {Miami, Florida, USA},
	title = {{HEART}-felt {Narratives}: {Tracing} {Empathy} and {Narrative} {Style} in {Personal} {Stories} with {LLMs}},
	shorttitle = {{HEART}-felt {Narratives}},
	url = {https://aclanthology.org/2024.emnlp-main.59/},
	doi = {10.18653/v1/2024.emnlp-main.59},
	urldate = {2026-02-11},
	booktitle = {Proceedings of the 2024 {Conference} on {Empirical} {Methods} in {Natural} {Language} {Processing}},
	publisher = {Association for Computational Linguistics},
	author = {Shen, Jocelyn and Mire, Joel and Park, Hae Won and Breazeal, Cynthia and Sap, Maarten},
	editor = {Al-Onaizan, Yaser and Bansal, Mohit and Chen, Yun-Nung},
	month = nov,
	year = {2024},
	pages = {1026--1046},
}

@article{brysbaertConcretenessRatings402014,
	title = {Concreteness ratings for 40 thousand generally known {English} word lemmas},
	volume = {46},
	issn = {1554-3528},
	url = {https://doi.org/10.3758/s13428-013-0403-5},
	doi = {10.3758/s13428-013-0403-5},
	language = {en},
	number = {3},
	urldate = {2026-02-10},
	journal = {Behavior Research Methods},
	author = {Brysbaert, Marc and Warriner, Amy Beth and Kuperman, Victor},
	month = sep,
	year = {2014},
	pages = {904--911},
}

@article{gretzLargeScaleDatasetArgument2020,
	title = {A {Large}-{Scale} {Dataset} for {Argument} {Quality} {Ranking}: {Construction} and {Analysis}},
	volume = {34},
	copyright = {Copyright (c) 2020 Association for the Advancement of Artificial Intelligence},
	issn = {2374-3468},
	shorttitle = {A {Large}-{Scale} {Dataset} for {Argument} {Quality} {Ranking}},
	url = {https://ojs.aaai.org/index.php/AAAI/article/view/6285},
	doi = {10.1609/aaai.v34i05.6285},
	language = {en},
	number = {05},
	urldate = {2026-01-27},
	journal = {Proceedings of the AAAI Conference on Artificial Intelligence},
	author = {Gretz, Shai and Friedman, Roni and Cohen-Karlik, Edo and Toledo, Assaf and Lahav, Dan and Aharonov, Ranit and Slonim, Noam},
	month = apr,
	year = {2020},
	pages = {7805--7813},
}

@inproceedings{rombergPerspectivistTurnArgument2025,
	address = {Albuquerque, New Mexico},
	title = {Towards a {Perspectivist} {Turn} in {Argument} {Quality} {Assessment}},
	isbn = {979-8-89176-189-6},
	url = {https://aclanthology.org/2025.naacl-long.382/},
	doi = {10.18653/v1/2025.naacl-long.382},
	urldate = {2026-01-26},
	booktitle = {Proceedings of the 2025 {Conference} of the {Nations} of the {Americas} {Chapter} of the {Association} for {Computational} {Linguistics}: {Human} {Language} {Technologies} ({Volume} 1: {Long} {Papers})},
	publisher = {Association for Computational Linguistics},
	author = {Romberg, Julia and Maurer, Maximilian and Wachsmuth, Henning and Lapesa, Gabriella},
	editor = {Chiruzzo, Luis and Ritter, Alan and Wang, Lu},
	month = apr,
	year = {2025},
	pages = {7458--7485},
}

@inproceedings{wachsmuthComputationalArgumentationQuality2017,
	address = {Valencia, Spain},
	title = {Computational {Argumentation} {Quality} {Assessment} in {Natural} {Language}},
	url = {https://aclanthology.org/E17-1017/},
	urldate = {2026-01-26},
	booktitle = {Proceedings of the 15th {Conference} of the {European} {Chapter} of the {Association} for {Computational} {Linguistics}: {Volume} 1, {Long} {Papers}},
	publisher = {Association for Computational Linguistics},
	author = {Wachsmuth, Henning and Naderi, Nona and Hou, Yufang and Bilu, Yonatan and Prabhakaran, Vinodkumar and Thijm, Tim Alberdingk and Hirst, Graeme and Stein, Benno},
	editor = {Lapata, Mirella and Blunsom, Phil and Koller, Alexander},
	month = apr,
	year = {2017},
	pages = {176--187},
}

@article{hackenburgLeversPoliticalPersuasion2025,
	title = {The levers of political persuasion with conversational artificial intelligence},
	volume = {390},
	url = {https://www.science.org/doi/10.1126/science.aea3884},
	doi = {10.1126/science.aea3884},
	number = {6777},
	urldate = {2025-12-06},
	journal = {Science},
	publisher = {American Association for the Advancement of Science},
	author = {Hackenburg, Kobi and Tappin, Ben M. and Hewitt, Luke and Saunders, Ed and Black, Sid and Lin, Hause and Fist, Catherine and Margetts, Helen and Rand, David G. and Summerfield, Christopher},
	month = dec,
	year = {2025},
	pages = {eaea3884},
}

@article{baiLLMgeneratedMessagesCan2025,
	title = {{LLM}-generated messages can persuade humans on policy issues},
	volume = {16},
	copyright = {2025 The Author(s)},
	issn = {2041-1723},
	url = {https://www.nature.com/articles/s41467-025-61345-5},
	doi = {10.1038/s41467-025-61345-5},
	language = {en},
	number = {1},
	urldate = {2025-11-20},
	journal = {Nature Communications},
	publisher = {Nature Publishing Group},
	author = {Bai, Hui and Voelkel, Jan G. and Muldowney, Shane and Eichstaedt, Johannes C. and Willer, Robb},
	month = jul,
	year = {2025},
	pages = {6037},
}

@inproceedings{gleizeAreYouConvinced2019,
	address = {Florence, Italy},
	title = {Are {You} {Convinced}? {Choosing} the {More} {Convincing} {Evidence} with a {Siamese} {Network}},
	shorttitle = {Are {You} {Convinced}?},
	url = {https://aclanthology.org/P19-1093/},
	doi = {10.18653/v1/P19-1093},
	urldate = {2025-10-29},
	booktitle = {Proceedings of the 57th {Annual} {Meeting} of the {Association} for {Computational} {Linguistics}},
	publisher = {Association for Computational Linguistics},
	author = {Gleize, Martin and Shnarch, Eyal and Choshen, Leshem and Dankin, Lena and Moshkowich, Guy and Aharonov, Ranit and Slonim, Noam},
	editor = {Korhonen, Anna and Traum, David and Màrquez, Lluís},
	month = jul,
	year = {2019},
	pages = {967--976},
}

@inproceedings{piskorskiSemEval2023Task32023,
	address = {Toronto, Canada},
	title = {{SemEval}-2023 {Task} 3: {Detecting} the {Category}, the {Framing}, and the {Persuasion} {Techniques} in {Online} {News} in a {Multi}-lingual {Setup}},
	shorttitle = {{SemEval}-2023 {Task} 3},
	url = {https://aclanthology.org/2023.semeval-1.317/},
	doi = {10.18653/v1/2023.semeval-1.317},
	urldate = {2025-10-27},
	booktitle = {Proceedings of the 17th {International} {Workshop} on {Semantic} {Evaluation} ({SemEval}-2023)},
	publisher = {Association for Computational Linguistics},
	author = {Piskorski, Jakub and Stefanovitch, Nicolas and Da San Martino, Giovanni and Nakov, Preslav},
	editor = {Ojha, Atul Kr. and Doğruöz, A. Seza and Da San Martino, Giovanni and Tayyar Madabushi, Harish and Kumar, Ritesh and Sartori, Elisa},
	month = jul,
	year = {2023},
	pages = {2343--2361},
}

@inproceedings{danescu-niculescu-mizilComputationalApproachPoliteness2013,
	address = {Sofia, Bulgaria},
	title = {A computational approach to politeness with application to social factors},
	url = {https://aclanthology.org/P13-1025/},
	urldate = {2025-10-21},
	booktitle = {Proceedings of the 51st {Annual} {Meeting} of the {Association} for {Computational} {Linguistics} ({Volume} 1: {Long} {Papers})},
	publisher = {Association for Computational Linguistics},
	author = {Danescu-Niculescu-Mizil, Cristian and Sudhof, Moritz and Jurafsky, Dan and Leskovec, Jure and Potts, Christopher},
	editor = {Schuetze, Hinrich and Fung, Pascale and Poesio, Massimo},
	month = aug,
	year = {2013},
	pages = {250--259},
}

@article{hackenburgComparingPersuasivenessRoleplaying2025,
	title = {Comparing the persuasiveness of role-playing large language models and human experts on polarized {U}.{S}. political issues},
	issn = {1435-5655},
	url = {https://doi.org/10.1007/s00146-025-02464-x},
	doi = {10.1007/s00146-025-02464-x},
	language = {en},
	urldate = {2025-10-19},
	journal = {AI \& SOCIETY},
	author = {Hackenburg, Kobi and Ibrahim, Lujain and Tappin, Ben M. and Tsakiris, Manos},
	month = jul,
	year = {2025},
}

@inproceedings{somasundaranDetectingArguingSentiment2007,
	address = {Antwerp, Belgium},
	title = {Detecting {Arguing} and {Sentiment} in {Meetings}},
	url = {https://aclanthology.org/2007.sigdial-1.5/},
	doi = {10.18653/v1/2007.sigdial-1.5},
	urldate = {2025-10-19},
	booktitle = {Proceedings of the 8th {SIGdial} {Workshop} on {Discourse} and {Dialogue}},
	publisher = {Association for Computational Linguistics},
	author = {Somasundaran, Swapna and Ruppenhofer, Josef and Wiebe, Janyce},
	editor = {Bunt, Harry and Keizer, Simon and Paek, Tim},
	month = sep,
	year = {2007},
	pages = {26--34},
}

@article{hackenburgEvaluatingPersuasiveInfluence2024,
	title = {Evaluating the persuasive influence of political microtargeting with large language models},
	volume = {121},
	url = {https://www.pnas.org/doi/full/10.1073/pnas.2403116121},
	doi = {10.1073/pnas.2403116121},
	number = {24},
	urldate = {2025-10-19},
	journal = {Proceedings of the National Academy of Sciences},
	publisher = {Proceedings of the National Academy of Sciences},
	author = {Hackenburg, Kobi and Margetts, Helen},
	month = jun,
	year = {2024},
	pages = {e2403116121},
}

@article{luuMeasuringOnlineDebaters2019,
	title = {Measuring {Online} {Debaters}' {Persuasive} {Skill} from {Text} over {Time}},
	volume = {7},
	url = {https://aclanthology.org/Q19-1031/},
	doi = {10.1162/tacl_a_00281},
	urldate = {2025-10-18},
	journal = {Transactions of the Association for Computational Linguistics},
	publisher = {MIT Press},
	author = {Luu, Kelvin and Tan, Chenhao and Smith, Noah A.},
	editor = {Lee, Lillian and Johnson, Mark and Roark, Brian and Nenkova, Ani},
	year = {2019},
	note = {Place: Cambridge, MA},
	pages = {537--550},
}

@inproceedings{vargheeseExploringSusceptibilityMeasures2020,
	address = {Cham},
	title = {Exploring {Susceptibility} {Measures} to {Persuasion}},
	isbn = {978-3-030-45712-9},
	doi = {10.1007/978-3-030-45712-9_2},
	language = {en},
	booktitle = {Persuasive {Technology}. {Designing} for {Future} {Change}},
	publisher = {Springer International Publishing},
	author = {Vargheese, John Paul and Collinson, Matthew and Masthoff, Judith},
	editor = {Gram-Hansen, Sandra Burri and Jonasen, Tanja Svarre and Midden, Cees},
	year = {2020},
	pages = {16--29},
}

@inproceedings{yangLetsMakeYour2019,
	address = {Minneapolis, Minnesota},
	title = {Let's {Make} {Your} {Request} {More} {Persuasive}: {Modeling} {Persuasive} {Strategies} via {Semi}-{Supervised} {Neural} {Nets} on {Crowdfunding} {Platforms}},
	shorttitle = {Let's {Make} {Your} {Request} {More} {Persuasive}},
	url = {https://aclanthology.org/N19-1364/},
	doi = {10.18653/v1/N19-1364},
	urldate = {2025-10-18},
	booktitle = {Proceedings of the 2019 {Conference} of the {North} {American} {Chapter} of the {Association} for {Computational} {Linguistics}: {Human} {Language} {Technologies}, {Volume} 1 ({Long} and {Short} {Papers})},
	publisher = {Association for Computational Linguistics},
	author = {Yang, Diyi and Chen, Jiaao and Yang, Zichao and Jurafsky, Dan and Hovy, Eduard},
	editor = {Burstein, Jill and Doran, Christy and Solorio, Thamar},
	month = jun,
	year = {2019},
	pages = {3620--3630},
}

@article{chenWeaklySupervisedHierarchicalModels2021,
	title = {Weakly-{Supervised} {Hierarchical} {Models} for {Predicting} {Persuasive} {Strategies} in {Good}-faith {Textual} {Requests}},
	volume = {35},
	issn = {2374-3468, 2159-5399},
	url = {https://ojs.aaai.org/index.php/AAAI/article/view/17498},
	doi = {10.1609/aaai.v35i14.17498},
	language = {en},
	number = {14},
	urldate = {2025-10-05},
	journal = {Proceedings of the AAAI Conference on Artificial Intelligence},
	author = {Chen, Jiaao and Yang, Diyi},
	month = may,
	year = {2021},
	pages = {12648--12656},
}

@article{matzPotentialGenerativeAI2024,
	title = {The potential of generative {AI} for personalized persuasion at scale},
	volume = {14},
	copyright = {2024 The Author(s)},
	issn = {2045-2322},
	url = {https://www.nature.com/articles/s41598-024-53755-0},
	doi = {10.1038/s41598-024-53755-0},
	language = {en},
	number = {1},
	urldate = {2025-09-30},
	journal = {Scientific Reports},
	publisher = {Nature Publishing Group},
	author = {Matz, S. C. and Teeny, J. D. and Vaid, S. S. and Peters, H. and Harari, G. M. and Cerf, M.},
	month = feb,
	year = {2024},
	pages = {4692},
}

@inproceedings{ghoshCoarsegrainedArgumentationFeatures2016,
	address = {Berlin, Germany},
	title = {Coarse-grained {Argumentation} {Features} for {Scoring} {Persuasive} {Essays}},
	url = {https://aclanthology.org/P16-2089/},
	doi = {10.18653/v1/P16-2089},
	urldate = {2025-09-30},
	booktitle = {Proceedings of the 54th {Annual} {Meeting} of the {Association} for {Computational} {Linguistics} ({Volume} 2: {Short} {Papers})},
	publisher = {Association for Computational Linguistics},
	author = {Ghosh, Debanjan and Khanam, Aquila and Han, Yubo and Muresan, Smaranda},
	editor = {Erk, Katrin and Smith, Noah A.},
	month = aug,
	year = {2016},
	pages = {549--554},
}

@article{dashPersuasivePotentialAIparaphrased2025,
	title = {The persuasive potential of {AI}-paraphrased information at scale},
	volume = {4},
	issn = {2752-6542},
	url = {https://doi.org/10.1093/pnasnexus/pgaf207},
	doi = {10.1093/pnasnexus/pgaf207},
	number = {7},
	urldate = {2025-09-29},
	journal = {PNAS Nexus},
	author = {Dash, Saloni and Xu, Yiwei and Jalbert, Madeline and Spiro, Emma S},
	month = jul,
	year = {2025},
	pages = {pgaf207},
}

@article{salviConversationalPersuasivenessGPT42025,
	title = {On the conversational persuasiveness of {GPT}-4},
	volume = {9},
	copyright = {2025 The Author(s)},
	issn = {2397-3374},
	url = {https://www.nature.com/articles/s41562-025-02194-6},
	doi = {10.1038/s41562-025-02194-6},
	language = {en},
	number = {8},
	urldate = {2025-09-29},
	journal = {Nature Human Behaviour},
	publisher = {Nature Publishing Group},
	author = {Salvi, Francesco and Horta Ribeiro, Manoel and Gallotti, Riccardo and West, Robert},
	month = aug,
	year = {2025},
	pages = {1645--1653},
}

@article{montiLanguageOpinionChange2022,
	title = {The language of opinion change on social media under the lens of communicative action},
	volume = {12},
	copyright = {2022 The Author(s)},
	issn = {2045-2322},
	url = {https://www.nature.com/articles/s41598-022-21720-4},
	doi = {10.1038/s41598-022-21720-4},
	language = {en},
	number = {1},
	urldate = {2025-09-29},
	journal = {Scientific Reports},
	publisher = {Nature Publishing Group},
	author = {Monti, Corrado and Aiello, Luca Maria and De Francisci Morales, Gianmarco and Bonchi, Francesco},
	month = oct,
	year = {2022},
	pages = {17920},
}

@inproceedings{toledoAutomaticArgumentQuality2019,
	address = {Hong Kong, China},
	title = {Automatic {Argument} {Quality} {Assessment} - {New} {Datasets} and {Methods}},
	url = {https://aclanthology.org/D19-1564/},
	doi = {10.18653/v1/D19-1564},
	urldate = {2025-09-29},
	booktitle = {Proceedings of the 2019 {Conference} on {Empirical} {Methods} in {Natural} {Language} {Processing} and the 9th {International} {Joint} {Conference} on {Natural} {Language} {Processing} ({EMNLP}-{IJCNLP})},
	publisher = {Association for Computational Linguistics},
	author = {Toledo, Assaf and Gretz, Shai and Cohen-Karlik, Edo and Friedman, Roni and Venezian, Elad and Lahav, Dan and Jacovi, Michal and Aharonov, Ranit and Slonim, Noam},
	editor = {Inui, Kentaro and Jiang, Jing and Ng, Vincent and Wan, Xiaojun},
	month = nov,
	year = {2019},
	pages = {5625--5635},
}

@inproceedings{mitraLanguageThatGets2014,
	address = {New York, NY, USA},
	series = {{CSCW} '14},
	title = {The language that gets people to give: phrases that predict success on kickstarter},
	isbn = {978-1-4503-2540-0},
	shorttitle = {The language that gets people to give},
	url = {https://doi.org/10.1145/2531602.2531656},
	doi = {10.1145/2531602.2531656},
	urldate = {2025-09-23},
	booktitle = {Proceedings of the 17th {ACM} conference on {Computer} supported cooperative work \& social computing},
	publisher = {Association for Computing Machinery},
	author = {Mitra, Tanushree and Gilbert, Eric},
	month = feb,
	year = {2014},
	pages = {49--61},
}

@article{althoffHowAskFavor2014,
	title = {How to {Ask} for a {Favor}: {A} {Case} {Study} on the {Success} of {Altruistic} {Requests}},
	volume = {8},
	issn = {2334-0770, 2162-3449},
	shorttitle = {How to {Ask} for a {Favor}},
	url = {https://ojs.aaai.org/index.php/ICWSM/article/view/14547},
	doi = {10.1609/icwsm.v8i1.14547},
	language = {en},
	number = {1},
	urldate = {2025-09-23},
	journal = {Proceedings of the International AAAI Conference on Web and Social Media},
	author = {Althoff, Tim and Danescu-Niculescu-Mizil, Cristian and Jurafsky, Dan},
	month = may,
	year = {2014},
	pages = {12--21},
}

@article{susmannIndependentEffectsSource2023,
	title = {The independent effects of source expertise and trustworthiness on retraction believability: {The} moderating role of vested interest},
	volume = {51},
	issn = {1532-5946},
	shorttitle = {The independent effects of source expertise and trustworthiness on retraction believability},
	doi = {10.3758/s13421-022-01374-3},
	language = {eng},
	number = {4},
	journal = {Memory \& Cognition},
	author = {Susmann, Mark W. and Wegener, Duane T.},
	month = may,
	year = {2023},
	pages = {845--861},
}

@article{allenMetaanalysisComparingPersuasiveness1991,
	title = {Meta‐analysis comparing the persuasiveness of one‐sided and two‐sided messages},
	volume = {55},
	issn = {0193-6700},
	url = {http://www.tandfonline.com/doi/abs/10.1080/10570319109374395},
	doi = {10.1080/10570319109374395},
	language = {en},
	number = {4},
	urldate = {2025-09-22},
	journal = {Western Journal of Speech Communication},
	author = {Allen, Mike},
	month = dec,
	year = {1991},
	pages = {390--404},
}

@article{wangWinningMeritsJoint2017,
	title = {Winning on the {Merits}: {The} {Joint} {Effects} of {Content} and {Style} on {Debate} {Outcomes}},
	volume = {5},
	shorttitle = {Winning on the {Merits}},
	url = {https://aclanthology.org/Q17-1016/},
	doi = {10.1162/tacl_a_00057},
	urldate = {2025-09-22},
	journal = {Transactions of the Association for Computational Linguistics},
	publisher = {MIT Press},
	author = {Wang, Lu and Beauchamp, Nick and Shugars, Sarah and Qin, Kechen},
	editor = {Lee, Lillian and Johnson, Mark and Toutanova, Kristina},
	year = {2017},
	note = {Place: Cambridge, MA},
	pages = {219--232},
}

@inproceedings{tanWinningArgumentsInteraction2016,
	address = {Republic and Canton of Geneva, CHE},
	series = {{WWW} '16},
	title = {Winning {Arguments}: {Interaction} {Dynamics} and {Persuasion} {Strategies} in {Good}-faith {Online} {Discussions}},
	isbn = {978-1-4503-4143-1},
	shorttitle = {Winning {Arguments}},
	url = {https://doi.org/10.1145/2872427.2883081},
	doi = {10.1145/2872427.2883081},
	urldate = {2025-09-22},
	booktitle = {Proceedings of the 25th {International} {Conference} on {World} {Wide} {Web}},
	publisher = {International World Wide Web Conferences Steering Committee},
	author = {Tan, Chenhao and Niculae, Vlad and Danescu-Niculescu-Mizil, Cristian and Lee, Lillian},
	month = apr,
	year = {2016},
	pages = {613--624},
}

@inproceedings{hideyAnalyzingSemanticTypes2017,
	address = {Copenhagen, Denmark},
	title = {Analyzing the {Semantic} {Types} of {Claims} and {Premises} in an {Online} {Persuasive} {Forum}},
	url = {https://aclanthology.org/W17-5102/},
	doi = {10.18653/v1/W17-5102},
	urldate = {2025-09-22},
	booktitle = {Proceedings of the 4th {Workshop} on {Argument} {Mining}},
	publisher = {Association for Computational Linguistics},
	author = {Hidey, Christopher and Musi, Elena and Hwang, Alyssa and Muresan, Smaranda and McKeown, Kathy},
	editor = {Habernal, Ivan and Gurevych, Iryna and Ashley, Kevin and Cardie, Claire and Green, Nancy and Litman, Diane and Petasis, Georgios and Reed, Chris and Slonim, Noam and Walker, Vern},
	month = sep,
	year = {2017},
	pages = {11--21},
}

@inproceedings{wangPersuasionGoodPersonalized2019,
	address = {Florence, Italy},
	title = {Persuasion for {Good}: {Towards} a {Personalized} {Persuasive} {Dialogue} {System} for {Social} {Good}},
	shorttitle = {Persuasion for {Good}},
	url = {https://aclanthology.org/P19-1566/},
	doi = {10.18653/v1/P19-1566},
	urldate = {2025-09-22},
	booktitle = {Proceedings of the 57th {Annual} {Meeting} of the {Association} for {Computational} {Linguistics}},
	publisher = {Association for Computational Linguistics},
	author = {Wang, Xuewei and Shi, Weiyan and Kim, Richard and Oh, Yoojung and Yang, Sijia and Zhang, Jingwen and Yu, Zhou},
	editor = {Korhonen, Anna and Traum, David and Màrquez, Lluís},
	month = jul,
	year = {2019},
	pages = {5635--5649},
}

@inproceedings{joshiArgAnalysis35KLargescaleDataset2023,
	address = {Toronto, Canada},
	title = {{ArgAnalysis35K} : {A} large-scale dataset for {Argument} {Quality} {Analysis}},
	shorttitle = {{ArgAnalysis35K}},
	url = {https://aclanthology.org/2023.acl-long.778/},
	doi = {10.18653/v1/2023.acl-long.778},
	urldate = {2025-09-22},
	booktitle = {Proceedings of the 61st {Annual} {Meeting} of the {Association} for {Computational} {Linguistics} ({Volume} 1: {Long} {Papers})},
	publisher = {Association for Computational Linguistics},
	author = {Joshi, Omkar and Pitre, Priya and Haribhakta, Yashodhara},
	editor = {Rogers, Anna and Boyd-Graber, Jordan and Okazaki, Naoaki},
	month = jul,
	year = {2023},
	pages = {13916--13931},
}

@inproceedings{habernalWhichArgumentMore2016,
	address = {Berlin, Germany},
	title = {Which argument is more convincing? {Analyzing} and predicting convincingness of {Web} arguments using bidirectional {LSTM}},
	shorttitle = {Which argument is more convincing?},
	url = {https://aclanthology.org/P16-1150/},
	doi = {10.18653/v1/P16-1150},
	urldate = {2025-09-16},
	booktitle = {Proceedings of the 54th {Annual} {Meeting} of the {Association} for {Computational} {Linguistics} ({Volume} 1: {Long} {Papers})},
	publisher = {Association for Computational Linguistics},
	author = {Habernal, Ivan and Gurevych, Iryna},
	editor = {Erk, Katrin and Smith, Noah A.},
	month = aug,
	year = {2016},
	pages = {1589--1599},
}

@inproceedings{modzelewskiPCoTPersuasionAugmentedChain2025,
	address = {Vienna, Austria},
	title = {{PCoT}: {Persuasion}-{Augmented} {Chain} of {Thought} for {Detecting} {Fake} {News} and {Social} {Media} {Disinformation}},
	isbn = {979-8-89176-251-0},
	shorttitle = {{PCoT}},
	url = {https://aclanthology.org/2025.acl-long.1215/},
	doi = {10.18653/v1/2025.acl-long.1215},
	urldate = {2025-09-15},
	booktitle = {Proceedings of the 63rd {Annual} {Meeting} of the {Association} for {Computational} {Linguistics} ({Volume} 1: {Long} {Papers})},
	publisher = {Association for Computational Linguistics},
	author = {Modzelewski, Arkadiusz and Sosnowski, Witold and Labruna, Tiziano and Wierzbicki, Adam and Da San Martino, Giovanni},
	editor = {Che, Wanxiang and Nabende, Joyce and Shutova, Ekaterina and Pilehvar, Mohammad Taher},
	month = jul,
	year = {2025},
	pages = {24959--24983},
}

@inproceedings{minFActScoreFinegrainedAtomic2023,
	address = {Singapore},
	title = {{FActScore}: {Fine}-grained {Atomic} {Evaluation} of {Factual} {Precision} in {Long} {Form} {Text} {Generation}},
	shorttitle = {{FActScore}},
	url = {https://aclanthology.org/2023.emnlp-main.741/},
	doi = {10.18653/v1/2023.emnlp-main.741},
	urldate = {2025-09-15},
	booktitle = {Proceedings of the 2023 {Conference} on {Empirical} {Methods} in {Natural} {Language} {Processing}},
	publisher = {Association for Computational Linguistics},
	author = {Min, Sewon and Krishna, Kalpesh and Lyu, Xinxi and Lewis, Mike and Yih, Wen-tau and Koh, Pang and Iyyer, Mohit and Zettlemoyer, Luke and Hajishirzi, Hannaneh},
	editor = {Bouamor, Houda and Pino, Juan and Bali, Kalika},
	month = dec,
	year = {2023},
	pages = {12076--12100},
}

@article{costelloDurablyReducingConspiracy2024,
	title = {Durably reducing conspiracy beliefs through dialogues with {AI}},
	volume = {385},
	url = {https://www.science.org/doi/10.1126/science.adq1814},
	doi = {10.1126/science.adq1814},
	number = {6714},
	urldate = {2025-09-15},
	journal = {Science},
	publisher = {American Association for the Advancement of Science},
	author = {Costello, Thomas H. and Pennycook, Gordon and Rand, David G.},
	month = sep,
	year = {2024},
	pages = {eadq1814},
}

@article{augensteinFactualityChallengesEra2024,
	title = {Factuality challenges in the era of large language models and opportunities for fact-checking},
	volume = {6},
	copyright = {2024 Springer Nature Limited},
	issn = {2522-5839},
	url = {https://www.nature.com/articles/s42256-024-00881-z},
	doi = {10.1038/s42256-024-00881-z},
	language = {en},
	number = {8},
	urldate = {2025-09-15},
	journal = {Nature Machine Intelligence},
	publisher = {Nature Publishing Group},
	author = {Augenstein, Isabelle and Baldwin, Timothy and Cha, Meeyoung and Chakraborty, Tanmoy and Ciampaglia, Giovanni Luca and Corney, David and DiResta, Renee and Ferrara, Emilio and Hale, Scott and Halevy, Alon and Hovy, Eduard and Ji, Heng and Menczer, Filippo and Miguez, Ruben and Nakov, Preslav and Scheufele, Dietram and Sharma, Shivam and Zagni, Giovanni},
	month = aug,
	year = {2024},
	pages = {852--863},
}
\appendix
\renewcommand{\thetable}{A\arabic{table}}
\setcounter{table}{0}

\renewcommand{\thefigure}{A\arabic{figure}}
\setcounter{figure}{0}
\section{Persuasion Strategies in the Persuasion Index}
\label{appendix:strategy_table}

Table~\ref{tab:pi_categories} provides the canonical reference for all 15 PI dimensions cited throughout the paper, organized under Aristotle's three modes of appeal, including \appeal{Logos}, \appeal{Ethos}, and \appeal{Pathos}. Each row lists the dimension's working definition with an illustrative example, the prior computational work informing its operationalization, and the theoretical literature grounding the underlying persuasive mechanism.

Table~\ref{tab:pi_methods_one_cue} maps each PI dimension to its constituent subfeatures and the specific linguistic cues used to detect them. The final column gives the regex, lexicon, or rule that produces each measurement; a subfeature score aggregates all of its assigned cues, normalized to per-token density unless the cue is marked \texttt{(binary)}, in which case the feature is computed as a 0/1 presence indicator per document. Together the two tables provide a complete trace from each Aristotelian appeal down to the operational rule that yields a feature value.

 \onecolumn

\begingroup
\footnotesize
\renewcommand{\arraystretch}{1.08} 

\begin{longtable}{p{0.10\textwidth} p{0.3\textwidth} p{0.3\textwidth} p{0.3\textwidth}}

\hline
\textbf{Strategy} & \textbf{Definition and Example} & \textbf{Computational Work} & \textbf{Theory Support} \\
\hline
\endfirsthead

\multicolumn{4}{c}{\tablename\ \thetable{} -- \textit{continued from previous page}} \\
\hline
\textbf{Strategy} & \textbf{Definition and Example} & \textbf{Computational Work} & \textbf{Theory Support} \\
\hline\hline
\endhead

\hline
\multicolumn{4}{r}{\textit{Continued on next page}} \\
\endfoot

\endlastfoot

Evidence &
Quantifies linguistic markers of statistical, attributed, and entity-based evidential support.
\emph{``Official statistics show unemployment fell by 20\% this year.''} &
\citealt{gretzLargescaleDatasetArgument2020} \newline
\citealt{habernalWhichArgumentMore2016} \newline
\citealt{toledoAutomaticArgumentQuality2019} \newline
\citealt{minFActScoreFinegrainedAtomic2023} & 
\citealt{zebregsDifferentialImpactStatistical2015} \newline
\citealt{allenComparingPersuasivenessNarrative1997} \newline
\citealt{kimTestingAdditiveModel2012} \newline
\citealt{bigsbyExemplificationTheoryReview2019} 
\\
\hline

Logic / Cohesion &
Measures reasoning structure and discourse connectivity that reduce cognitive load in processing.
\emph{``Because the rate is fixed, your monthly payment will not increase.''} &
\citealt{hideyPersuasiveInfluenceDetection2018} \newline
\citealt{wangWinningMeritsJoint2017} &  
\citealt{pettyElaborationLikelihoodModel1986} \newline
\citealt{chaikenHeuristicSystematicInformation1980} \newline
\citealt{kaakinenInfluenceTextCohesion2011} \newline
\citealt{sebaldaAnalyzingCoherenceCohesion2025} \newline
\citealt{decheneTruthTruthMetaanalytic2010}\\
\hline

Argumentation &
Captures structural explicitness of claims, premises, and argumentative intensity within a text.
\emph{``Our city should expand transit because commute times doubled.''} &
\citealt{somasundaranDetectingArguingSentiment2007} \newline
\citealt{hideyAnalyzingSemanticTypes2017} \newline
\citealt{gretzLargescaleDatasetArgument2020} \newline
\citealt{joshiArgAnalysis35KLargescaleDataset2023} \newline
\citealt{wachsmuthComputationalArgumentationQuality2017} \newline
\citealt{stabIdentifyingArgumentativeDiscourse2014} &
\citealt{lukinArgumentStrengthEye2017} \newline
\citealt{blumenauVariablePersuasivenessPolitical2024} \newline
\citealt{okeefeStandpointExplicitnessPersuasive1997a} \newline
\citealt{hovlandExperimentalComparisonConclusiondrawing1952} \newline
\citealt{pettyEffectsInvolvementResponses1984} \newline
\citealt{dillardLanguageStylePersuasion2014} \newline
\citealt{zhaoMeasurePerceivedArgument2011}
\\
\hline

Specificity &
Analyzes linguistic concreteness and psychological proximity to increase perceived truth.
\emph{``I contributed \$25 last Friday so the shelter can vaccinate three more puppies.''} &
\citealt{althoffHowAskFavor2014} \newline
\citealt{mitraLanguageThatGets2014} \newline
\citealt{ghoshCoarsegrainedArgumentationFeatures2016} \newline
\citealt{carlileGiveMeMore2018} \newline
\citealt{brysbaertConcretenessRatings402014} \newline
\citealt{rietscheSpecificityHelpfulnessPeertoPeer2022} \newline
\citealt{murakiConcretenessRatings620002023} & 
\citealt{tropeConstrualLevelTheoryPsychological2010} \newline
\citealt{hansenTruthLanguageTruth2010} \newline
\citealt{dillardLanguageStylePersuasion2014} \newline
\citealt{ciukMoralConvictionEmotion2020}  \newline
\citealt{fujitaInfluencingAttitudesDistant2008} \newline
\citealt{cruzSecondPersonPronouns2017} \newline
\citealt{packardHowConcreteLanguage2021}
\\
\hline

Opponent's View &
Detects concession-refutation sequences that strategically acknowledge and rebut counterarguments.
\emph{``Some argue rent control reduces supply, but recent data disagree.''} &
\citealt{tanWinningArgumentsInteraction2016} \newline
\citealt{wangWinningMeritsJoint2017} \newline
\citealt{luuMeasuringOnlineDebaters2019} & 
\citealt{okeefeHowHandleOpposing1999} \newline
\citealt{allenMetaanalysisComparingPersuasiveness1991} \\
\hline
\hline

Authority / Credibility &
Captures linguistic signals of expertise and institutional trust that function as credibility heuristics.
\emph{``Researchers at the national institute recommend this booster.''} &
\citealt{susmannIndependentEffectsSource2023} \newline
\citealt{vargheeseExploringSusceptibilityMeasures2020} \newline
\citealt{SAGEHandbookPersuasion2025} \newline
\citealt{farkasCoNLL2010SharedTask2010} & 
\citealt{hovlandInfluenceSourceCredibility1951} \newline
\citealt{dillardLanguageStylePersuasion2014} \newline
\citealt{songSourceEffectsPsychological2018} \newline
\citealt{pornpitakpanPersuasivenessSourceCredibility2004} \newline
\citealt{bulleePersuasionSecurityAwareness2015} \newline
\citealt{smithEffectsPowerfulPowerless1998} \newline
\citealt{mccroskeyEthosCredibilityConstruct1981} \newline
\citealt{mccroskeyGoodwillReexaminationConstruct1999}
 \\
\hline

Politeness &
Measures face-saving language that reduces psychological reactance and mitigates persuasive imposition.
\emph{``I see your point, but may I suggest an alternative?''} &
\citealt{danescu-niculescu-mizilComputationalApproachPoliteness2013} \newline
\citealt{tanWinningArgumentsInteraction2016} \newline
\citealt{luuMeasuringOnlineDebaters2019} \newline
\citealt{farkasCoNLL2010SharedTask2010} & 
\citealt{brownPolitenessUniversalsLanguage1987} \newline
\citealt{carpenterMetaAnalysisEffectivenessYou2013} \newline
\citealt{rainsNaturePsychologicalReactance2013}
\\
\hline

Commitment &
Signals the communicator's resolve and prior investment to establish a consistent, credible persona.
\emph{``I already pledged to bike to work, so I support this proposal.''} &
\citealt{vargheeseExploringSusceptibilityMeasures2020} \newline
\citealt{yangLetsMakeYour2019} \newline
\citealt{wangPersuasionGoodPersonalized2019} &  
\citealt{burgerFootintheDoorComplianceProcedure1999} \newline
\citealt{pallakCommitmentEnergyConservation1980} \newline
\citealt{chenFrameworkModeratorsSocial2024}
\\
\hline

Surface / Style &
Captures perceptual fluency and surface-level competence cues that serve as heuristics for truth.
\emph{``We will act, listen, and adapt.''} &
\citealt{ghoshCoarsegrainedArgumentationFeatures2016} \newline
\citealt{wangWinningMeritsJoint2017} \newline
\citealt{gretzLargescaleDatasetArgument2020} \newline
\citealt{farkasCoNLL2010SharedTask2010} &
\citealt{kempPeriodReallyPissed2025} \newline
\citealt{reberEffectsPerceptualFluency1999}
\\
\hline

Sentiment &
Measures affective intensity through polarity, discrete emotional categories, and dimensional affect scores.
\emph{``It is heartbreaking that so many families cannot afford care.''} &
\citealt{somasundaranDetectingArguingSentiment2007} \newline
\citealt{montiLanguageOpinionChange2022} \newline
\citealt{breumPersuasivePowerLarge2024} \newline
\citealt{mohammadWordAffectIntensities2018} & 
\citealt{dillardLanguageStylePersuasion2014} \newline
\citealt{dillardDiscreteEmotionsPersuasion} \newline
\citealt{witteMetaanalysisFearAppeals2000} \newline
\citealt{hornikQuantitativeEvaluationPersuasive2016}
\\
\hline

Impact &
Leverages Prospect Theory by framing projected gains, losses, and future consequences.
\emph{``Your donation funds three counseling sessions.''} &
\citealt{montiLanguageOpinionChange2022} \newline
\citealt{costelloDurablyReducingConspiracy2024} \newline
\citealt{matzPotentialGenerativeAI2024} \newline
\citealt{baiLLMgeneratedMessagesCan2025}. & 
\citealt{okeefeRelativePersuasivenessGainframed2007} \newline
\citealt{kimMetaanalysisGainLoss2025} \newline
\citealt{witteMetaanalysisFearAppeals2000} \newline
\citealt{rothmanStrategicUseGain2006} \newline
\citealt{dillardLanguageStylePersuasion2014} \newline
\citealt{carpenterMetaAnalysisEffectivenessYou2013} \newline
\citealt{gallagherHealthMessageFraming2012}
 \\
\hline

Engagement &
Promotes narrative transportation via direct address, self-referencing, and vivid imagery.
\emph{``Have you ever wondered where your taxes go?''} &
\citealt{tanWinningArgumentsInteraction2016} \newline
\citealt{luuMeasuringOnlineDebaters2019} & 
\citealt{dillardLanguageStylePersuasion2014} \newline
\citealt{greenRoleTransportationPersuasiveness2000} \newline
\citealt{burgerFootintheDoorComplianceProcedure1999} \newline
\citealt{shenHEARTfeltNarrativesTracing2024} \newline
\citealt{burnkrantEffectsSelfreferencingPersuasion1995} \newline
\citealt{blankenshipRhetoricalQuestionUse2006} \newline
\citealt{zebregsDifferentialImpactStatistical2015}

 \\
\hline

Reciprocity &
Triggers the norm of social exchange by signaling concession or indebtedness to prompt compliance.
\emph{``I have been sharing my notes all semester---could you sign this petition?''} &
\citealt{vargheeseExploringSusceptibilityMeasures2020} \newline
\citealt{althoffHowAskFavor2014} \newline
\citealt{mitraLanguageThatGets2014} \newline
\citealt{yangLetsMakeYour2019} \newline  
\citealt{okeefeDoorintheFaceInfluenceStrategy1998} \newline
\citealt{burgerIncreasingComplianceImproving1986}\\
\hline

Scarcity / Urgency &
Heightens perceived value by invoking time limits or limited availability to intensify decision pressure.
\emph{``Only 24 hours remain before matching funds expire.''} &
\citealt{vargheeseExploringSusceptibilityMeasures2020} \newline
\citealt{SAGEHandbookPersuasion2025} &  
\citealt{ladeiraMetaanalysisEffectsProduct2023} \newline
\citealt{grantMultipleRolesScarcity2014}\\
\hline

Propaganda &
Detects identity-based heuristics and logical distortions that secure compliance regardless of logical depth.
\emph{``Everyone who cares about freedom is joining.''} &
\citealt{dimitrovSemEval2021Task62021} \newline
\citealt{dasanmartinoSemEval2020Task112020} \newline
\citealt{piskorskiSemEval2023Task32023} \newline
\citealt{modzelewskiPCoTPersuasionAugmentedChain2025} &  
\citealt{witteMetaanalysisFearAppeals2000} \newline
\citealt{shenDoesAuthoritarianPropaganda2025} \newline
\citealt{karasmanPropagandaMechanismManipulation} \\
\hline

\caption{Fifteen persuasion dimensions in the Persuasion Index (PI), organized under the Aristotelian triad. \appeal{Logos} (reasoning and evidence): Evidence, Logic/Cohesion, Argumentation, Specificity, Opponent's View. \appeal{Ethos} (credibility and character): Authority/Credibility, Politeness, Commitment, Style. \appeal{Pathos} (emotion and compliance): Sentiment, Impact, Engagement, Reciprocity, Scarcity/Urgency, Propaganda. Each row lists the dimension definition with an illustrative example, prior computational work that informs the operationalization, and the theoretical literature grounding the persuasive mechanism.}
\label{tab:pi_categories}
\end{longtable}
\endgroup

\providecommand{\variable}[1]{\texttt{\detokenize{#1}}}

\onecolumn

\begingroup
\footnotesize
\setlength{\tabcolsep}{3pt}
\renewcommand{\arraystretch}{1.15}

\begin{longtable}{
@{}
>{\raggedright\arraybackslash}p{0.17\textwidth}
>{\raggedright\arraybackslash}p{0.19\textwidth}
>{\raggedright\arraybackslash}p{0.25\textwidth}
>{\raggedright\arraybackslash}p{0.34\textwidth}
@{}
}

\toprule
\textbf{Dimension} &
\textbf{Subfeature} &
\textbf{Linguistic cue} &
\textbf{Detector / representative examples} \\
\midrule
\endfirsthead

\multicolumn{4}{c}{
\tablename\ \thetable{} -- \textit{continued from previous page}
} \\
\toprule
\textbf{Dimension} &
\textbf{Subfeature} &
\textbf{Linguistic cue} &
\textbf{Detector / representative examples} \\
\midrule
\endhead

\midrule
\multicolumn{4}{r}{\textit{Continued on next page}} \\
\endfoot

\endlastfoot


\multicolumn{4}{@{}l}{
\textbf{\large Logos: Reasoning and Evidence}
} \\
\specialrule{0.5pt}{2pt}{3pt}


\textbf{Evidence} &
Statistical &
Numeric expressions &
Regex \variable{RE_NUMBER}; e.g., 42, 3.5, 20\% \\
&
Statistical &
Percentage expressions &
Regex \variable{RE_PERCENT}; e.g., 25\%, 25 percent \\
&
Statistical &
Measurement and statistical units &
Lexicon \variable{EVI_UNITS}; e.g., kg, percent, million \\

\cmidrule{2-4}

&
Attribution &
APA-style citations &
Regex \variable{RE_APA} (\texttt{binary}); e.g., Smith (2024) \\
&
Attribution &
Bracketed numeric citations &
Regex \variable{RE_NUM_CIT} (\texttt{binary}); e.g., [1], [1, 2] \\
&
Attribution &
URLs and web references &
Regex \variable{RE_URL} (\texttt{binary}); HTTP/HTTPS and WWW links \\
&
Attribution &
Source-attribution phrases &
Lexicon \variable{AUTHORITY_PHRASE} (\texttt{binary});
e.g., \textit{according to}, \textit{studies show}, \textit{data from} \\

\cmidrule{2-4}

&
Named entities &
Named entities as evidential anchors &
spaCy NER; PERSON, ORG, and GPE entities \\


\midrule

\textbf{Specificity} &
Psychological nearness &
Explicit date and time cues &
Regex \variable{RE_DATE_MD} and \variable{RE_TIME_SPECIFIC}
(\texttt{binary}); e.g., Jan.\ 1, 2020; 10:30 PM \\
&
Psychological nearness &
Psychologically near anchors &
Lexicon \variable{SPEC_PSYCH_NEAR} (\texttt{binary});
e.g., \textit{minute}, \textit{hour}, \textit{day} \\
&
Psychological nearness &
Psychologically distant anchors &
Lexicon \variable{SPEC_PSYCH_FAR} (\texttt{binary});
e.g., \textit{year}, \textit{decade}, \textit{century} \\
&
Psychological nearness &
Specificity anchors &
Lexicon \variable{SPEC_ANCHORS} (\texttt{binary});
e.g., \textit{specifically}, \textit{namely}, \textit{for instance} \\
&
Psychological nearness &
Vague quantifiers and adverbs &
Lexicon \variable{SPEC_VAGUE} (\texttt{binary});
e.g., \textit{many}, \textit{some}, \textit{most} \\

\cmidrule{2-4}

&
Lexical concreteness &
Single-word concreteness &
Brysbaert et al.\ concreteness ratings \\
&
Lexical concreteness &
Multiword concreteness &
Muraki et al.\ multiword-expression concreteness ratings \\

\cmidrule{2-4}

&
Interactional immediacy &
Second-person address &
LIWC \textit{You} category \\
&
Interactional immediacy &
Person names near second-person address &
spaCy PERSON proximity rule with \textit{you}/\textit{your} \\


\midrule

\textbf{Logic/Cohesion} &
Structural reasoning &
Causal connectives &
Lexicon \variable{LOGIC_CAUSAL};
e.g., \textit{because}, \textit{since}, \textit{as a result} \\
&
Structural reasoning &
Inferential cues &
Lexicon \variable{LOGIC_INFERENCE};
e.g., \textit{implies}, \textit{suggests}, \textit{indicates} \\
&
Structural reasoning &
Referential cohesion cues &
Lexicon \variable{LOGIC_REFERENCE};
e.g., \textit{this}, \textit{that}, \textit{these} \\
&
Structural reasoning &
If--then structures &
Regex \variable{IF_THEN}; explicit \textit{if ... then ...} constructions \\

\cmidrule{2-4}

&
Discourse cohesion &
Contrastive discourse markers &
Lexicon \variable{LOGIC_CONTRAST};
e.g., \textit{however}, \textit{although}, \textit{though} \\
&
Discourse cohesion &
Additive and sequencing markers &
Lexicon \variable{LOGIC_ADDITIVE};
e.g., \textit{moreover}, \textit{furthermore}, \textit{in addition} \\
&
Discourse cohesion &
Repeated non-trivial content words &
Rule detecting repeated content words of four or more characters \\


\midrule

\textbf{Argumentation} &
Conclusion explicitness &
Claim markers &
Lexicon \variable{ARG_CLAIM};
e.g., \textit{should}, \textit{must}, \textit{need to} \\
&
Conclusion explicitness &
Conclusion markers &
Lexicon \variable{ARG_CONCLUSION};
e.g., \textit{in conclusion}, \textit{to sum up}, \textit{therefore} \\

\cmidrule{2-4}

&
Premise density &
Premise and support markers &
Lexicon \variable{ARG_PREMISE} (\texttt{binary});
e.g., \textit{because}, \textit{since}, \textit{given that} \\

\cmidrule{2-4}

&
Quantity intensity &
Argumentative sentences &
Sentence-level rule combining claim, conclusion, and premise markers \\

\cmidrule{2-4}

&
Style sophistication &
Definition cues &
Lexicon \variable{ARG_STYLE.definition};
e.g., \textit{means}, \textit{refers to}, \textit{defined as} \\
&
Style sophistication &
Analogy cues &
Lexicon \variable{ARG_STYLE.analogy};
e.g., \textit{like}, \textit{as if}, \textit{as though} \\
&
Style sophistication &
Example cues &
Lexicon \variable{ARG_STYLE.example};
e.g., \textit{for example}, \textit{for instance}, \textit{such as} \\
&
Style sophistication &
Cause--effect cues &
Lexicon \variable{ARG_STYLE.cause_effect};
e.g., \textit{because}, \textit{therefore}, \textit{thus} \\
&
Style sophistication &
Counterargument cues &
Lexicon \variable{ARG_STYLE.counter};
e.g., \textit{however}, \textit{although}, \textit{though} \\
&
Style sophistication &
Concession--refutation cues &
Lexicon \variable{ARG_STYLE.concession_refute};
e.g., \textit{I agree that ... but ...} \\
&
Style sophistication &
Conditional cues &
Lexicon \variable{ARG_STYLE.conditional};
e.g., \textit{if}, \textit{unless}, \textit{provided that} \\


\midrule

\textbf{Opponent's View} &
Acknowledge &
Opponent-view acknowledgment phrases &
Lexicon \variable{OPP_VIEW} (\texttt{binary});
e.g., \textit{I understand your point}, \textit{I see your point},
\textit{I agree that} \\
&
Acknowledge &
Soft acknowledgment frames &
Regex \variable{RE_SOFT_ACK} (\texttt{binary});
e.g., \textit{current approach}, \textit{existing approach},
\textit{limitations with the current approach} \\

\cmidrule{2-4}

&
Refutation strength &
Acknowledgment followed by contrastive refutation &
Sequential rule combining \variable{OPP_VIEW} with
\variable{LOGIC_CONTRAST};
e.g., \textit{I see your point ... however ...} \\


\specialrule{0.8pt}{5pt}{2pt}
\multicolumn{4}{@{}l}{
\textbf{\large Ethos: Credibility and Character}
} \\
\specialrule{0.5pt}{2pt}{3pt}


\textbf{Authority/Credibility} &
Titles &
Professional and academic titles &
Regex \variable{RE_TITLES};
e.g., \textit{Dr.}, \textit{Professor}, \textit{PhD} \\

\cmidrule{2-4}

&
Organizations &
Institutional authority names &
Regex \variable{RE_AUTH_ORG};
e.g., \textit{World Health Organization}, \textit{Harvard University},
\textit{CDC} \\
&
Organizations &
Organization named entities &
spaCy ORG named-entity recognition \\

\cmidrule{2-4}

&
Phrases &
Authority and source phrases &
Lexicon \variable{AUTHORITY_PHRASE} (\texttt{binary});
e.g., \textit{according to}, \textit{studies show},
\textit{research indicates} \\

\cmidrule{2-4}

&
Consensus &
Consensus and majority cues &
Regex \variable{RE_CONSENSUS} (\texttt{binary});
e.g., \textit{majority of}, \textit{most people},
\textit{nearly everyone} \\

\cmidrule{2-4}

&
Speech power &
Hedges as powerless-speech cues &
Lexicon \variable{HEDGES};
e.g., \textit{may}, \textit{might}, \textit{can} \\
&
Speech power &
Hesitation and filler cues &
Lexicon \variable{HESITATIONS};
e.g., \textit{um}, \textit{uh}, \textit{err} \\
&
Speech power &
Tag questions &
Lexicon \variable{AUT_TAG_QUESTION};
e.g., \textit{right?}, \textit{don't you think?}, \textit{isn't it?} \\


\midrule

\textbf{Politeness} &
Professional courtesy &
Request-politeness markers &
Lexicon \variable{PLEASE} (\texttt{binary});
e.g., \textit{please}, \textit{plz}, \textit{kindly} \\
&
Professional courtesy &
Gratitude markers &
Lexicon \variable{THANKS} (\texttt{binary});
e.g., \textit{thanks}, \textit{thank you}, \textit{appreciated} \\
&
Professional courtesy &
Formal courtesy terms &
Lexicon \variable{COURTESY} (\texttt{binary});
e.g., \textit{sir}, \textit{ma'am}, \textit{Mr.} \\
&
Professional courtesy &
Apology, reassurance, and greeting cues &
Binary lexicons for apology, reassurance, and greeting;
e.g., \textit{sorry}, \textit{it's okay}, \textit{hello} \\

\cmidrule{2-4}

&
Rapport building &
Rapport-building phrases &
Lexicon \variable{RAPPORT} (\texttt{binary});
e.g., \textit{good point}, \textit{well said}, \textit{totally agree} \\
&
Rapport building &
Conversational rapport markers &
Binary greeting, filler, topic-shift, and agreement lexicons;
e.g., \textit{hi}, \textit{um}, \textit{by the way},
\textit{actually} \\

\cmidrule{2-4}

&
Non-imposition &
Hedges and softeners &
Lexicon \variable{HEDGES} (\texttt{binary});
e.g., \textit{may}, \textit{might}, \textit{can} \\
&
Non-imposition &
Indirect request forms &
Lexicon \variable{INDIRECTS} (\texttt{binary});
e.g., \textit{would you}, \textit{could you},
\textit{would it be possible} \\
&
Non-imposition &
Subjunctive request forms &
Binary subjunctive-request lexicon;
e.g., \textit{could you}, \textit{would you} \\
&
Non-imposition &
Indicative request forms &
Binary indicative-request lexicon;
e.g., \textit{can you}, \textit{will you} \\
&
Non-imposition &
Perspective softeners &
Binary perspective lexicons;
e.g., \textit{for me}, \textit{for us}, \textit{for you} \\

\cmidrule{2-4}

&
Domineering &
Directive and command cues &
Lexicon \variable{DOMINEERING} (\texttt{binary});
e.g., \textit{must}, \textit{have to}, \textit{need to} \\


\midrule

\textbf{Commitment} &
Statements &
Personal or group commitment statements &
Lexicon \variable{COMMITMENT} (\texttt{binary});
e.g., \textit{I have already}, \textit{I've already},
\textit{I have been} \\

\cmidrule{2-4}

&
Power &
Commitment followed by evidence of action &
Sequential rule linking \variable{COMMITMENT} to
\variable{COMMITMENT_PROOF};
e.g., a commitment statement followed by documentation,
verification, or other evidence of action \\


\midrule

\textbf{Style} &
Fluency &
Lexical rarity &
\texttt{wordfreq} Zipf frequency;
words below 2.5 treated as low-frequency items \\

\cmidrule{2-4}

&
Length &
Argument length &
Formula \variable{LENGTH}: log-transformed token count,
$\log(1+\text{tokens})/7$ \\

\cmidrule{2-4}

&
Rhetorical punctuation &
Quotation marks &
Regex detection of quotation marks \\
&
Rhetorical punctuation &
Question marks &
Regex detection of question marks \\
&
Rhetorical punctuation &
Exclamation marks &
Regex detection of exclamation marks \\


\specialrule{0.8pt}{5pt}{2pt}
\multicolumn{4}{@{}l}{
\textbf{\large Pathos: Emotion and Affection}
} \\
\specialrule{0.5pt}{2pt}{3pt}


\textbf{Sentiment} &
VADER compound &
Overall sentiment polarity &
VADER compound sentiment score \\

\cmidrule{2-4}

&
Language intensity &
Extreme-intensity language &
Lexicon \variable{SENT_INTENSITY.extreme};
e.g., \textit{catastrophic}, \textit{devastating}, \textit{horrific} \\
&
Language intensity &
Strong-intensity language &
Lexicon \variable{SENT_INTENSITY.strong};
e.g., \textit{absolutely}, \textit{extremely}, \textit{critical} \\
&
Language intensity &
Moderate-intensity language &
Lexicon \variable{SENT_INTENSITY.moderate};
e.g., \textit{significant}, \textit{substantial},
\textit{considerable} \\
&
Language intensity &
Weak intensifiers &
Lexicon \variable{SENT_INTENSITY.weak};
e.g., \textit{really}, \textit{very}, \textit{quite} \\
&
Language intensity &
Directive and urgency force &
Weighted combination of \variable{DOMINEERING},
\variable{URGENCY_TIME}, \variable{ARG_CLAIM}, and
\variable{IMPACT_THREAT};
e.g., \textit{must}, \textit{urgent}, \textit{should},
\textit{catastrophic} \\

\cmidrule{2-4}

&
Fear/threat &
Anxiety language &
LIWC \textit{Anx} category \\
&
Fear/threat &
Fear and threat language &
Lexicons \variable{PROP_FEAR} and \variable{IMPACT_THREAT};
e.g., \textit{threat}, \textit{danger}, \textit{deadly} \\

\cmidrule{2-4}

&
Joy/gain &
Positive-emotion language &
LIWC \textit{Posemo} category \\

\cmidrule{2-4}

&
Anger &
Anger language &
LIWC \textit{Anger} category \\

\cmidrule{2-4}

&
Sadness &
Sadness language &
LIWC \textit{Sad} category \\

\cmidrule{2-4}

&
Valence &
Affective valence &
NRC-VAD valence ratings \\

\cmidrule{2-4}

&
Arousal &
Affective arousal &
NRC-VAD arousal ratings \\

\cmidrule{2-4}

&
Dominance &
Affective dominance &
NRC-VAD dominance ratings \\


\midrule

\textbf{Impact} &
Gain framing &
Impact and enabling phrases &
Lexicon \variable{IMPACT} (\texttt{binary});
e.g., \textit{make a difference}, \textit{have an impact},
\textit{change my life} \\
&
Gain framing &
Gain and benefit language &
Lexicon \variable{IMPACT_GAIN} (\texttt{binary});
e.g., \textit{benefit}, \textit{benefits}, \textit{advantage} \\
&
Gain framing &
Soft gain frames &
Regex \variable{RE_SOFT_GAIN} (\texttt{binary});
e.g., \textit{better solution}, \textit{better approach},
\textit{better strategy} \\


&
Loss framing &
Loss and cost language &
Lexicon \variable{IMPACT_LOSS} (\texttt{binary});
e.g., \textit{loss}, \textit{waste}, \textit{cost} \\

\cmidrule{2-4}

&
Threat severity &
Threat and severity language &
Lexicon \variable{IMPACT_THREAT} (\texttt{binary});
e.g., \textit{catastrophic}, \textit{deadly}, \textit{fatal} \\

\cmidrule{2-4}

&
Future projection &
Future-oriented modal cues &
Lexicon \variable{IMPACT_WILL} (\texttt{binary});
e.g., \textit{will}, \textit{going to}, \textit{gonna} \\
&
Future projection &
Time-span projections &
Regex \variable{RE_TIME_SPAN} (\texttt{binary});
e.g., \textit{within 3 days}, \textit{in two weeks},
\textit{over 1 year} \\
&
Future projection &
Future-frame imagery &
Regex \variable{RE_FUTURE_FRAME} (\texttt{binary});
e.g., \textit{future generations}, \textit{imagine a world},
\textit{too late} \\


\midrule

\textbf{Scarcity/Urgency} &
Temporal urgency &
Urgency time cues &
Lexicon \variable{URGENCY_TIME};
e.g., \textit{urgent}, \textit{urgently}, \textit{immediately} \\
&
Temporal urgency &
Near-time cues &
Lexicon \variable{SPEC_PSYCH_NEAR};
e.g., \textit{minute}, \textit{hour}, \textit{day} \\

\cmidrule{2-4}

&
Exclusivity/quantity &
Scarcity and exclusivity cues &
Lexicon \variable{SCARCITY_QUANTITY};
e.g., \textit{only}, \textit{just}, \textit{few left} \\


\midrule

\textbf{Engagement} &
Identification &
First-person identity cues &
LIWC \textit{I} category \\
&
Identification &
Collective pronouns &
Regex \variable{RE_WE};
e.g., \textit{we}, \textit{our}, \textit{us} \\
&
Identification &
Collective-care phrases &
Regex \variable{RE_WE_care};
e.g., \textit{we care about}, \textit{our future},
\textit{our children} \\

\cmidrule{2-4}

&
Self-reference &
Second-person reference &
LIWC \textit{You} category \\

\cmidrule{2-4}

&
Inquiry &
Question marks &
Regex \variable{RE_QUESTION} (\texttt{binary}) \\

\cmidrule{2-4}

&
Past &
Past-tense language &
LIWC \textit{Past} category \\

\cmidrule{2-4}

&
Imagery &
Visual-perception language &
LIWC \textit{See} category \\
&
Imagery &
Imagery prompts &
Regex \variable{RE_IMAGERY};
e.g., \textit{imagine}, \textit{picture}, \textit{visualize} \\

\cmidrule{2-4}

&
Characters &
Personal pronouns and characters &
LIWC \textit{Ppron} category \\


\midrule

\textbf{Reciprocity} &
Direct promise &
Direct reciprocity-promise patterns &
Regex \variable{RECIPROCITY_PATTERNS} (\texttt{binary});
e.g., \textit{return the favor}, \textit{do the same for you},
\textit{pass it on} \\

\cmidrule{2-4}

&
Mutual benefit &
Mutual-benefit and exchange cues &
Lexicon \variable{REC_MUTUAL_EXCHANGE} (\texttt{binary});
e.g., \textit{together}, \textit{mutual}, \textit{shared} \\

\cmidrule{2-4}

&
Concession framing &
Concession and compromise cues &
Lexicon \variable{REC_CONCESSION} (\texttt{binary});
e.g., \textit{least I can do}, \textit{compromise},
\textit{meet in the middle} \\
&
Concession framing &
Soft concession frames &
Regex \variable{RE_SOFT_CONCESSION} (\texttt{binary});
e.g., \textit{would you be open to}, \textit{just wanted to},
\textit{might be helpful} \\


\midrule

\textbf{Propaganda} &
Emotional charge &
Loaded language &
Lexicon \variable{PROP_LOADED} (\texttt{binary});
e.g., \textit{corrupt}, \textit{evil}, \textit{traitor} \\
&
Emotional charge &
Name-calling cues &
Lexicon \variable{PROP_NAMECALL} (\texttt{binary});
e.g., \textit{idiot}, \textit{fool}, \textit{liar} \\
&
Emotional charge &
Fear, loss, and threat cues &
Combined \variable{PROP_FEAR}, \variable{IMPACT_LOSS}, and
\variable{IMPACT_THREAT} lexicons;
e.g., \textit{danger}, \textit{loss}, \textit{deadly} \\

\cmidrule{2-4}

&
Logical distortion &
Whataboutism cues &
Lexicon \variable{PROP_WHATABOUT} (\texttt{binary});
e.g., \textit{what about}, \textit{meanwhile},
\textit{they ignore} \\
&
Logical distortion &
Strawman cues &
Lexicon \variable{PROP_STRAW} (\texttt{binary});
e.g., \textit{they claim}, \textit{they say},
\textit{but actually} \\
&
Logical distortion &
Vagueness-as-authority cues &
Lexicon \variable{PROP_VAGUE} (\texttt{binary});
e.g., \textit{some say}, \textit{experts agree},
\textit{it is known} \\
&
Logical distortion &
Oversimplification cues &
Lexicon \variable{PROP_OVERSIMPLIFY} (\texttt{binary});
e.g., \textit{it's simple}, \textit{the only reason},
\textit{just because} \\

\cmidrule{2-4}

&
Heuristic identity appeals &
National-identity cues &
Lexicon \variable{PROP_FLAG};
e.g., \textit{nation}, \textit{country}, \textit{people} \\
&
Heuristic identity appeals &
Bandwagon cues &
Lexicon \variable{PROP_BANDWAGON};
e.g., \textit{everyone}, \textit{all of us},
\textit{people know} \\
&
Heuristic identity appeals &
Slogan cues &
Lexicon \variable{PROP_SLOGAN};
e.g., \textit{make America great again}, \textit{no more},
\textit{never forget} \\
&
Heuristic identity appeals &
Propaganda authority cues &
Lexicon \variable{PROP_AUTH};
e.g., \textit{according to}, \textit{scientists say},
\textit{researchers found} \\

\midrule


\caption{
Persuasion Index (PI) subfeatures and their linguistic operationalizations.
\textbf{Dimension} denotes one of the 15 PI dimensions;
\textbf{Subfeature} denotes one of the 55 finer-grained components;
\textbf{Linguistic cue} describes the textual phenomenon being measured; and
\textbf{Detector / examples} identifies the computational resource, lexicon,
regex, or rule together with representative examples.
Exact regex expressions, complete frozen lexicons, and implementation details
are provided in the released codebase.
Unless otherwise indicated, lexical cue counts are normalized to
per-100-token density; detectors marked \texttt{binary} are scored as
document-level presence/absence indicators.
}
\label{tab:pi_methods_one_cue}
\\

\end{longtable}
\endgroup

\twocolumn

\section{Datasets}
\label{appendix:datasets}

\begin{table*}[t]
\centering
\resizebox{\textwidth}{!}{%
\small
\begin{tabular}{lllllll}
\toprule
\textbf{Dataset} & \textbf{Size} & \textbf{Domain} & \textbf{Label Type} & \textbf{Persuasiveness Signal} & \textbf{Setting} & \textbf{Annotator} \\
\midrule
UKPConvArg1 \citep{habernalWhichArgumentMore2016} & 11,650 pairs & Web debate forums (16 topics) & Pairwise winner & Crowd convincingness judgment & Pairwise & Crowd \\
CMV \citep{tanWinningArgumentsInteraction2016} & 8,526 replies & Reddit /r/CMV (diverse topics) & Binary ($\Delta$ / no $\Delta$) & Original poster $\Delta$ award & Single-arg & Community \\
IBM Arg. Quality \citep{gretzLargescaleDatasetArgument2020} & 14,003 args & Debating (71 topics) & Quality score + pairs & Expert quality rating & Pairwise & Expert \\
Anthropic Persuasion \citep{MeasuringPersuasivenessLanguage} & 3,882 args & Diverse (human \& LLM-gen.) & Continuous ($\Delta$ rating) & Pre/post attitude shift & Single-arg & Human subjects \\
\bottomrule
\end{tabular}%
}
\caption{Overview of the four evaluation datasets.}
\label{tab:datasets}
\end{table*}

We construct and evaluate the PI using four publicly released datasets (shown in Table~\ref{tab:datasets}) that together cover pairwise convincingness, topic- and side-informed arguments, and expert or crowd ratings of individual arguments.

\paragraph{UKPConvArg1 (UKP) \cite{habernalWhichArgumentMore2016}.}
UKPConvArg1 is a \textbf{pairwise} convincingness dataset (n=11,650) in which each instance contains two short arguments and crowd judgments indicating which argument is more convincing. Because arguments often appear in multiple pairwise comparisons within a topic, the dataset supports both topic-level and side-level analyses. UKPConvArg1 provides our primary supervision signal for estimating PI weights.

Table~\ref{tab:ukp_topic_distribution} reports the full topic distribution in UKPConvArg1, listing the original topic identifiers, their shortened display labels, and the number of unique arguments per topic. The display labels are used in all figures and coefficient plots throughout the paper for readability.

\begin{table*}[htbp!]
\centering
\small
\resizebox{.8\textwidth}{!}{%
\begin{tabular}{llr}
\toprule
\textbf{Original Identifier} & \textbf{Display Label} & \textbf{N} \\
\midrule
is-the-school-uniform-a-good-or-bad-idea- & School Uniforms & 70 \\
pro-choice-vs-pro-life & Abortion Debate & 70 \\
human-growth-and-development-should-parents-use-spanking-as-an-option-to-discipline- & Spanking Discipline & 70 \\
is-it-better-to-have-a-lousy-father-or-to-be-fatherless- & Father vs Fatherless & 70 \\
if-your-spouse-committed-murder-and-he-or-she-confided-in-you-would-you-turn-them-in- & Reporting a Spouse & 70 \\
gay-marriage-right-or-wrong & Gay Marriage & 70 \\
personal-pursuit-or-advancing-the-common-good- & Individual vs Common Good & 70 \\
william-farquhar-ought-to-be-honoured-as-the-rightful-founder-of-singapore & Farquhar as Founder & 69 \\
evolution-vs-creation & Evolution Debate & 68 \\
india-has-the-potential-to-lead-the-world- & India Leading & 67 \\
tv-is-better-than-books & TV vs Books & 62 \\
firefox-vs-internet-explorer & Firefox vs IE & 62 \\
christianity-or-atheism- & Christianity vs Atheism & 62 \\
ban-plastic-water-bottles & Ban Plastic Bottles & 60 \\
is-porn-wrong- & Porn Morality & 56 \\
should-physical-education-be-mandatory-in-schools- & Mandatory PE & 56 \\
\midrule
\textbf{Total} & & \textbf{1,052} \\
\bottomrule
\end{tabular}%
}
\caption{Topic distribution in UKPConvArg1. $N$ is the number of unique arguments per topic after deduplication across pairwise comparisons. Display labels are shortened versions of the original identifiers used in all figures and coefficient plots throughout the paper.}
\label{tab:ukp_topic_distribution}
\end{table*}

\paragraph{ChangeMyView (CMV) \cite{tanWinningArgumentsInteraction2016}.}
Rather than the full CMV corpus, we use the ConvoKit ``Winning Arguments’’ subset (n=8,526), where each reply is labeled according to whether the original poster awarded a \(\Delta\), signaling successful persuasion. Although CMV is not pairwise, the \(\Delta\) annotation naturally produces a binary persuasive outcome: positive \(\Delta\) indicates a persuasive (winning) argument, and vice versa.

\paragraph{IBM Argument Quality (IBM) \cite{gretzLargescaleDatasetArgument2020}.}
The IBM Argument Quality dataset (n=14,003) contains expert-annotated assessments of argument quality and includes \textbf{pairwise} comparisons within topics. Although framed as argument quality rather than persuasion, prior work shows that these dimensions correlate strongly with theoretical constructs of persuasive effectiveness, making it suitable for evaluating PI’s interpretability and generalization.

Table~\ref{tab:ibm_topic_distribution} reports the full topic distribution in IBM Argument Quality, listing the original topic identifiers, their shortened display labels, the stance either PRO or CON, and the number of unique arguments per topic. The display labels are used in all figures and coefficient plots throughout the paper for readability.

\begin{table*}[htbp!]
\centering
\small
\resizebox{.75\textwidth}{!}{%
\begin{tabular}{lllr}
\toprule
\textbf{Original Source File} & \textbf{Topic} & \textbf{Stance} & \textbf{N} \\
\midrule
Social-media-brings-MORE-GOOD-than-harm & Social Media & PRO & 654 \\
Social-media-brings-MORE-HARM-than-good & Social Media & CON & 520 \\
We-should-support-information-privacy-laws-(PRO) & Information Privacy & PRO & 253 \\
We-should-promote-autonomous-cars-(CON) & Autonomous Cars & CON & 205 \\
We-should-adopt-vegetarianism-(PRO) & Vegetarianism & PRO & 163 \\
We-should-abandon-vegetarianism-(CON) & Vegetarianism & CON & 146 \\
We-should-legalize-doping-in-sport-(PRO) & Doping in Sport & PRO & 141 \\
Flu-vaccination-should-be-mandatory-(PRO) & Flu Vaccination & PRO & 131 \\
We-should-ban-doping-in-sport-(CON) & Doping in Sport & CON & 129 \\
We-should-adopt-cryptocurrency-(PRO) & Cryptocurrency & PRO & 127 \\
Flu-vaccination-should-not-be-mandatory-(CON) & Flu Vaccination & CON & 121 \\
We-should-abandon-cryptocurrency-(CON) & Cryptocurrency & CON & 120 \\
We-should-limit-autonomous-cars-(PRO) & Autonomous Cars & PRO & 120 \\
Gambling-should-be-banned-(PRO) & Gambling & PRO & 109 \\
We-should-ban-the-sale-of-violent-video-games-to-minors-(PRO) & Sale Violent Games & PRO & 101 \\
Gambling-should-not-be-banned-(CON) & Gambling & CON & 94 \\
We-should-allow-the-sale-of-violent-video-games-to-minors-(CON) & Sale Violent Games & CON & 91 \\
We-should-ban-fossil-fuels-(PRO) & Fossil Fuels & PRO & 69 \\
Online-shopping-brings-more-harm-than-good-(PRO) & Online Shopping & PRO & 69 \\
We-should-not-ban-fossil-fuels-(CON) & Fossil Fuels & CON & 64 \\
Online-shopping-brings-more-good-than-harm-(CON) & Online Shopping & CON & 63 \\
We-should-discourage-information-privacy-laws-(CON) & Information Privacy & CON & 43 \\
\midrule
\textbf{Total} & & & \textbf{3,354} \\
\bottomrule
\end{tabular}%
}
\caption{Source file distribution in the IBM Argument Quality dataset. Each row corresponds to one topic--stance combination. Topic labels are used in all figures and coefficient plots throughout the paper.}
\label{tab:ibm_topic_distribution}
\end{table*}

\paragraph{Anthropic Dataset (Anthropic) \cite{MeasuringPersuasivenessLanguage}.}
We include the publicly released Anthropic Persuasion dataset (n=3,882), a collection of human- and model-generated arguments with participants’ pre- and post-exposure agreement ratings and derived persuasiveness scores. Each instance consists of a single argument paired with participant pre- and post-exposure attitude ratings, enabling argument-level analysis without requiring pairwise comparisons. Because the dataset directly encodes persuasive behaviors across a wide range of topics and styles, it provides a complementary evaluation setting for testing the generality and interpretability of the Persuasion Index.

\section{Persuasiveness Signal Construction}
\label{sec:persuasiveness-signal-construction}

Our datasets provide two types of supervision: pairwise convincingness comparisons (UKP, IBM) and single-instance persuasion outcomes (CMV, Anthropic).

\paragraph{Pairwise datasets (UKP and IBM).} For each argument pair (A,B), we construct a difference vector $\Delta\mathbf{x} = \mathbf{x}_A - \mathbf{x}_B$ from each argument's PI feature vectors, with binary label y=1 if argument A is judged more convincing and y=0 otherwise. This formulation encodes the contrastive signal between paired arguments without aggregating win--loss statistics across the full dataset, avoiding the confound where an argument's label depends on which other arguments it happens to be paired with.

\paragraph{CMV.} Each reply is labeled y=1 if the original poster awarded a $\Delta$, signaling that the reply successfully changed their view, and y=0 otherwise.

\paragraph{Anthropic.} Each argument includes participant attitude ratings before and after exposure. We first remove participants with extreme baseline attitudes, those who initially selected the strongest agree or strongest disagree positions, as these show strong ceiling and floor effects unlikely to reflect genuine persuasion: 97\% of participants with the strongest initial agreement maintained that position after reading, and 74.4\% of those with the strongest initial disagreement did the same. After filtering, we label y=1 if the post-exposure attitude rating improves by more than 1 point relative to the pre-exposure baseline, capturing substantive attitude change while reducing noise from marginal shifts.

\section{Model}
\label{appendix:model}
We evaluate the Persuasion Index by comparing logistic regression classifiers fit on PI feature vectors to fine-tuned \model{RoBERTa} and zero-shot \model{GPT-4o}. All models are evaluated on the same held-out 20\% test split; supervised models are trained on the remaining 80\%.

\paragraph{Logistic Regression.} Our primary predictive models are $\ell_2$-regularized logistic regressions trained on either the 15-dimensional PI category-mean vector (\model{\model{PI-mean}}) or the full 55-dimensional sub-feature vector (\model{\model{PI-sub}}). Given feature vector $\mathbf{x}$ for argument $i$, the model estimates $P(y_i = 1 \mid \mathbf{x}_i) = \sigma(\mathbf{w}^{\top}\mathbf{x}_i + b)$, where $y_i \in \{0, 1\}$ is the binary persuasion label (\S\ref{sec:persuasiveness-signal-construction}). We use a \textsc{StandardScaler} followed by \textsc{LogisticRegression} with the \texttt{liblinear} solver, \texttt{max\_iter=2000}, and \texttt{class\_weight=``balanced''} to correct for class imbalance. We report accuracy, F1 score, and ROC--AUC on the held-out test set. 

\vspace{1ex}
\noindent\textbf{Pairwise setting (UKP and IBM).} For pairwise datasets, we construct difference-vector features by subtracting two PI feature vectors: $\Delta\mathbf{x} = \mathbf{x}_A - \mathbf{x}_B$, and train a standard logistic regression on $\Delta\mathbf{x}$ with binary label $y = 1$ if argument $A$ is judged more convincing. This formulation isolates the specific PI dimensions that distinguish winning arguments from losing ones, rather than modeling each argument's absolute score. We evaluate via pairwise accuracy, defined as the proportion of correctly identified winners across all held-out pairs.

\paragraph{\model{RoBERTa} Neural Baseline.} We compare PI to \model{RoBERTa-base} \citep{liuRoBERTaRobustlyOptimized2019}. Arguments are tokenized with a maximum sequence length of 128 using \texttt{RobertaTokenizer}, and we fine-tune \texttt{RobertaForSequenceClassification} with a two-class softmax head for binary persuasion prediction. We train for two epochs using AdamW with learning rate $5 \times 10^{-5}$ and batch size 8. To account for stochastic variation in fine-tuning, we train \model{RoBERTa} with 10 random seeds and report mean performance across runs, with standard deviations provided to characterize variability. The original seed-42 run is retained separately for the paired instance-level statistical comparisons in Appendix~\ref{appendix:test_across_models} and is not included in the 10-run averages. At evaluation time, predicted probabilities $p(y{=}1)$ are obtained via softmax over the output logits. Rather than applying the default threshold of 0.5, we select a decision threshold on a held-out validation set using Youden's $J$ statistic \citep{youdenIndexRatingDiagnostic1950}, $t^*=\arg\max_t[\mathrm{TPR}(t)-\mathrm{FPR}(t)]$, and apply this threshold unchanged to the test set. Test labels are not used for threshold selection. Seed-level results are reported in Appendix~\ref{appendix:roberta_seeds}.

\paragraph{\model{GPT-4o} Zero-shot Baseline.} We additionally evaluate \model{GPT-4o} as a zero-shot baseline to benchmark PI against frontier LLM judgment without any fine-tuning. For pairwise datasets (UKP and IBM), we prompt \model{GPT-4o} to directly compare two arguments and output which is more persuasive. For single-instance datasets (CMV and Anthropic), we prompt the model to produce a binary judgment (1 = persuasive, 0 = not persuasive) for each argument independently. Full prompts are as follows:

\begin{lstlisting}
TASK:
Read the two texts below and decide which one is more persuasive.

OUTPUT:
a1 = the first text ("a1") is more persuasive than the second text ("a2")
a2 = the second text ("a2") is more persuasive than the first text ("a1")

OUTPUT FORMAT:
Return only one label: a1 or a2.
Do not output anything else.
\end{lstlisting}
\begin{lstlisting}
TASK:
Read the text below and decide whether or not it is persuasive.

OUTPUT:
0 = not persuasive
1 = persuasive

OUTPUT FORMAT:
Return only one character: 0 or 1.
Do not output anything else.
\end{lstlisting}

\section{Prediction Results}
\label{appendix:results}

\subsection{Model Comparisons}
\label{appendix:test_across_models}

We report three statistical tests to characterize how PI models compare to \model{\model{RoBERTa}} and \model{\model{GPT-4o}}. For these pairwise tests, we use the original \model{RoBERTa} model trained with random seed 42, yielding a single fixed set of predictions for instance-level comparison. The 10-seed \model{RoBERTa} results reported in the main text characterize variation due to stochastic fine-tuning and are detailed separately in Section~\ref{appendix:roberta_seeds}. 

\textbf{McNemar's test} examines whether two models make systematically different errors on the same test instances, with $p<0.05$ indicating significantly different error patterns. \textbf{Bootstrap difference tests} estimate the sampling distribution of the accuracy gap via 1,000 resamples, reporting 95\% CIs and two-sided p-values for the observed difference. \textbf{TOST (Two One-Sided Tests)} tests practical equivalence within a $\pm 0.05$ accuracy margin; $p<0.05$ confirms the two models are equivalent, while failure to reject is inconclusive. 

\begin{table}[t]
\centering
\small
\resizebox{\columnwidth}{!}{
\begin{tabular}{lrrrrrrr}
\toprule
Comparison & Acc(M1) & Acc(M2) & $\Delta$ & Bootstrap 95\% CI & Boot $p$ & McNemar $p$ & TOST $p$ \\
\midrule
PI-sub vs PI-mean & 0.775 & 0.690 & \textbf{+0.085} & [0.035, 0.135] & \textbf{0.001} & \textbf{0.001} & 0.869 \\
PI-sub vs RoBERTa & 0.775 & 0.595 & \textbf{+0.180} & [0.122, 0.237] & \textbf{0.000} & \textbf{0.000} & 1.000 \\
PI-sub vs GPT-4o & 0.775 & 0.855 & -0.080 & [-0.128, -0.030] & \textbf{0.001} & \textbf{0.002} & 0.864 \\
PI-mean vs RoBERTa & 0.690 & 0.595 & \textbf{+0.095} & [0.035, 0.158] & \textbf{0.002} & \textbf{0.004} & 0.909 \\
PI-mean vs GPT-4o & 0.690 & 0.855 & -0.165 & [-0.220, -0.110] & \textbf{0.000} & \textbf{0.000} & 1.000 \\
RoBERTa vs GPT-4o & 0.595 & 0.855 & -0.260 & [-0.318, -0.205] & \textbf{0.000} & \textbf{0.000} & 1.000 \\
\bottomrule
\end{tabular}
}
\caption{UKP pairwise setting. Bold $\Delta$ marks significant wins for the first-listed model; bold $p$-values indicate $p<0.05$. }
\label{tab:test_ukp}
\end{table}

On UKP (Table~\ref{tab:test_ukp}), \model{\model{PI-sub}} significantly outperforms \model{\model{PI-mean}} (+8.5 pp, p=.001). \model{PI-sub} also significantly beats \model{\model{RoBERTa}} by 18.0 pp (p<.001) but trails \model{GPT-4o} by 8.0 pp (p=.001). No pair is confirmed equivalent under TOST.

\begin{table}[t]
\centering
\small
\resizebox{\columnwidth}{!}{
\begin{tabular}{lrrrrrrr}
\toprule
Comparison & Acc(M1) & Acc(M2) & $\Delta$ & Bootstrap 95\% CI & Boot $p$ & McNemar $p$ & TOST $p$ \\
\midrule
PI-sub vs PI-mean & 0.585 & 0.560 & \textbf{+0.025} & [0.003, 0.049] & \textbf{0.032} & \textbf{0.040} & 0.077 \\
PI-sub vs GPT-4o & 0.585 & 0.528 & \textbf{+0.057} & [0.025, 0.088] & \textbf{0.000} & \textbf{0.001} & 0.650 \\
PI-sub vs RoBERTa & 0.585 & 0.520 & \textbf{+0.065} & [0.030, 0.099] & \textbf{0.000} & \textbf{0.000} & 0.801 \\
PI-mean vs GPT-4o & 0.560 & 0.528 & +0.031 & [-0.001, 0.064] & 0.055 & 0.059 & 0.144 \\
PI-mean vs RoBERTa & 0.560 & 0.520 & \textbf{+0.040} & [0.004, 0.075] & \textbf{0.028} & \textbf{0.031} & 0.273 \\
GPT-4o vs RoBERTa & 0.528 & 0.520 & +0.008 & [-0.026, 0.043] & 0.662 & 0.678 & \textbf{0.008} \\
\bottomrule
\end{tabular}
}
\caption{CMV setting. Bold $\Delta$ marks significant wins for the first-listed model; bold $p$-values indicate $p<0.05$.}
\label{tab:test_cmv}
\end{table}

CMV (Table~\ref{tab:test_cmv}) is the only dataset where \model{PI-sub} leads outright. \model{PI-sub} significantly outperforms \model{PI-mean} (+2.5 pp, p=.032), \model{GPT-4o} (+5.7 pp, p<.001), and \model{RoBERTa} (+6.5 pp, p<.001), demonstrating stronger generalization on organic Reddit persuasion data than either fine-tuned \model{RoBERTa} or zero-shot \model{GPT-4o}. \model{PI-mean} is significantly above \model{RoBERTa} (+4.0 pp, p=.028) and trends above \model{GPT-4o} (+3.1 pp, p=.055, borderline). 

\begin{table}[t]
\centering
\small
\resizebox{\columnwidth}{!}{
\begin{tabular}{lrrrrrrr}
\toprule
Comparison & Acc(M1) & Acc(M2) & $\Delta$ & Bootstrap 95\% CI & Boot $p$ & McNemar $p$ & TOST $p$ \\
\midrule
PI-sub vs PI-mean & 0.590 & 0.575 & +0.015 & [-0.037, 0.068] & 0.569 & 0.634 & 0.158 \\
PI-sub vs RoBERTa & 0.590 & 0.517 & \textbf{+0.072} & [0.003, 0.142] & \textbf{0.038} & \textbf{0.046} & 0.740 \\
PI-sub vs GPT-4o & 0.590 & 0.650 & -0.060 & [-0.122, 0.003] & 0.062 & 0.079 & 0.615 \\
PI-mean vs RoBERTa & 0.575 & 0.517 & +0.058 & [-0.013, 0.125] & 0.104 & 0.123 & 0.585 \\
PI-mean vs GPT-4o & 0.575 & 0.650 & -0.075 & [-0.140, -0.007] & \textbf{0.026} & \textbf{0.033} & 0.767 \\
RoBERTa vs GPT-4o & 0.517 & 0.650 & -0.133 & [-0.193, -0.072] & \textbf{0.000} & \textbf{0.000} & 0.992 \\
\bottomrule
\end{tabular}
}
\caption{IBM pairwise setting. Bold $\Delta$ marks significant wins for the first-listed model; bold $p$-values indicate $p<0.05$.}
\label{tab:test_ibm}
\end{table}

On IBM (Table~\ref{tab:test_ibm}), \model{GPT-4o} is the strongest model overall, significantly outperforming \model{PI-mean} (+7.5 pp, p=.026) and \model{RoBERTa} (+13.3 pp, p<.001). \model{PI-sub} closes much of this gap: the 6.0 pp difference from \model{GPT-4o} does not reach significance (boot p=.062, McNemar p=.079), while \model{PI-sub} significantly beats \model{RoBERTa} (+7.2 pp, p=.038). \model{PI-sub} and \model{PI-mean} remain indistinguishable from each other (+1.5 pp, p=.569). No pair is confirmed equivalent under TOST, so non-significant comparisons here may be underpowered rather than truly equivalent.

\begin{table}[t]
\centering
\small
\resizebox{\columnwidth}{!}{
\begin{tabular}{lrrrrrrr}
\toprule
Comparison & Acc(M1) & Acc(M2) & $\Delta$ & Bootstrap 95\% CI & Boot $p$ & McNemar $p$ & TOST $p$ \\
\midrule
PI-sub vs PI-mean & 0.533 & 0.516 & +0.017 & [-0.022, 0.055] & 0.391 & 0.434 & 0.095 \\
PI-sub vs RoBERTa & 0.533 & 0.633 & -0.100 & [-0.143, -0.058] & \textbf{0.000} & \textbf{0.000} & 0.979 \\
PI-sub vs GPT-4o & 0.533 & 0.328 & \textbf{+0.205} & [0.158, 0.252] & \textbf{0.000} & \textbf{0.000} & 1.000 \\
PI-mean vs RoBERTa & 0.516 & 0.633 & -0.117 & [-0.161, -0.073] & \textbf{0.000} & \textbf{0.000} & 0.997 \\
PI-mean vs GPT-4o & 0.516 & 0.328 & \textbf{+0.188} & [0.142, 0.234] & \textbf{0.000} & \textbf{0.000} & 1.000 \\
RoBERTa vs GPT-4o & 0.633 & 0.328 & \textbf{+0.305} & [0.251, 0.359] & \textbf{0.000} & \textbf{0.000} & 1.000 \\
\bottomrule
\end{tabular}
}
\caption{Anthropic setting. Bold $\Delta$ marks significant wins for the first-listed model; bold $p$-values indicate $p<0.05$.}
\label{tab:test_anth}
\end{table}

Anthropic (Table~\ref{tab:test_anth}) presents a striking inversion of the UKP ordering. \model{RoBERTa} is the strongest model here (acc.\ 0.633), significantly outperforming \model{PI-sub} (+10.0 pp, p<.001) and \model{PI-mean} (+11.7 pp, p<.001). Both PI models substantially outperform \model{GPT-4o}: \model{PI-sub} beats \model{GPT-4o} by 20.5 pp and \model{PI-mean} by 18.8 pp (p<.001 across all tests). \model{PI-sub} and \model{PI-mean} are indistinguishable from each other (+1.7 pp, p=.391), suggesting that mean-level scores are sufficient on this task. 

\begin{table}[t] \centering
  \footnotesize
  \setlength{\tabcolsep}{3pt}
  \caption{Logistic regression: Pairwise Mean-feature Differences (UKP)}
  \label{tab:ukp_regression}
\begin{tabular}{@{\extracolsep{5pt}}lc}
\\[-1.8ex]\hline
\hline \\[-1.8ex]
& \multicolumn{1}{c}{\textit{Pairwise outcome}} \
\cr \cline{2-2}
\\[-1.8ex] & \multicolumn{1}{c}{PI-mean}  \\
\hline \\[-1.8ex]
 Argumentation & 0.147$^{**}$ \\
& (0.071) \\
 Authority/Credibility & 0.093$^{}$ \\
& (0.065) \\
 Commitment & 0.182$^{***}$ \\
& (0.060) \\
 Engagement & -0.140$^{**}$ \\
& (0.070) \\
 Evidence & 0.491$^{***}$ \\
& (0.064) \\
 Impact & 0.254$^{***}$ \\
& (0.064) \\
 Logic/Cohesion & 0.487$^{***}$ \\
& (0.076) \\
 Politeness & 0.128$^{*}$ \\
& (0.067) \\
 Propaganda & 0.054$^{}$ \\
& (0.063) \\
 Reciprocity & 0.054$^{}$ \\
& (0.059) \\
 Scarcity/Urgency & 0.052$^{}$ \\
& (0.063) \\
 Sentiment & 0.260$^{***}$ \\
& (0.068) \\
 Specificity & -0.311$^{***}$ \\
& (0.069) \\
 Style & 0.263$^{***}$ \\
& (0.069) \\
 const & -0.075$^{}$ \\
& (0.057) \\
\hline \\[-1.8ex]
 Observations & 1600 \\
 Pseudo $R^2$ & 0.178 \\
\hline
\hline \\[-1.8ex]
\textit{Note:} & \multicolumn{1}{r}{$^{*}$p$<$0.1; $^{**}$p$<$0.05; $^{***}$p$<$0.01} \\
\end{tabular}
\end{table}

\begin{table}[t] \centering
  \footnotesize
  \setlength{\tabcolsep}{3pt}
  \caption{Logistic regression: Mean-feature (CMV)}
  \label{tab:cmv_regression}
\begin{tabular}{@{\extracolsep{5pt}}lc}
\\[-1.8ex]\hline
\hline \\[-1.8ex]
& \multicolumn{1}{c}{\textit{Single outcome}} \
\cr \cline{2-2}
\\[-1.8ex] & \multicolumn{1}{c}{PI-mean}  \\
\hline \\[-1.8ex]
 Argumentation & -0.053$^{*}$ \\
& (0.028) \\
 Authority/Credibility & -0.006$^{}$ \\
& (0.032) \\
 Commitment & 0.028$^{}$ \\
& (0.027) \\
 Engagement & -0.017$^{}$ \\
& (0.031) \\
 Evidence & 0.148$^{***}$ \\
& (0.032) \\
 Impact & 0.084$^{***}$ \\
& (0.032) \\
 Logic/Cohesion & 0.021$^{}$ \\
& (0.028) \\
 Opponent’s View & -0.024$^{}$ \\
& (0.026) \\
 Politeness & 0.089$^{***}$ \\
& (0.029) \\
 Propaganda & 0.035$^{}$ \\
& (0.030) \\
 Reciprocity & 0.012$^{}$ \\
& (0.027) \\
 Scarcity/Urgency & 0.019$^{}$ \\
& (0.027) \\
 Sentiment & -0.029$^{}$ \\
& (0.028) \\
 Specificity & -0.017$^{}$ \\
& (0.028) \\
 Style & 0.092$^{***}$ \\
& (0.033) \\
 const & 0.095$^{***}$ \\
& (0.025) \\
\hline \\[-1.8ex]
 Observations & 6484 \\
 Pseudo $R^2$ & 0.014 \\
\hline
\hline \\[-1.8ex]
\textit{Note:} & \multicolumn{1}{r}{$^{*}$p$<$0.1; $^{**}$p$<$0.05; $^{***}$p$<$0.01} \\
\end{tabular}
\end{table}
\begin{table}[t] \centering
  \footnotesize
  \setlength{\tabcolsep}{3pt}
  \caption{Logistic regression: Pairwise Mean-feature Differences (IBM)}
  \label{tab:ibm_regression}
\begin{tabular}{@{\extracolsep{5pt}}lc}
\\[-1.8ex]\hline
\hline \\[-1.8ex]
& \multicolumn{1}{c}{\textit{Pairwise outcome}} \
\cr \cline{2-2}
\\[-1.8ex] & \multicolumn{1}{c}{PI-mean}  \\
\hline \\[-1.8ex]
 Argumentation & 0.017$^{}$ \\
& (0.085) \\
 Authority/Credibility & 0.015$^{}$ \\
& (0.058) \\
 Commitment & -0.178$^{***}$ \\
& (0.061) \\
 Engagement & -0.267$^{***}$ \\
& (0.058) \\
 Evidence & 0.135$^{***}$ \\
& (0.052) \\
 Impact & -0.062$^{}$ \\
& (0.052) \\
 Logic/Cohesion & 0.052$^{}$ \\
& (0.083) \\
 Politeness & 0.029$^{}$ \\
& (0.061) \\
 Propaganda & 0.059$^{}$ \\
& (0.052) \\
 Reciprocity & 0.037$^{}$ \\
& (0.056) \\
 Scarcity/Urgency & -0.096$^{}$ \\
& (0.071) \\
 Sentiment & 0.224$^{***}$ \\
& (0.053) \\
 Specificity & 0.022$^{}$ \\
& (0.074) \\
 Style & 0.041$^{}$ \\
& (0.057) \\
 const & -0.032$^{}$ \\
& (0.051) \\
\hline \\[-1.8ex]
 Observations & 1600 \\
 Pseudo $R^2$ & 0.030 \\
\hline
\hline \\[-1.8ex]
\textit{Note:} & \multicolumn{1}{r}{$^{*}$p$<$0.1; $^{**}$p$<$0.05; $^{***}$p$<$0.01} \\
\end{tabular}
\end{table}
\begin{table}[t] \centering
  \footnotesize
  \setlength{\tabcolsep}{3pt}
  \caption{Logistic regression: Mean-feature (Anthropic)}
  \label{tab:anth_regression}
\begin{tabular}{@{\extracolsep{5pt}}lc}
\\[-1.8ex]\hline
\hline \\[-1.8ex]
& \multicolumn{1}{c}{\textit{Single outcome}} \
\cr \cline{2-2}
\\[-1.8ex] & \multicolumn{1}{c}{PI-mean}  \\
\hline \\[-1.8ex]
 Argumentation & 0.013$^{}$ \\
& (0.041) \\
 Authority/Credibility & 0.118$^{*}$ \\
& (0.061) \\
 Commitment & -0.016$^{}$ \\
& (0.041) \\
 Engagement & -0.060$^{}$ \\
& (0.047) \\
 Evidence & -0.032$^{}$ \\
& (0.061) \\
 Impact & 0.038$^{}$ \\
& (0.043) \\
 Logic/Cohesion & -0.002$^{}$ \\
& (0.041) \\
 Opponent’s View & 0.935$^{}$ \\
& (5582.655) \\
 Politeness & -0.044$^{}$ \\
& (0.039) \\
 Propaganda & -0.122$^{***}$ \\
& (0.041) \\
 Reciprocity & 0.027$^{}$ \\
& (0.040) \\
 Scarcity/Urgency & -0.020$^{}$ \\
& (0.044) \\
 Sentiment & -0.033$^{}$ \\
& (0.042) \\
 Specificity & 0.002$^{}$ \\
& (0.044) \\
 Style & -0.036$^{}$ \\
& (0.043) \\
 const & -0.696$^{}$ \\
& (200.503) \\
\hline \\[-1.8ex]
 Observations & 3105 \\
 Pseudo $R^2$ & 0.009 \\
\hline
\hline \\[-1.8ex]
\textit{Note:} & \multicolumn{1}{r}{$^{*}$p$<$0.1; $^{**}$p$<$0.05; $^{***}$p$<$0.01} \\
\end{tabular}
\end{table}

\subsection{RoBERTa Variation Across Random Seeds}
\label{appendix:roberta_seeds}

To characterize variation due to stochastic fine-tuning, we train \model{RoBERTa} independently with 10 random seeds on each dataset while holding the data splits and evaluation procedure fixed. We report accuracy, precision, recall, and F1 for each run, together with the mean and standard deviation across these 10 runs.

For reference, each table also reports the original seed-42 \model{RoBERTa} result used in the paired statistical comparisons in Section~\ref{appendix:test_across_models}. Seed 42 is shown separately and is not included in the mean or standard deviation of the 10 additional runs.

\begin{table}[t]
\centering
\small
\setlength{\tabcolsep}{5pt}
\begin{tabular}{lrrrr}
\toprule
Seed & Acc & Precision & Recall & F1 \\
\midrule
42$^\dagger$ & 0.5950 & 0.5900 & 0.5410 & 0.5650 \\
\midrule
101  & 0.8250 & 0.7520 & 0.9536 & 0.8409 \\
203  & 0.7925 & 0.7734 & 0.8093 & 0.7909 \\
307  & 0.8450 & 0.8204 & 0.8711 & 0.8450 \\
409  & 0.7700 & 0.8493 & 0.6392 & 0.7294 \\
503  & 0.5175 & 0.5017 & 0.7526 & 0.6021 \\
607  & 0.5900 & 0.5658 & 0.6649 & 0.6114 \\
709  & 0.7675 & 0.7463 & 0.7887 & 0.7669 \\
811  & 0.5250 & 0.5182 & 0.2938 & 0.3750 \\
907  & 0.8125 & 0.7874 & 0.8402 & 0.8130 \\
1009 & 0.5250 & 0.5097 & 0.5412 & 0.5250 \\
\midrule
Mean (10 runs) & 0.6970 & 0.6824 & 0.7155 & 0.6900 \\
SD (10 runs)   & 0.1390 & 0.1407 & 0.1911 & 0.1563 \\
\bottomrule
\end{tabular}
\caption{Seed-level \model{RoBERTa} performance on UKP.
$^\dagger$Seed 42 is the original run used for the paired model-comparison
tests and is shown for reference only. Mean and SD are computed over the
10 additional runs and exclude seed 42.}
\label{tab:roberta_seeds_ukp}
\end{table}

Table~\ref{tab:roberta_seeds_ukp} reports the seed-level results on UKP. Across the 10 additional runs, accuracy is $0.697 \pm 0.139$ and F1 is
$0.690 \pm 0.156$, indicating substantial variation across fine-tuning
runs.

\begin{table}[t]
\centering
\small
\setlength{\tabcolsep}{5pt}
\begin{tabular}{lrrrr}
\toprule
Seed & Acc & Precision & Recall & F1 \\
\midrule
42$^\dagger$ & 0.5200 & 0.5590 & 0.3890 & 0.4590 \\
\midrule
1103 & 0.4932 & 0.5100 & 0.7795 & 0.6166 \\
1201 & 0.4766 & 0.4979 & 0.1380 & 0.2161 \\
1301 & 0.5234 & 0.5389 & 0.6132 & 0.5736 \\
1409 & 0.5296 & 0.5478 & 0.5743 & 0.5607 \\
1511 & 0.4938 & 0.5212 & 0.3915 & 0.4471 \\
1601 & 0.5160 & 0.5300 & 0.6568 & 0.5866 \\
1709 & 0.5518 & 0.5463 & 0.8420 & 0.6626 \\
1801 & 0.5173 & 0.5242 & 0.8302 & 0.6426 \\
1901 & 0.5000 & 0.5496 & 0.2417 & 0.3358 \\
2003 & 0.5012 & 0.5141 & 0.8373 & 0.6371 \\
\midrule
Mean (10 runs) & 0.5103 & 0.5280 & 0.5904 & 0.5279 \\
SD (10 runs)   & 0.0217 & 0.0176 & 0.2557 & 0.1485 \\
\bottomrule
\end{tabular}
\caption{Seed-level \model{RoBERTa} performance on CMV.
$^\dagger$Seed 42 is shown for reference; mean and SD exclude seed 42.}
\label{tab:roberta_seeds_cmv}
\end{table}

Table~\ref{tab:roberta_seeds_cmv} reports the corresponding results on CMV. Accuracy is comparatively stable across the 10 runs ($0.510 \pm 0.022$), while recall ($0.590 \pm 0.256$) and F1 ($0.528 \pm 0.148$) exhibit considerably greater variation.

\begin{table}[t]
\centering
\small
\setlength{\tabcolsep}{5pt}
\begin{tabular}{lrrrr}
\toprule
Seed & Acc & Precision & Recall & F1 \\
\midrule
42$^\dagger$ & 0.5180 & 0.5380 & 0.1420 & 0.2250 \\
\midrule
2111 & 0.5100 & 0.5015 & 0.8376 & 0.6274 \\
2203 & 0.5350 & 0.5307 & 0.4822 & 0.5053 \\
2309 & 0.4650 & 0.4536 & 0.4213 & 0.4368 \\
2411 & 0.5150 & 0.5366 & 0.1117 & 0.1849 \\
2503 & 0.5825 & 0.5620 & 0.6904 & 0.6196 \\
2609 & 0.5075 & 0.0000 & 0.0000 & 0.0000 \\
2707 & 0.6625 & 0.6890 & 0.5736 & 0.6260 \\
2801 & 0.6200 & 0.6257 & 0.5685 & 0.5957 \\
2903 & 0.5325 & 0.5676 & 0.2132 & 0.3100 \\
3001 & 0.4975 & 0.4902 & 0.5076 & 0.4988 \\
\midrule
Mean (10 runs) & 0.5427 & 0.4957 & 0.4406 & 0.4405 \\
SD (10 runs)   & 0.0608 & 0.1869 & 0.2613 & 0.2133 \\
\bottomrule
\end{tabular}
\caption{Seed-level \model{RoBERTa} performance on IBM.
$^\dagger$Seed 42 is shown for reference; mean and SD exclude seed 42.}
\label{tab:roberta_seeds_ibm}
\end{table}

As shown in Table~\ref{tab:roberta_seeds_ibm}, IBM also exhibits substantial seed sensitivity, particularly for recall ($0.441 \pm 0.261$) and F1 ($0.440 \pm 0.213$). One run yields zero positive-class precision, recall, and F1, illustrating the instability that can arise from individual fine-tuning runs.

\begin{table}[t]
\centering
\small
\setlength{\tabcolsep}{5pt}
\begin{tabular}{lrrrr}
\toprule
Seed & Acc & Precision & Recall & F1 \\
\midrule
42$^\dagger$ & 0.6330 & 0.4320 & 0.3730 & 0.4000 \\
\midrule
3109 & 0.6010 & 0.3129 & 0.1804 & 0.2289 \\
3203 & 0.3784 & 0.3328 & 0.8902 & 0.4845 \\
3301 & 0.4723 & 0.3159 & 0.5216 & 0.3935 \\
3407 & 0.6281 & 0.3692 & 0.1882 & 0.2494 \\
3511 & 0.3848 & 0.3368 & 0.9020 & 0.4904 \\
3607 & 0.5830 & 0.3480 & 0.3098 & 0.3278 \\
3701 & 0.4659 & 0.3319 & 0.6196 & 0.4323 \\
3803 & 0.6731 & 0.5062 & 0.1608 & 0.2440 \\
3907 & 0.5508 & 0.3454 & 0.4118 & 0.3757 \\
4001 & 0.6589 & 0.3750 & 0.0588 & 0.1017 \\
\midrule
Mean (10 runs) & 0.5396 & 0.3574 & 0.4243 & 0.3328 \\
SD (10 runs)   & 0.1083 & 0.0560 & 0.3026 & 0.1258 \\
\bottomrule
\end{tabular}
\caption{Seed-level \model{RoBERTa} performance on Anthropic.
$^\dagger$Seed 42 is shown for reference; mean and SD exclude seed 42.}
\label{tab:roberta_seeds_anthropic}
\end{table}

Table~\ref{tab:roberta_seeds_anthropic} shows substantial variation on Anthropic, particularly in recall ($0.424 \pm 0.303$), with accuracy of $0.540 \pm 0.108$ and F1 of $0.333 \pm 0.126$ across the 10 additional runs.

Overall, these results show that the fine-tuned \model{RoBERTa} baseline can be sensitive to random initialization and training order, with the degree of variation differing across datasets.

\section{Full regression coefficients}

Tables \ref{tab:ukp_regression}--~\ref{tab:anth_regression} show all logistic regression coefficients for UKP, CMV, IBM, and Anthropic.

\subsection{Full Coefficients by Topic}
\label{appendix:results-ibm-topic}

\begin{figure}[t]
  \includegraphics[width=\columnwidth]{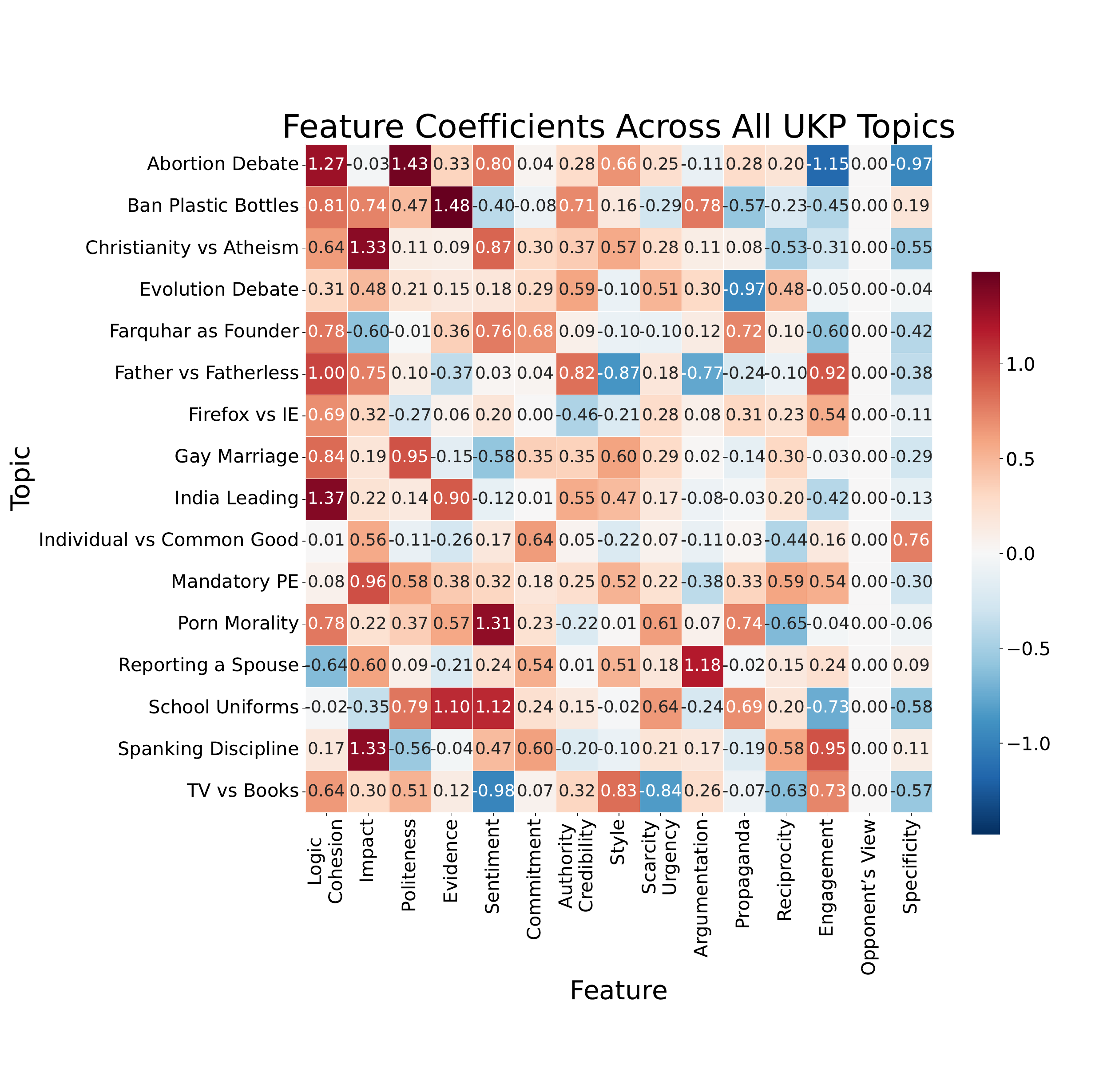}
  \caption{Coefficient Values of all features in UKP across different topics}
  \label{fig:ukp_coefficient_topic}
\end{figure}

Figure~\ref{fig:ukp_coefficient_topic} shows topic-level coefficient patterns for all 16 UKP topics. Several cross-topic regularities are visible. \pidim{Logic and Cohesion} and \pidim{Evidence} are among the most consistently positive predictors, appearing with moderate to strong positive coefficients across the majority of topics, reinforcing their role as stable Logos-oriented signals. \pidim{Sentiment} is broadly positive but shows greater variance, peaking in \textit{Porn Morality} (1.31), \textit{School Uniforms} (1.12), and \textit{Abortion Debate} (0.80), while near-zero or negative in others. \pidim{Engagement} is predominantly negative across topics, most strongly in \textit{TV vs Books} (-0.84) and \textit{Father vs Fatherless} (-0.87), consistent with the cross-dataset pattern observed in Section~\ref{result-dataset}. \pidim{Opponent's View} shows near-zero coefficients across all topics, reflecting its rarity as a rhetorical move in this corpus. Topic-specific patterns also emerge: \textit{Reporting a Spouse} is uniquely driven by \textit{Scarcity and Urgency} (1.18), \textit{Abortion Debate} by \pidim{Politeness} (1.43), and \textit{Individual vs Common Good} by \pidim{Specificity} (0.76), suggesting that the rhetorical demands of each topic activate distinct persuasion dimensions beyond the cross-topic baseline.

\begin{figure}[t]
  \includegraphics[width=\columnwidth]{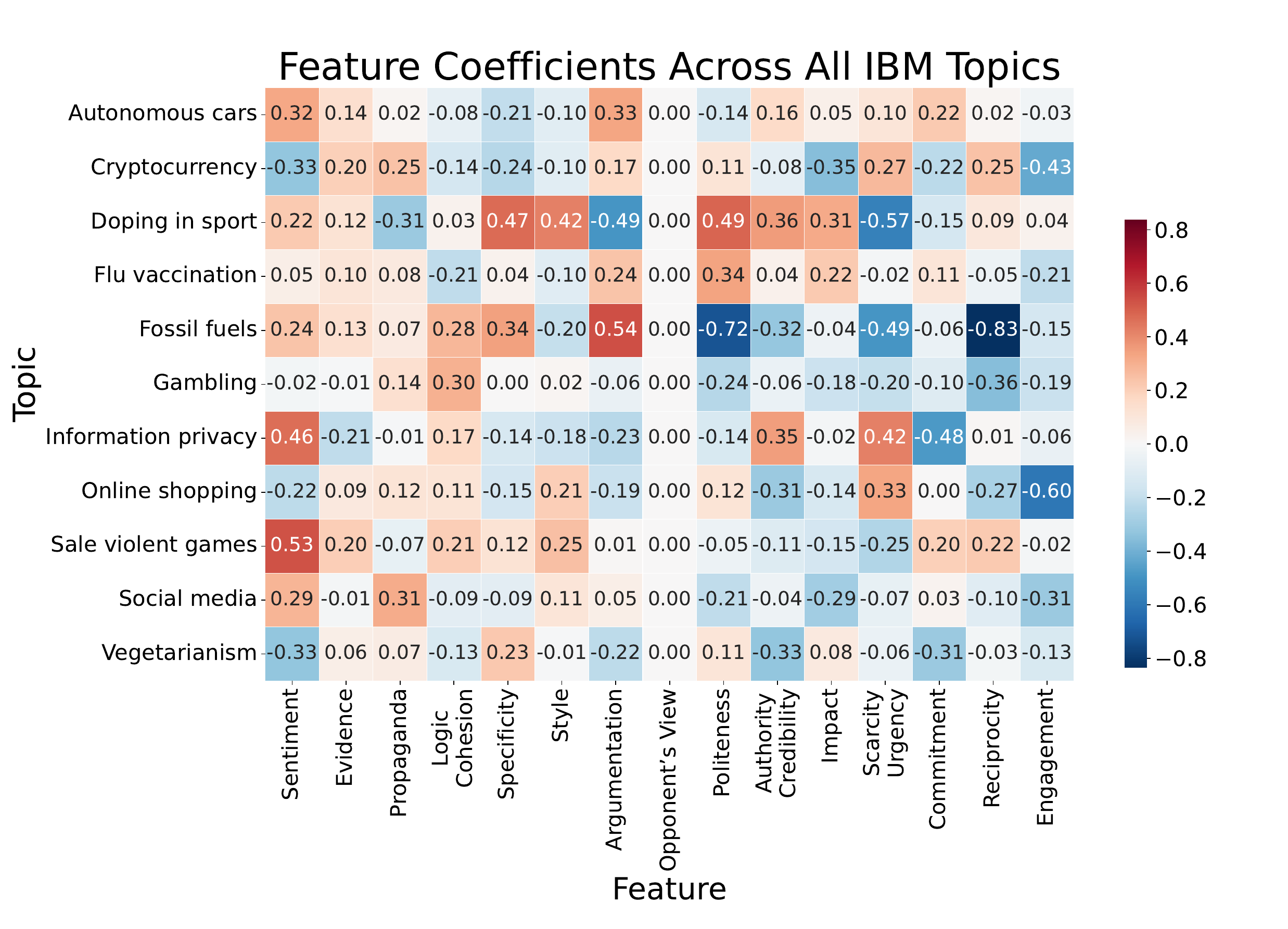}
  \caption{Coefficient Values of all features in IBM across different topics}
  \label{fig:ibm_coefficient_topic}
\end{figure}

Figure~\ref{fig:ibm_coefficient_topic} shows topic-level coefficient patterns for all 11 IBM topics. Unlike UKP, where \pidim{Evidence} and \pidim{Logic and Cohesion} dominate broadly, IBM topics show more heterogeneous persuasion signatures with no single dimension consistently predictive across topics. \pidim{Sentiment} shows the most variable pattern, strongly positive for \textit{Sale of Violent Games} (0.53) and \textit{Information Privacy} (0.46) but negative for \textit{Cryptocurrency} (-0.33) and \textit{Vegetarianism} (-0.33). \textit{Fossil Fuels} is driven by \pidim{Argumentation} (0.54) with strongly negative \pidim{Commitment} (-0.83) and \pidim{Politeness} (-0.72), suggesting that on contested scientific policy, explicit claim structure matters while conciliatory framing does not. \textit{Doping in Sport} similarly relies on \pidim{Argumentation} (0.42) and \pidim{Sentiment} (0.47) but penalizes \pidim{Commitment} (-0.57), consistent with a topic where bold claims outweigh personal pledges. \textit{Information Privacy} shows a distinctive pattern with strong positive \pidim{Sentiment} (0.46), \pidim{Reciprocity} (0.42), and \pidim{Authority and Credibility} (0.35), reflecting appeals to personal rights and institutional trust. \pidim{Opponent's View} shows near-zero coefficients across all topics, consistent with the UKP pattern. Together, these results reinforce that persuasive strategies are highly topic-dependent, and demonstrate PI's capacity to surface fine-grained rhetorical variation in an interpretable and auditable way.

\section{Feature Correlations}
\begin{figure}[htbp!]
  \includegraphics[width=\columnwidth]{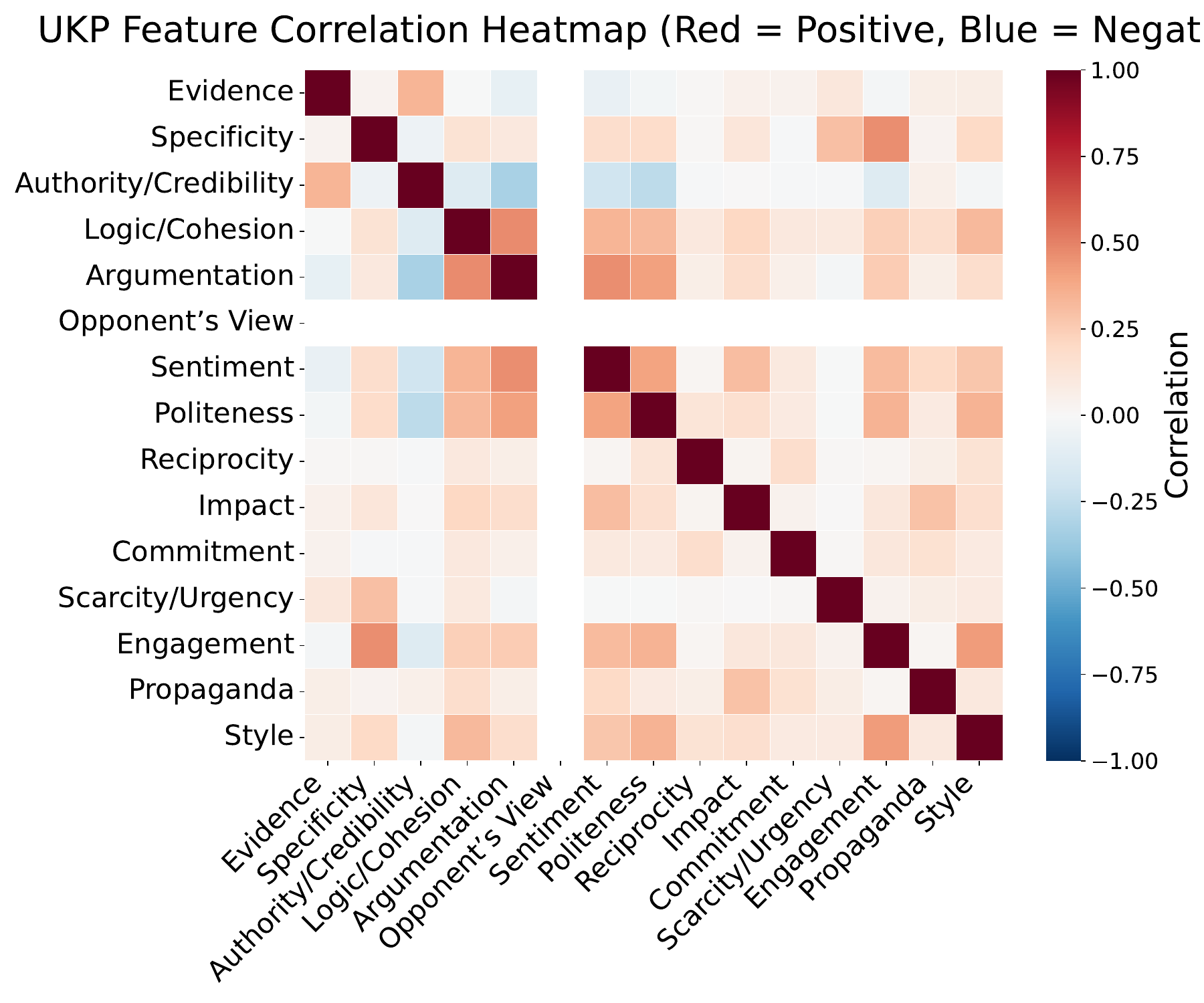}
  \caption{Pairwise Pearson correlations between the 15 PI category-mean scores on the UKP dataset.}
  \label{fig:ukp_correlation_mean}
\end{figure}

Figure~\ref{fig:ukp_correlation_mean} shows correlations between the 15 PI category-mean scores on UKP. Two clear correlation clusters emerge, broadly corresponding to the Logos and Pathos groupings. Within Logos, \pidim{Evidence}, \pidim{Logic and Cohesion}, and \pidim{Argumentation} are positively correlated, reflecting that well-structured arguments tend to deploy factual grounding and explicit reasoning jointly. Within Pathos, \pidim{Sentiment}, \pidim{Reciprocity}, \pidim{Impact}, and \pidim{Engagement} form a second positive cluster, consistent with affective and relational appeals co-occurring in persuasive text. Notably, the two clusters show weak to negligible cross-cluster correlations, suggesting that Logos- and Pathos-oriented rhetorical strategies are largely orthogonal in this corpus. \pidim{Opponent's View} is isolated with near-zero correlations to all other dimensions, reflecting its rarity as a rhetorical move.

\begin{figure*}[t]
  \includegraphics[width=2\columnwidth]{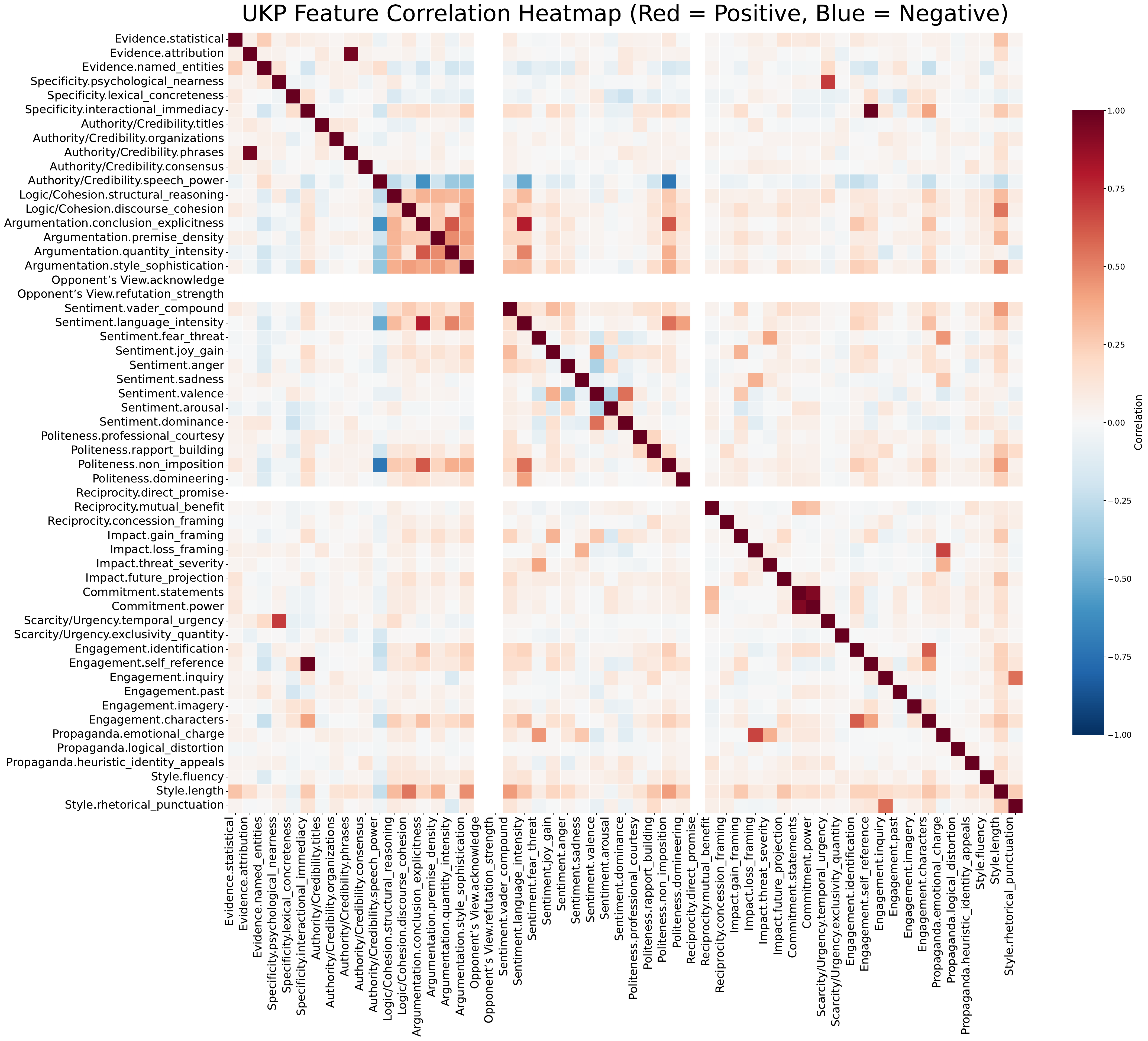}
  \caption{Pairwise Pearson correlations between the 55 PI sub-features on the UKP dataset.}
  \label{fig:ukp_correlation_sub}
\end{figure*}

Figure~\ref{fig:ukp_correlation_sub} reveals finer-grained structure at the sub-feature level. Within-dimension correlations are strong by construction for some dimensions (e.g., \pidim{Sentiment} sub-features cluster tightly), but several cross-dimension correlations are also visible. \pidim{Authority and Credibility.speech\_power} shows a notable negative correlation with \pidim{Argumentation} sub-features, suggesting that powerless language co-occurs with sparser argumentative structure. \pidim{Engagement} sub-features (\pisub{self\_reference}, \pisub{past}, \pisub{characters}) form a tight positive cluster, consistent with their shared reliance on personal pronoun and narrative language. The overall sparsity of strong off-diagonal correlations supports the discriminant validity of the PI sub-feature space — most sub-features capture distinct rhetorical signals.

\section{Robustness of Coefficient Interpretation}
\label{appendix:robustness}

To assess whether PI coefficients can be interpreted as estimates of rhetorical importance rather than artifacts of correlated features, we conduct two robustness analyses on UKP: (1) multicollinearity diagnostics via Variance Inflation Factors (VIFs) across the 15 dimensions, and (2) leave-one-out ablations that quantify the marginal contribution of each dimension to predictive performance.

\subsection{Multicollinearity Diagnostics}
\label{appendix:vif}

We compute VIFs by treating each dimension in turn as the outcome variable in an auxiliary regression against the remaining 14 dimensions, after standardizing the design matrix to mean zero and unit variance. By convention, VIF $> 10$ indicates problematic multicollinearity, while VIF $> 5$ warrants attention. Table~\ref{tab:vif} reports per-dimension VIFs on UKP.

\begin{table}[h]
\centering
\small
\begin{tabular}{lr}
\toprule
Dimension & VIF \\
\midrule
Style                  & 27.39 \\
Sentiment              & 22.64 \\
Logic/Cohesion         & 14.09 \\
Argumentation          &  8.36 \\
Engagement             &  7.39 \\
Specificity            &  4.61 \\
Politeness             &  4.19 \\
Evidence               &  2.64 \\
Authority/Credibility  &  2.55 \\
Propaganda             &  2.10 \\
Scarcity/Urgency       &  1.96 \\
Impact                 &  1.89 \\
Commitment             &  1.18 \\
Reciprocity            &  1.17 \\
Opponent's View        &  --   \\
\midrule
\textbf{Median}        &  3.41 \\
\textbf{Max}           & 27.39 \\
\bottomrule
\end{tabular}
\caption{Per-dimension VIFs on UKP. Opponent's View is omitted (NaN) because its sub-feature scores are near-constant on this dataset.}
\label{tab:vif}
\end{table}

The median VIF of 3.41 indicates that most dimensions carry largely independent variance, but three dimensions exceed the conventional threshold of 10: \pidim{Style} (27.4), \pidim{Sentiment} (22.6), and \pidim{Logic and Cohesion} (14.1). These elevated values reflect interpretable structural properties of the framework rather than redundancy in the taxonomy:

\begin{itemize}
\itemsep0em
    \item \pidim{Style} aggregates surface markers including \texttt{length}, which correlates with the rate-based sub-features in every other dimension that longer arguments accumulate more matches across the framework as a whole.
    \item \pidim{Sentiment} aggregates nine partially overlapping sub-features spanning polarity (VADER), discrete emotions (anger, sadness, fear, joy), and dimensional affect (valence, arousal, dominance from NRC-VAD), with deliberate redundancy across these complementary affective scales.
    \item \pidim{Logic and Cohesion} relies on connective-density sub-features that scale with text length, producing partial correlation with \pidim{Style} and other length-sensitive dimensions.
\end{itemize}

Two design choices mitigate the practical impact of these correlations on coefficient interpretation. First, our $\ell_2$-regularized logistic regression explicitly shrinks correlated coefficients toward zero, stabilizing estimates in the regime where multicollinearity would otherwise produce instability. Second, the 95\% Wald confidence intervals reported in Figures~\ref{fig:mean_coefficient_8_comparison} and \ref{fig:ibm_stance_features_comparison} explicitly reflect the standard-error inflation that VIF would predict; dimensions with high VIFs naturally exhibit wider CIs in our reported figures, allowing readers to calibrate confidence in individual estimates directly.

We therefore interpret PI coefficients at the \emph{dimension} level as our primary analytic claim. Sub-feature-level interpretation, where some sub-features (e.g., \texttt{length} under \pidim{Style}, \texttt{named\_entities} under \pidim{Evidence}) may carry surface or topicality confounds in addition to rhetorical signal, requires the more targeted detectors that PI's modular design is intended to support (Section~\ref{sec:methods}). 

\subsection{Leave-One-Out Dimension Ablations}
\label{appendix:ablations}

\begin{figure*}[t]
  \centering
  \includegraphics[width=\textwidth]{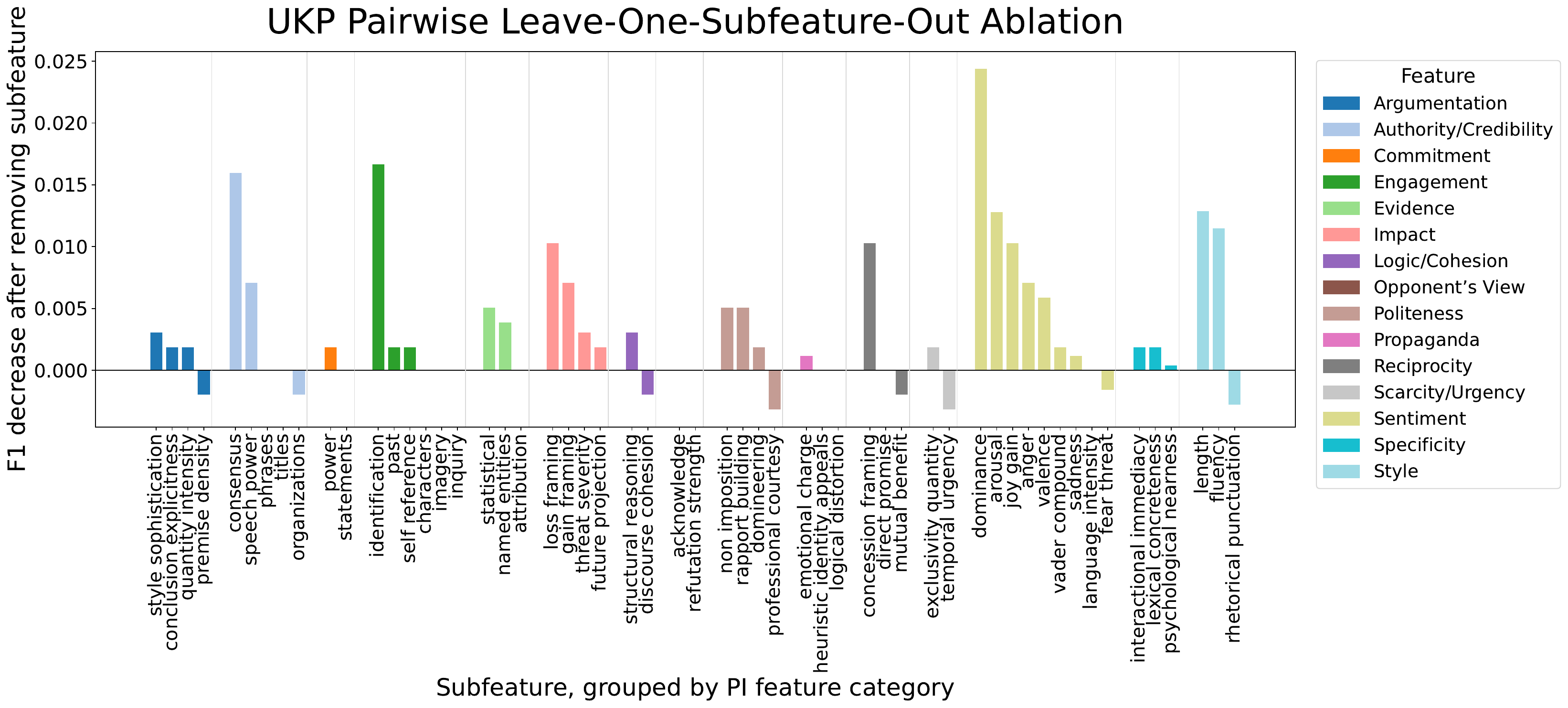}
  \caption{Leave-one-subfeature-out ablation results on UKP. Bars show the change in F1 after removing each subfeature and refitting the PI-sub model on the remaining features. Subfeatures are grouped by their corresponding PI dimensions. Positive values indicate a decrease in F1 after removal, whereas negative values indicate an improvement.}
  \label{fig:sub_ablation}
\end{figure*}

To complement the multicollinearity diagnostics with a behavioral measure of dimension importance, we conduct leave-one-out ablations at both the dimension and subfeature levels. At the dimension level, we remove each of the 15 dimensions in turn, dropping all of its constituent sub-features, refit the regression on the reduced feature matrix, and measure the resulting change in accuracy, F1, and AUC on UKP. At the subfeature level, we repeat the procedure for each of the 55 individual subfeatures, removing one at a time while retaining the remaining 54. Table~\ref{tab:ablation} and~\ref{tab:ablation_sub} report the dimension- and subfeature-level results, sorted by $\Delta$AUC (most damaging first). Figure~\ref{fig:sub_ablation} provides a more fine-grained view of these effects.

\begin{table*}[h]
\centering
\small
\begin{tabular}{lrrrrrrr}
\toprule
Removed Dimension & \# Sub-features & Acc & F1 & AUC & $\Delta$Acc & $\Delta$F1 & $\Delta$AUC \\
\midrule
Style                  & 3 & 0.7450 & 0.7358 & 0.8299 & $-$0.0300 & $-$0.0322 & $-$0.0281 \\
Sentiment              & 9 & 0.7375 & 0.7273 & 0.8352 & $-$0.0375 & $-$0.0407 & $-$0.0228 \\
Engagement             & 6 & 0.7750 & 0.7692 & 0.8426 &    0.0000 &    0.0012 & $-$0.0154 \\
Authority/Credibility  & 5 & 0.7725 & 0.7624 & 0.8538 & $-$0.0025 & $-$0.0056 & $-$0.0042 \\
Propaganda             & 3 & 0.7775 & 0.7700 & 0.8565 &    0.0025 &    0.0020 & $-$0.0015 \\
Reciprocity            & 3 & 0.7700 & 0.7617 & 0.8571 & $-$0.0050 & $-$0.0063 & $-$0.0009 \\
Impact                 & 4 & 0.7700 & 0.7641 & 0.8577 & $-$0.0050 & $-$0.0039 & $-$0.0003 \\
Evidence               & 3 & 0.7625 & 0.7595 & 0.8578 & $-$0.0125 & $-$0.0085 & $-$0.0002 \\
Commitment             & 2 & 0.7700 & 0.7641 & 0.8580 & $-$0.0050 & $-$0.0039 &    0.0000 \\
Opponent's View        & 2 & 0.7750 & 0.7680 & 0.8580 &    0.0000 &    0.0000 &    0.0000 \\
Logic/Cohesion         & 2 & 0.7725 & 0.7649 & 0.8585 & $-$0.0025 & $-$0.0031 &    0.0005 \\
Scarcity/Urgency       & 2 & 0.7750 & 0.7680 & 0.8588 &    0.0000 &    0.0000 &    0.0008 \\
Argumentation          & 4 & 0.7750 & 0.7692 & 0.8598 &    0.0000 &    0.0012 &    0.0018 \\
Politeness             & 4 & 0.7700 & 0.7641 & 0.8629 & $-$0.0050 & $-$0.0039 &    0.0049 \\
Specificity            & 3 & 0.7850 & 0.7772 & 0.8651 &    0.0100 &    0.0092 &    0.0071 \\
\midrule
\textbf{Full PI-sub}           & 55 & 0.7750 & 0.7680 & 0.8580 & --- & --- & --- \\
\bottomrule
\end{tabular}
\caption{Leave-one-out ablations on UKP. Each row reports performance when the indicated dimension is removed and the regression is refit on the remaining feature matrix. Sorted by $\Delta$AUC (most damaging removal first).}
\label{tab:ablation}
\end{table*}

\begin{table*}[!p]
\centering
\small
\setlength{\tabcolsep}{3pt}
\begin{tabular}{llrrrrrr}
\toprule
Removed Dimension & Removed Subfeature & Acc & F1 & AUC & $\Delta$Acc & $\Delta$F1 & $\Delta$AUC \\
\midrule
Style                  & length                       & 0.7600 & 0.7551 & 0.8297 & $-$0.0150 & $-$0.0129 & $-$0.0283 \\
Engagement             & identification               & 0.7550 & 0.7513 & 0.8433 & $-$0.0200 & $-$0.0167 & $-$0.0147 \\
Sentiment              & dominance                    & 0.7500 & 0.7436 & 0.8513 & $-$0.0250 & $-$0.0244 & $-$0.0067 \\
Sentiment              & anger                        & 0.7675 & 0.7609 & 0.8551 & $-$0.0075 & $-$0.0071 & $-$0.0029 \\
Style                  & rhetorical\_punctuation      & 0.7800 & 0.7708 & 0.8554 &    0.0050 &    0.0028 & $-$0.0026 \\
Style                  & fluency                      & 0.7650 & 0.7565 & 0.8559 & $-$0.0100 & $-$0.0115 & $-$0.0021 \\
Sentiment              & arousal                      & 0.7650 & 0.7552 & 0.8562 & $-$0.0100 & $-$0.0128 & $-$0.0018 \\
Politeness             & non\_imposition              & 0.7700 & 0.7629 & 0.8562 & $-$0.0050 & $-$0.0051 & $-$0.0018 \\
Authority/Credibility  & consensus                    & 0.7625 & 0.7520 & 0.8565 & $-$0.0125 & $-$0.0160 & $-$0.0015 \\
Sentiment              & joy\_gain                    & 0.7650 & 0.7577 & 0.8565 & $-$0.0100 & $-$0.0103 & $-$0.0015 \\
Authority/Credibility  & speech\_power                & 0.7675 & 0.7609 & 0.8565 & $-$0.0075 & $-$0.0071 & $-$0.0015 \\
Propaganda             & logical\_distortion          & 0.7750 & 0.7680 & 0.8566 &    0.0000 &    0.0000 & $-$0.0014 \\
Engagement             & characters                   & 0.7750 & 0.7680 & 0.8570 &    0.0000 &    0.0000 & $-$0.0010 \\
Sentiment              & sadness                      & 0.7750 & 0.7668 & 0.8573 &    0.0000 & $-$0.0012 & $-$0.0007 \\
Impact                 & threat\_severity             & 0.7725 & 0.7649 & 0.8574 & $-$0.0025 & $-$0.0031 & $-$0.0006 \\
Argumentation          & style\_sophistication        & 0.7725 & 0.7649 & 0.8575 & $-$0.0025 & $-$0.0031 & $-$0.0005 \\
Reciprocity            & concession\_framing          & 0.7650 & 0.7577 & 0.8576 & $-$0.0100 & $-$0.0103 & $-$0.0004 \\
Propaganda             & emotional\_charge            & 0.7750 & 0.7668 & 0.8576 &    0.0000 & $-$0.0012 & $-$0.0004 \\
Authority/Credibility  & titles                       & 0.7750 & 0.7680 & 0.8577 &    0.0000 &    0.0000 & $-$0.0003 \\
Engagement             & imagery                      & 0.7750 & 0.7680 & 0.8577 &    0.0000 &    0.0000 & $-$0.0003 \\
Impact                 & gain\_framing                & 0.7675 & 0.7609 & 0.8578 & $-$0.0075 & $-$0.0071 & $-$0.0002 \\
Scarcity/Urgency       & exclusivity\_quantity        & 0.7725 & 0.7661 & 0.8578 & $-$0.0025 & $-$0.0019 & $-$0.0002 \\
Logic/Cohesion         & discourse\_cohesion          & 0.7775 & 0.7700 & 0.8578 &    0.0025 &    0.0020 & $-$0.0002 \\
Evidence               & statistical                  & 0.7700 & 0.7629 & 0.8579 & $-$0.0050 & $-$0.0051 & $-$0.0001 \\
Authority/Credibility  & phrases                      & 0.7750 & 0.7680 & 0.8579 &    0.0000 &    0.0000 & $-$0.0001 \\
Propaganda             & heuristic\_identity\_appeals & 0.7750 & 0.7680 & 0.8579 &    0.0000 &    0.0000 & $-$0.0001 \\
Authority/Credibility  & organizations                & 0.7775 & 0.7700 & 0.8579 &    0.0025 &    0.0020 & $-$0.0001 \\
Argumentation          & quantity\_intensity          & 0.7725 & 0.7661 & 0.8580 & $-$0.0025 & $-$0.0019 &    0.0000 \\
Evidence               & attribution                  & 0.7750 & 0.7680 & 0.8580 &    0.0000 &    0.0000 &    0.0000 \\
Opponent's View        & acknowledge                  & 0.7750 & 0.7680 & 0.8580 &    0.0000 &    0.0000 &    0.0000 \\
Opponent's View        & refutation\_strength         & 0.7750 & 0.7680 & 0.8580 &    0.0000 &    0.0000 &    0.0000 \\
Reciprocity            & direct\_promise              & 0.7750 & 0.7680 & 0.8580 &    0.0000 &    0.0000 &    0.0000 \\
Reciprocity            & mutual\_benefit              & 0.7775 & 0.7700 & 0.8580 &    0.0025 &    0.0020 &    0.0000 \\
Evidence               & named\_entities              & 0.7700 & 0.7641 & 0.8581 & $-$0.0050 & $-$0.0039 &    0.0001 \\
Argumentation          & conclusion\_explicitness     & 0.7725 & 0.7661 & 0.8582 & $-$0.0025 & $-$0.0019 &    0.0002 \\
Politeness             & domineering                  & 0.7725 & 0.7661 & 0.8582 & $-$0.0025 & $-$0.0019 &    0.0002 \\
Impact                 & loss\_framing                & 0.7650 & 0.7577 & 0.8585 & $-$0.0100 & $-$0.0103 &    0.0005 \\
Engagement             & self\_reference              & 0.7725 & 0.7661 & 0.8585 & $-$0.0025 & $-$0.0019 &    0.0005 \\
Sentiment              & vader\_compound              & 0.7725 & 0.7661 & 0.8585 & $-$0.0025 & $-$0.0019 &    0.0005 \\
Commitment             & power                        & 0.7725 & 0.7661 & 0.8586 & $-$0.0025 & $-$0.0019 &    0.0006 \\
Specificity            & lexical\_concreteness        & 0.7725 & 0.7661 & 0.8586 & $-$0.0025 & $-$0.0019 &    0.0006 \\
Sentiment              & fear\_threat                 & 0.7800 & 0.7696 & 0.8587 &    0.0050 &    0.0016 &    0.0007 \\
Logic/Cohesion         & structural\_reasoning        & 0.7725 & 0.7649 & 0.8588 & $-$0.0025 & $-$0.0031 &    0.0008 \\
Impact                 & future\_projection           & 0.7725 & 0.7661 & 0.8588 & $-$0.0025 & $-$0.0019 &    0.0008 \\
Engagement             & inquiry                      & 0.7750 & 0.7680 & 0.8588 &    0.0000 &    0.0000 &    0.0008 \\
Commitment             & statements                   & 0.7750 & 0.7680 & 0.8589 &    0.0000 &    0.0000 &    0.0009 \\
Scarcity/Urgency       & temporal\_urgency            & 0.7775 & 0.7712 & 0.8589 &    0.0025 &    0.0032 &    0.0009 \\
Engagement             & past                         & 0.7725 & 0.7661 & 0.8590 & $-$0.0025 & $-$0.0019 &    0.0010 \\
Sentiment              & language\_intensity          & 0.7750 & 0.7680 & 0.8590 &    0.0000 &    0.0000 &    0.0010 \\
Specificity            & interactional\_immediacy     & 0.7725 & 0.7661 & 0.8591 & $-$0.0025 & $-$0.0019 &    0.0011 \\
Politeness             & rapport\_building            & 0.7700 & 0.7629 & 0.8601 & $-$0.0050 & $-$0.0051 &    0.0021 \\
Sentiment              & valence                      & 0.7675 & 0.7621 & 0.8605 & $-$0.0075 & $-$0.0059 &    0.0025 \\
Argumentation          & premise\_density             & 0.7775 & 0.7700 & 0.8605 &    0.0025 &    0.0020 &    0.0025 \\
Politeness             & professional\_courtesy       & 0.7775 & 0.7712 & 0.8618 &    0.0025 &    0.0032 &    0.0038 \\
Specificity            & psychological\_nearness      & 0.7775 & 0.7676 & 0.8631 &    0.0025 & $-$0.0004 &    0.0051 \\
\midrule
\textbf{Full PI-sub}   & ---                          & 0.7750 & 0.7680 & 0.8580 & --- & --- & --- \\
\bottomrule
\end{tabular}
\caption{Leave-one-subfeature-out ablations on UKP. Each row reports performance when the indicated subfeature is removed and the regression is refit on the remaining 54-subfeature feature matrix. Sorted by $\Delta$AUC (most damaging removal first).}
\label{tab:ablation_sub}
\end{table*}

At the dimension level, the ablation results converge with the coefficient analysis (Section~\ref{sec:results}). \pidim{Style} (--0.028 AUC) and \pidim{Sentiment} (--0.023 AUC) are the two most consequential dimensions for predictive performance on UKP, consistent with their large coefficient magnitudes and their elevated VIFs, where they carry both unique and shared variance with downstream persuasion judgments. \pidim{Engagement} ranks third in importance ($\Delta$AUC = --0.015) despite its consistently negative coefficient, suggesting that the dimension carries genuine discriminative signal even where its sign is interpretable as a negative predictor of persuasion in our corpora.

The subfeature-level results provide a more fine-grained view of these effects. The three most consequential individual subfeatures are \textit{length} within \pidim{Style} ($\Delta$AUC = $-$0.028), \textit{identification} within \pidim{Engagement} ($-$0.015), and \textit{dominance} within \pidim{Sentiment} ($-$0.007). Notably, these subfeatures belong to the same three dimensions that show the largest performance decreases in the dimension-level ablation. The effect for \textit{length} is particularly pronounced: removing this subfeature alone produces nearly the same AUC decrease as removing the entire \pidim{Style} dimension, suggesting that much of Style's predictive contribution on UKP is concentrated in this feature. In contrast, the effect of \pidim{Sentiment} is more distributed across its constituent subfeatures: although \textit{dominance} produces the largest individual decrease, its effect is substantially smaller than the decrease observed when the full Sentiment dimension is removed.

Most remaining dimensions and subfeatures have comparatively small marginal effects. At the dimension level, the other twelve dimensions each have $|\Delta\mathrm{AUC}| \leq 0.007$. At the subfeature level, all but three features have $|\Delta\mathrm{AUC}| < 0.007$, and many produce changes near zero. Several removals slightly improve AUC, including \textit{psychological\_nearness}, \textit{professional\_courtesy}, and \textit{premise\_density}. Such improvements do not necessarily indicate that these features are theoretically uninformative. In a correlated feature space, they may contribute information that overlaps with other predictors or introduce small amounts of dataset-specific noise.

We therefore interpret the leave-one-out results as measures of marginal predictive contribution within the UKP model rather than as a universal ranking of rhetorical importance. PI is designed as a comprehensive theoretical framework rather than a minimal predictive feature set, and features with small or redundant contributions on UKP may remain informative in other domains, persuasion settings, or audience conditions. The dimension- and subfeature-level analyses together show that predictive importance is heterogeneous both across dimensions and within them, while retaining the broader taxonomy enables PI to support interpretation beyond this individual prediction task.

\subsection{Interpretation and Caveats}

Taken together, the VIF and ablation analyses support three claims about PI coefficient interpretation:

\begin{enumerate}[leftmargin=*,nosep]
\itemsep0em
    \item Coefficient estimates are stable enough at the dimension level to support comparative interpretation across datasets, topics, and stances, particularly for dimensions with VIF $< 10$, which covers twelve of fifteen dimensions.
    \item For the three dimensions with elevated VIFs (\pidim{Style}, \pidim{Sentiment}, \pidim{Logic and Cohesion}), individual coefficients should be read with the wider confidence intervals reported in the main figures, and conclusions should rely on the joint behavior of the dimension across multiple datasets rather than a single point estimate.
    \item Sub-feature-level interpretation requires more caution than dimension-level interpretation. Sub-features such as \texttt{length}, \texttt{named\_entities}, and frequency-based connective counts may capture verbosity or topicality confounds in addition to rhetorical signal. PI's modular design is explicitly intended to support upgrading these sub-features with confound-resistant detectors without altering the dimension-level taxonomy or its theoretical anchors.
\end{enumerate}

Bootstrap confidence intervals on PI coefficients and group-lasso analyses penalizing each Aristotelian triad as a unit are natural extensions of these checks, which we leave for future work.

\section{Lexicon Construction and Expansion}
\label{appendix:llm_expansion}

Here we provide full details for lexicon construction and validation.

\mypar{Step 1: Manual Seed Lexicon Construction.} 
We construct small seed lexicons based on foundational persuasion theories (Table~\ref{tab:pi_categories}). For example, the \pisub{Causal} sub-feature of the \pidim{Logic and Cohesion} dimension has a seed lexicon containing causal connectives (``because'', ``therefore'') drawn from argumentation theory \citep{toulminUsesArgument2003}. Example seeds for all linguistic cues can be found in Table~\ref{tab:pi_methods_one_cue}, and full seed and expanded lexicons are available at \url{https://github.com/krystalgong/Persuasion\_Index\_Code}.

\mypar{Step 2: Manual Seed Validation.}
We conduct a three-stage validation of the seed lexicons to ensure construct validity. First, we review each seed lexicon by rating the fit of every term against its theoretical definition and removing terms with low face validity. Second, we run the seed-based PI on the datasets and examine the distribution and frequency of each feature to flag implementation errors or incomplete lexicons. Third, we inspect sub-feature scoring at the top and bottom for each manipulated argument. We examine high-scoring arguments to confirm the detected signals reflect the intended persuasion dimension and inspect low-scoring arguments to identify missing patterns.

\mypar{Step 3: LLM Expansion.} 
Expansion begins with a structured prompt encoding the sub-feature label, theoretical anchor, and explicit inclusion and exclusion criteria derived from Steps 1 and 2 (example in Appendix~\ref{appendix:llm_expansion-category_definition}). Seven register- and morphology-conditioned prompts are issued to \model{GPT-5.4-mini} per lexicon category, targeting: formal single tokens, formal phrases, colloquial single tokens, colloquial phrases, domain-specific items (legal, academic, political), archaic or literary forms, and inflectional variants of canonical seeds. Since a single sub-feature may correspond to multiple lexicon categories (e.g., \pidim{Logic and Cohesion} covers separate categories for causal, contrastive, and additive connectives), each category is expanded independently with its own structured prompt (Appendix~\ref{appendix:llm_expansion-7slices_prompts}). Generated candidates are merged by lowercased surface forms and punctuation-based near-duplicates are collapsed. The model assigns a confidence level (high, medium, or low) to each candidate (prompt in Appendix~\ref{appendix:llm_expansion-expansion_prompts}).

\mypar{Step 4: Post-expansion Filtering.} 
Three trained annotators (authors of this work) independently conducted a pilot annotation of 218 sampled items stratified across model-generated confidence tiers. The task involved binary labeling for whether or not a lexical item belongs to the sub-feature for which it was generated. All disagreements were resolved via consensus-coding. Overall Krippendorff's $\alpha = 0.558$ across three annotators, with majority agreement (at least 2 of 3) on all 218 items (67.0\% unanimous) (Table~\ref{tab:iaa_confidence}). Rejection rates diverge substantially by confidence level: high-confidence items show a 5.4\% rejection rate while low-confidence items reach 67.6\%. Based on these rates, high-confidence items are admitted after a spot-check and low-confidence items are removed outright. For the 1,440 medium-confidence items, for which annotator agreement on the pilot sample was highest ($\alpha = 0.674$), remaining items were divided equally for independent annotation. Closed subfeature categories, such as ``by the way'', are dropped directly. The final lexicon takes the union of audited LLM-expanded candidates and the seed lexicon from Step 1.

\begin{table}[t]
\caption{Inter-annotator agreement by LLM confidence level}
\label{tab:iaa_confidence}
\resizebox{\columnwidth}{!}{%
\begin{tabular}{lrrrrrrr}
\toprule
Confidence & \% Excl. & Krippendorff's $\alpha$ & $N$ & All 3 Agree & Incl. & Excl. \\
\midrule
High & 5.4\%  & 0.344 & 52  & 36 & 44 & 8  \\
Low  & 67.6\% & 0.420 & 108 & 65 & 35 & 73 \\
Med  & 31.0\% & 0.674 & 58  & 45 & 40 & 18 \\
\bottomrule
\end{tabular}%
}
\end{table}

\mypar{Step 5: Final Lexicon Validation.} 
A split-half internal consistency analysis verifies lexicon robustness. Each sub-category lexicon is partitioned into two equal, non-overlapping halves. The corpus is then scored independently using each half. Features driven entirely by non-lexical resources are excluded to prevent inflated results. The correlation between aggregate mean vectors is $r=0.90$ ($p<0.001$) across 30 lexicon-driven sub-features. Across the 12 mean-features, the correlation reaches $r=0.97$ ($p<0.001$). These confirm that the final lexicons preserve stable feature scales across the full PI framework. Sub-feature internal consistency is shown in Table~\ref{tab:internal_consistency} 

\begin{table}[t]
\caption{Sub-feature internal consistency (lexicon-driven features only), sorted by Pearson $r$ (worst first). Computed on UKP ($n=1{,}052$) using two random half-splits.}
\label{tab:internal_consistency}
\resizebox{\columnwidth}{!}{%
\begin{tabular}{lrrr}
\toprule
Feature & $r$ & $p$ & MAD \\
\midrule
Scarcity/Urgency.exclusivity\_quantity & $-0.023$ & .460 & 0.157 \\
Propaganda.logical\_distortion        & $-0.002$ & .957 & 0.004 \\
Impact.loss\_framing                  &  0.012  & .700 & 0.073 \\
Commitment.statements                 &  0.027  & .389 & 0.079 \\
Reciprocity.concession\_framing       &  0.038  & .217 & 0.057 \\
Argumentation.conclusion\_explicitness&  0.067  & .029 & 0.369 \\
Impact.threat\_severity               &  0.079  & .011 & 0.021 \\
Authority/Credibility.phrases         &  0.089  & .004 & 0.080 \\
Impact.gain\_framing                  &  0.096  & .002 & 0.220 \\
Argumentation.premise\_density        &  0.097  & .002 & 0.452 \\
Propaganda.emotional\_charge          &  0.099  & .001 & 0.137 \\
Scarcity/Urgency.temporal\_urgency    &  0.101  & .001 & 0.155 \\
Authority/Credibility.speech\_power   &  0.103  & $<$.001 & 0.402 \\
Logic/Cohesion.structural\_reasoning  &  0.135  & $<$.001 & 0.413 \\
Politeness.professional\_courtesy     &  0.137  & $<$.001 & 0.075 \\
Propaganda.heuristic\_identity\_appeals& 0.163  & $<$.001 & 0.243 \\
Politeness.rapport\_building          &  0.165  & $<$.001 & 0.264 \\
Politeness.non\_imposition            &  0.166  & $<$.001 & 0.427 \\
Sentiment.language\_intensity         &  0.201  & $<$.001 & 0.362 \\
Specificity.psychological\_nearness   &  0.205  & $<$.001 & 0.226 \\
Reciprocity.mutual\_benefit           &  0.215  & $<$.001 & 0.031 \\
Argumentation.style\_sophistication   &  0.228  & $<$.001 & 0.217 \\
Evidence.attribution                  &  0.303  & $<$.001 & 0.075 \\
Commitment.power                      &  0.305  & $<$.001 & 0.038 \\
Argumentation.quantity\_intensity     &  0.314  & $<$.001 & 0.288 \\
Impact.future\_projection             &  0.332  & $<$.001 & 0.171 \\
Evidence.statistical                  &  0.343  & $<$.001 & 0.286 \\
Politeness.domineering                &  0.346  & $<$.001 & 0.183 \\
Logic/Cohesion.discourse\_cohesion    &  0.838  & $<$.001 & 0.069 \\
Sentiment.fear\_threat                &  0.844  & $<$.001 & 0.021 \\
\bottomrule
\end{tabular}%
}
\end{table}

\mypar{Step 6: Dimension-Level Human Validation.}
We additionally validate whether PI's computed dimension scores correspond to human judgments of the rhetorical constructs they are intended to measure. For each of the 15 dimensions, we randomly sample 20 high-scoring and 20 low-scoring texts and pair them to produce 20 high--low comparisons. The annotator, without access to the PI scores, judges which text in each pair more strongly exhibits the target dimension. We then measure agreement between the human judgment and the ordering implied by PI's scores.

\begin{table}[t]
\centering
\small
\setlength{\tabcolsep}{5pt}
\begin{tabular}{lrrr}
\toprule
Dimension & Matches & Pairs & Agreement \\
\midrule
Argumentation              & 19 & 20 & 95.0\% \\
Authority/Credibility      & 19 & 20 & 95.0\% \\
Commitment                 & 16 & 20 & 80.0\% \\
Engagement                 & 17 & 20 & 85.0\% \\
Evidence                   & 17 & 20 & 85.0\% \\
Impact                     & 18 & 20 & 90.0\% \\
Logic/Cohesion             & 17 & 20 & 85.0\% \\
Opponent's View$^\dagger$  & 5  & 9  & 55.6\% \\
Politeness                 & 20 & 20 & 100.0\% \\
Propaganda                 & 17 & 20 & 85.0\% \\
Reciprocity                & 16 & 20 & 80.0\% \\
Scarcity/Urgency           & 15 & 20 & 75.0\% \\
Sentiment                  & 20 & 20 & 100.0\% \\
Specificity                & 13 & 20 & 65.0\% \\
Style                      & 20 & 20 & 100.0\% \\
\midrule
Overall                    & 249 & 289 & 86.2\% \\
\bottomrule
\end{tabular}
\caption{Dimension-level human validation of PI scores. Agreement indicates the proportion of high--low pairs for which the human judgment matches the ordering implied by PI. $^\dagger$Only nine valid pairs were available for \pidim{Opponent's View} because sufficiently high-scoring examples were sparse.}
\label{tab:dimension_validation}
\end{table}

As shown in Table~\ref{tab:dimension_validation}, human judgments agree with PI's score ordering in 249 of 289 comparisons (86.2\%). Agreement is at least 75\% for 12 of the 15 dimensions and reaches 100\% for \pidim{Politeness}, \pidim{Sentiment}, and \pidim{Style}. \pidim{Argumentation} and \pidim{Authority and Credibility} each reach 95\%, and \pidim{Impact} reaches 90\%. The lowest agreement among dimensions with the full 20 comparisons is observed for \pidim{Specificity} (65\%).

\pidim{Opponent's View} is an exception because sufficiently high-scoring examples are rare in the sampled corpus. Only nine valid high--low pairs could therefore be constructed for this dimension, of which five (55.6\%) matched the human judgment. We report this result for completeness but interpret it cautiously given the smaller number of available comparisons. Overall, the results provide direct human validation that PI's dimension-level scores generally preserve the intended ordering of rhetorical constructs, complementing the split-half analysis of lexicon stability above.

\subsection{Category Definitions for LLM Expansion}
\label{appendix:llm_expansion-category_definition}
Below is one category definition example to expand the seed lexicon ``EVI\_UNITS''. Full category definitions are available in Supplementary Material.
\begin{lstlisting}
"EVI_UNITS": {
"parent_appeal": "Logos",
"parent_dimension": "Evidence",
"parent_subfeature": "statistical",
"function": "Units of measurement, statistical markers, currencies, and quantitative indicators that ground claims in verifiable numerical evidence.",
"theory_anchor": "Zebregs et al. (2015) on statistical evidence; targets cognitive belief change via factual grounding.",
"include": [
  "SI / imperial units of measure (kg, mi, ml, ft, C, mph)",
  "Currency symbols and codes (USD, EUR, \$, JPY)",
  "Statistical indicators (p-value, percentage, ratio, coefficient, significance)",
  "Magnitude words used in numerical context (million, billion, thousand)"
],
"exclude": [
  "Polysemous letters with non-measurement meanings in dominant English use",
  "Generic quantifiers without measurement function (some, many, several) those belong to SPEC_VAGUE",
  "Number words used as ordinals/names (first, second as ranks)"
],
"common_drift_traps": [
  {
    "word": "size",
    "drift_type": "HYPERNYM_DILUTION",
    "why": "Generic noun, not a measurement unit"
  },
  {
    "word": "many",
    "drift_type": "TOPICAL_NEIGHBOR",
    "why": "Vague quantifier; belongs to SPEC_VAGUE"
  }
],
"canonical_seeds": [
  "%",
  "kg",
  "million",
  "percentage",
  "p-value",
  "USD"
]
}
\end{lstlisting}

\subsection{Prompts for 7 slices}
\label{appendix:llm_expansion-7slices_prompts}
We broaden lexicon coverage by prompting with 7 distinct register ``slices''. Each slice is a configuration object substituted into the \texttt{[SLICE]} block of that template: \texttt{instructions} specifies the linguistic target (register, token granularity, and whether the slice generates positive examples or hard negatives), \texttt{type\_field} and \texttt{register\_hint} constrain the expected output, and \texttt{target\_count} bounds the requested number of items. For each lexicon category we issue one LLM call per (category, slice) pair, then aggregate and deduplicate the returned items before human review.

\begin{lstlisting}
    {
    "formal_single": {
        "instructions": (
            "Generate FORMAL-register SINGLE-WORD items. Targets: written, "
            "academic, legal, news-editorial, professional discourse. "
            "All items must be single tokens (no spaces). Set type='single'."
        ),
        "type_field": "single",
        "register_hint": "formal",
        "is_hard_negative": False,
        "target_count": 30,
    },
    "formal_phrase": {
        "instructions": (
            "Generate FORMAL-register N-GRAM items (multi-word expressions, "
            "fixed phrases, idiom-like locutions). Targets: 'in conclusion', "
            "'on the grounds that', 'it follows that'. Each item must contain "
            "at least one space (>=2 tokens). Set type='phrase'."
        ),
        "type_field": "phrase",
        "register_hint": "formal",
        "is_hard_negative": False,
        "target_count": 25,
    },
    "colloquial_single": {
        "instructions": (
            "Generate COLLOQUIAL / SPOKEN single-word items. Targets: spoken "
            "English, informal writing, social-media discourse. Include "
            "interjections, informal contractions where they function "
            "lexically (e.g., 'yeah', 'nah', 'totally'). All items single "
            "tokens. Set type='single'."
        ),
        "type_field": "single",
        "register_hint": "colloquial",
        "is_hard_negative": False,
        "target_count": 20,
    },
    "colloquial_phrase": {
        "instructions": (
            "Generate COLLOQUIAL / SPOKEN N-grams that DIRECTLY perform "
            "the rhetorical function of THIS category, expressed in casual "
            "register. Each item >=2 tokens.\n\n"
            "CRITICAL: Do NOT output generic conversational fillers, "
            "hedges, or topic-management phrases (e.g. 'you know what', "
            "'I mean', 'to be honest', 'at the end of the day', 'the "
            "bottom line is', 'that being said') unless they LITERALLY "
            "encode the category's function. These items belong to other "
            "PI categories (Filler, By.The.Way, etc.) generating them "
            "here is category leakage.\n"
            "Test each candidate: 'If I delete the candidate from the "
            "sentence, does the {category} function disappear?' If the "
            "answer is no (the rhetorical function survives without it), "
            "do not output it.\n"
            "Examples of good output: for IMPACT_GAIN, 'pays off big', "
            "'comes out ahead', 'good deal'; for LOGIC_CAUSAL, 'that's "
            "why', 'that's because'.\n"
            "Set type='phrase'."
        ),
        "type_field": "phrase",
        "register_hint": "colloquial",
        "is_hard_negative": False,
        "target_count": 20,
    },
    "domain_specific": {
        "instructions": (
            "Generate items where THIS category's function appears in a "
            "specialised register (legal, academic, scientific, "
            "political, business). The item must still perform THIS "
            "category's rhetorical function not a different one that "
            "happens to share the domain.\n\n"
            "CRITICAL: Do NOT mechanically reach for examples from other "
            "PI categories (e.g., academic stance verbs like 'postulate' "
            "or 'we contend' are ARG_CLAIM, not LOGIC_CAUSAL). Re-read "
            "the FUNCTION block above and confirm the candidate satisfies "
            "it. Mark the source domain in the rationale (e.g., 'legal', "
            "'academic'). type may be 'single' or 'phrase'."
        ),
        "type_field": "single_or_phrase",
        "register_hint": "domain",
        "is_hard_negative": False,
        "target_count": 20,
    },
    "archaic_literary": {
        "instructions": (
            "Generate items that perform the category function in a "
            "GENUINELY archaic, literary, or historically-marked register. "
            "Targets: items still encountered in religious, legal, or "
            "literary text but rare in modern conversational/journalistic "
            "prose.\n\n"
            "CRITICAL: Do NOT relabel modern, high-frequency items as "
            "archaic. 'therefore', 'hence', 'thus', 'reward', 'prosperity', "
            "'advantage' are common modern words; they are not archaic and "
            "should NOT appear in this slice. True archaic examples: "
            "'wherefore', 'verily', 'forsooth', 'whence', 'henceforth', "
            "'inasmuch as', 'thereupon', 'hereby'. If the category has no "
            "genuine archaic-register variants, output an EMPTY list "
            "this is the correct answer for many categories. Quality over "
            "count. type='single' or 'phrase'."
        ),
        "type_field": "single_or_phrase",
        "register_hint": "archaic",
        "is_hard_negative": False,
        "target_count": 15,
    },
    "inflectional": {
        "instructions": (
            "Generate INFLECTIONAL / DERIVATIONAL VARIANTS of the canonical "
            "seeds and other category members. Include third-person -s, "
            "past -ed, gerund -ing, agent -er/-or, negation prefixes when "
            "they preserve function.\n\n"
            "CRITICAL: If the canonical seeds are CLOSED-CLASS items "
            "(connectives, modals, demonstratives, fixed phrases) that do "
            "not inflect, output an EMPTY list. Examples of categories "
            "where you should output nothing: LOGIC_CAUSAL ('because' / "
            "'therefore' / 'hence' do not inflect); LOGIC_CONTRAST; "
            "LOGIC_REFERENCE; SPEC_PSYCH_NEAR (deictic 'now' / 'here'); "
            "PROP_BANDWAGON ('everyone' as quantifier).\n"
            "Never invent non-existent forms like 'therefored' or "
            "'consequentlying'. Zero output is the correct answer when "
            "the category is non-inflectional.\n\n"
            "Only items whose function is genuinely preserved. Set "
            "type='inflection'."
        ),
        "type_field": "inflection",
        "register_hint": "neutral",
        "is_hard_negative": False,
        "target_count": 25,
    },
}
\end{lstlisting}

\subsection{Prompts to expand lexicons}
\label{appendix:llm_expansion-expansion_prompts}

The full prompt template, instantiated for the \texttt{EVI\_UNITS} category and the \texttt{formal\_single} slice, is shown below. Bracketed section headers in color demarcate the framework structure.
\begin{lstlisting}[style=promptstyle]
(*\textbf{\textcolor{sectionhead}{[CATEGORY]}}*)
Code: EVI_UNITS
Aristotelian appeal: Logos
PI dimension: Evidence
PI sub-feature: statistical
(*\textbf{\textcolor{sectionhead}{[FUNCTION]}}*)
Units of measurement, statistical markers, currencies, and quantitative indicators that ground claims in verifiable numerical evidence.
(*\textbf{\textcolor{sectionhead}{[THEORETICAL ANCHOR]}}*)
Zebregs et al. (2015) on statistical evidence; targets cognitive belief change via factual grounding.
(*\textbf{\textcolor{sectionhead}{[INCLUSION CRITERIA]}}*)
- SI / imperial units of measure (kg, mi, ml, ft, C, mph)
- Currency symbols and codes (USD, EUR, $, JPY)
- Statistical indicators (p-value, percentage, ratio, coefficient, significance)
- Magnitude words used in numerical context (million, billion, thousand)
(*\textbf{\textcolor{sectionhead}{[EXCLUSION CRITERIA]}}*)
- Polysemous letters with non-measurement meanings in dominant English use
- Generic quantifiers without measurement function (some, many, several) -- those belong to SPEC_VAGUE
- Number words used as ordinals/names (first, second as ranks)
(*\textbf{\textcolor{sectionhead}{[KNOWN DRIFT TRAPS]}}*)
- "size" (HYPERNYM_DILUTION): Generic noun, not a measurement unit
- "many" (TOPICAL_NEIGHBOR): Vague quantifier; belongs to SPEC_VAGUE
(*\textbf{\textcolor{sectionhead}{[CANONICAL SEEDS]}}*)
%, kg, million, percentage, p-value, USD
(*\textbf{\textcolor{sectionhead}{[SLICE: formal\_single]}}*)
Generate FORMAL-register SINGLE-WORD items. Targets: written, academic, legal, news-editorial, professional discourse. All items must be single tokens (no spaces). Set type='single'.
Target output count: ~30 items (more is fine if quality holds; fewer is fine if the category is genuinely small).

(*\textbf{\textcolor{sectionhead}{[OUTPUT SCHEMA]}}*)
Return STRICT JSON in exactly this shape -- no prose, no fences:
{
  "items": [
    {
      "word": "<the lexicon item, lowercase unless capitalisation is intrinsic>",
      "register": "<formal | colloquial | domain | archaic | neutral>",
      "type": "<single | phrase | inflection | antonym | boundary>",
      "confidence": "<high | medium | low>",
      "rationale": "<one sentence; cite the specific INCLUDE or EXCLUDE rule that applies>"
    },
    ...
  ]
}
Field rules:
- "word" must be the surface form (not a regex pattern, not a list).
- "type" must equal "single" for this slice unless the slice is "domain_specific" or "archaic_literary" (in which case use "single" or "phrase").
- "register" should reflect the dominant register for the item.
- "confidence": use "high" only when no reasonable linguist would dispute admission; use "low" for borderline items that should go to human review.
- "rationale": one sentence. Cite the operative rule by paraphrase. No hedging.
Do not include items already given by the canonical seeds -- those are inputs, not outputs.
\end{lstlisting}

\section{Public Web Interface}
\label{appendix:web_interface}

\begin{figure*}[t]
\centering
\includegraphics[width=\textwidth]{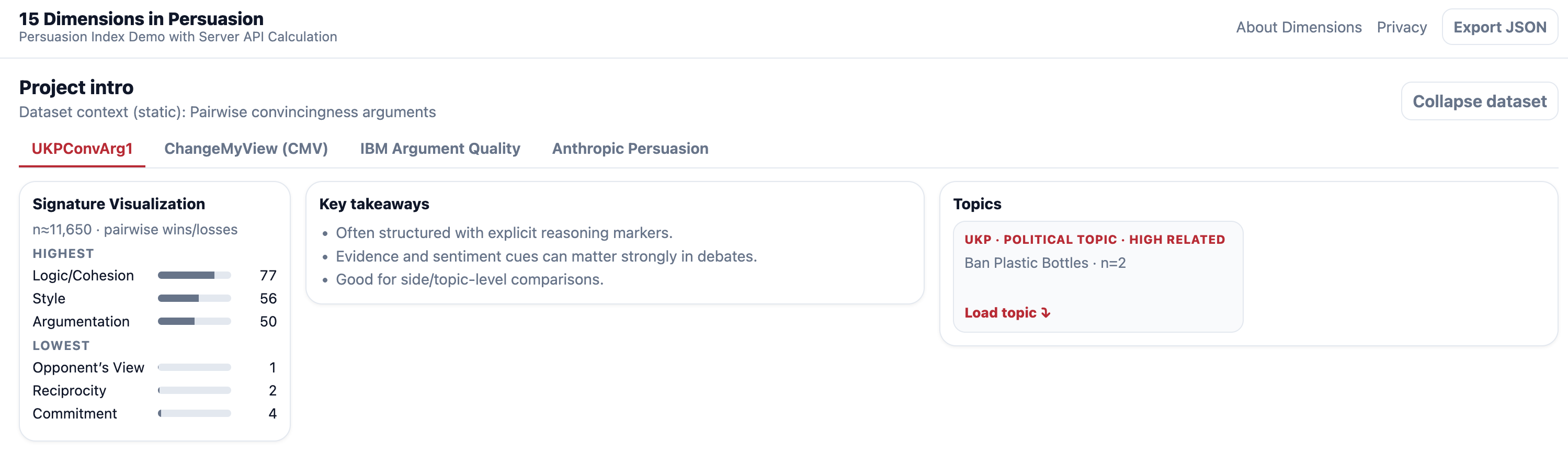}
\caption{Dataset overview panel showing corpus-level PI dimension distributions and key takeaways.}
\label{fig:web_overview}
\end{figure*}

\begin{figure*}[t]
\centering
\includegraphics[width=\textwidth]{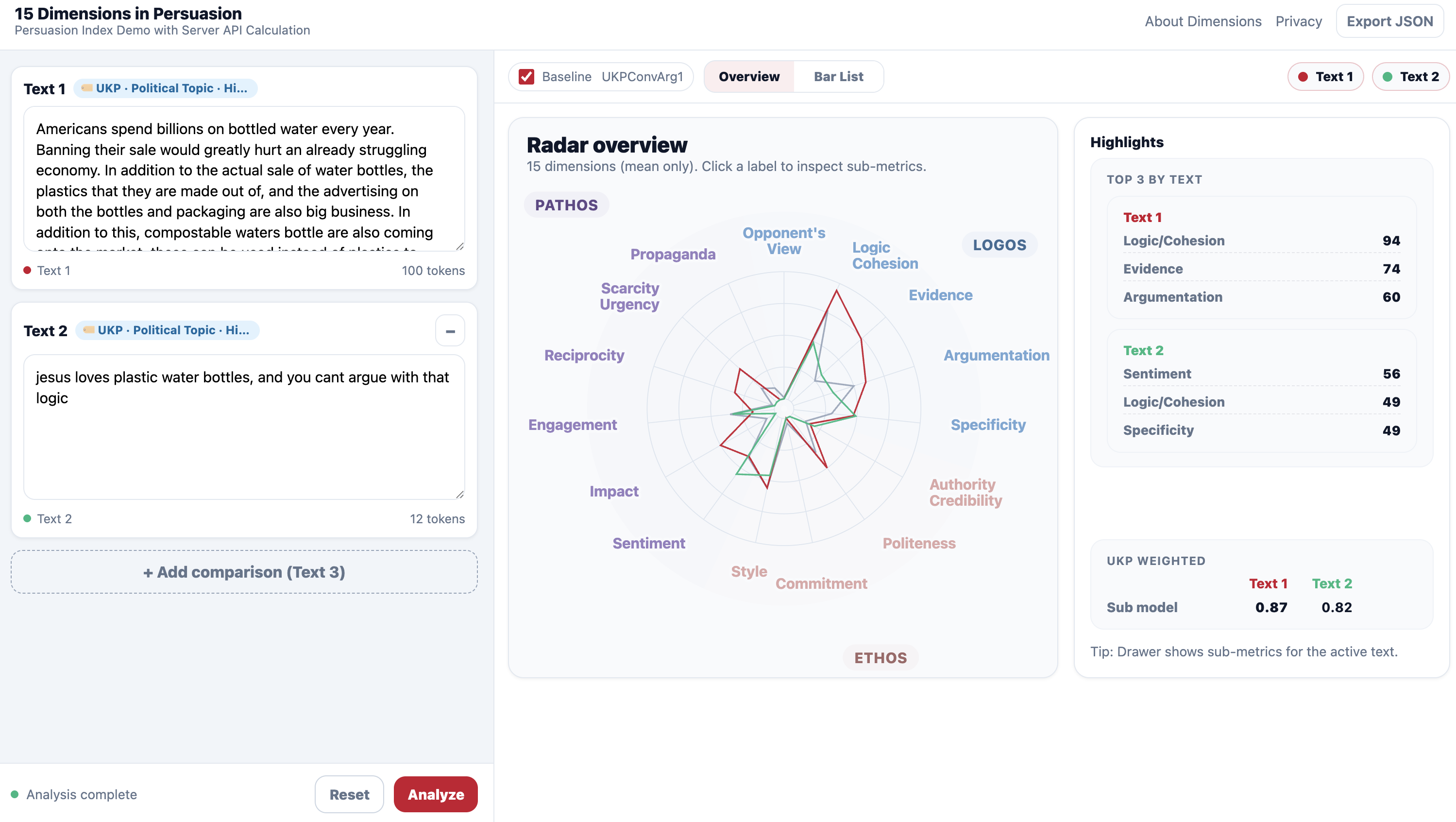}
\caption{Analysis panel showing pairwise radar visualization across 15 PI dimensions with per-text highlights and weighted persuasiveness scores.}
\label{fig:web_radar}
\end{figure*}

Figure~\ref{fig:web_overview} and Figure~\ref{fig:web_radar} illustrate the public-facing Persuasion Index web interface, accessible at \url{https://pi.umiacs.umd.edu}. The tool allows users to input one or more arguments and receive real-time PI scores computed via server-side API calls.
The dataset overview panel (Figure~\ref{fig:web_overview}) provides dataset-level context for each of the four evaluation corpora, displaying the distribution of PI dimension scores across the corpus and surfacing the highest- and lowest-scoring dimensions as a reference baseline. Users can select a dataset context (e.g., UKPConvArg1) to calibrate their interpretation of scores against empirical norms.
The analysis panel (Figure~\ref{fig:web_radar}) supports pairwise or single-argument analysis. For each submitted text, the interface renders a radar chart across all 15 PI dimensions organized under Logos, Ethos, and Pathos, highlights the top-3 dimensions per text, and optionally computes a dataset-weighted persuasiveness score using the logistic regression coefficients estimated in our experiments. Raw scores can be exported as JSON for downstream use.

\section{Use of AI Assistants}

AI Assistants are used for less than 10\% of coding and writing in this work.

\end{document}